\documentclass[11pt]{article}

\PassOptionsToPackage{hyphens}{url}

\usepackage[final]{acl}

\usepackage{times}
\usepackage{latexsym}
\usepackage[T1]{fontenc}
\usepackage[utf8]{inputenc}
\usepackage{microtype}
\usepackage[htt]{hyphenat}
\usepackage{inconsolata}
\usepackage{graphicx}
\usepackage[most]{tcolorbox}
\tcbset{promptbox/.style={
  colback=white, colframe=black!55, boxrule=0.5pt, arc=1.5pt,
  fonttitle=\bfseries\small, coltitle=white, colbacktitle=black!62,
  left=4pt, right=4pt, top=3pt, bottom=3pt, toptitle=1pt, bottomtitle=1pt,
  fontupper=\footnotesize}}
\usepackage{amsmath}
\usepackage{amssymb}
\usepackage{booktabs}
\usepackage{multirow}
\usepackage{xcolor}
\usepackage{subcaption}
\usepackage{algorithm}
\usepackage{algpseudocode}
\usepackage{tikz}
\usetikzlibrary{shapes.geometric, arrows.meta, positioning, calc, fit, backgrounds}
\usepackage{hyperref}


\usepackage{caption}
\title{Re:CAP -- Auditing Retrieval Coverage in Production RAG Pipelines}

\author{
  \textbf{Aviral Joshi}, \textbf{Hanoz Bhathena},
  \textbf{Max Nelson}, \textbf{Saket Sharma}
\\
  Machine Learning Center of Excellence, JPMorgan Chase \& Co.
\\
  \small{
    \texttt{\{aviral.joshi, hanoz.bhathena,}
    \texttt{max.nelson, saket.sharma\}@jpmchase.com}
  }
}

\begin{document}
\maketitle

\begin{abstract}
Retrieval-augmented generation (RAG) is hard to monitor in
production: exhaustive relevance labels do not exist for non-stationary
multi-million-passage corpora that re-index in real time. As a result, retrieval
quality is generally understudied and often deprioritised in favour of generation-oriented metrics.
In this work, we propose auditing retrieval coverage by probing
for evidence of missing documents rather than enumerating every
relevant one. Our method Re:CAP (REtrieval Coverage Audit by iterative
Probing) is a reference-free audit loop applied to a deployed
RAG pipeline's initial answer and retrieved context: it
identifies the topics already covered, generates probing
questions for plausibly missing topics, retrieves candidate
documents, and applies an LLM-as-judge to retain only
those that introduce previously-unretrieved information.  On
four public benchmarks, Re:CAP recovers $9$--$29\%$ of gold labels
that flat BM25 top-$500$ cannot reach, rising to $48\%$ on
TREC-COVID. On MuSiQue Re:CAP beats flat hybrid top-$500$ by $+12.9$\,pp on
recall at less than half the document budget. An ensemble
BM25, dense, and hybrid baseline (top-$500$ each) still leaves out $21.2\%$ of
gold docs on TREC-COVID that Re:CAP recovers; human annotators judge that $78.9\%$ of
those structurally distinct documents add new information to the
baseline answer (Fleiss $\kappa = 0.79$, $n = 123$), and $73.9\%$ on live
production traffic ($n = 180$).
End-to-end recall is reproducible to within
$\pm 1\%$ across three independent runs, making Re:CAP a stable instrument for periodic retrieval audits.
\end{abstract}

\section{Introduction}
\label{sec:intro}

\begin{figure*}[t]
\centering
\begin{tikzpicture}[
    corpusframe/.style={draw=black!14, rounded corners=1.5pt, fill=black!3},
    corpusdot/.style={black!18},
    beamnear/.style={fill=blue!30, opacity=0.72},
    beamdeep/.style={fill=blue!18, opacity=0.72},
    beamspent/.style={fill=black!8, opacity=0.68},
    beamedge/.style={draw=blue!55, line width=0.5pt, densely dashed},
    probefill/.style={fill=green!65!black!55, opacity=0.6},
    probeedge/.style={draw=green!60!black, line width=0.5pt, densely dashed},
    goldfound/.style={draw=orange!55!black, fill=orange!72, line width=0.35pt},
    goldmissed/.style={draw=orange!60!black, fill=white, line width=0.5pt},
    goldrecovered/.style={draw=orange!55!black, fill=orange!72, line width=0.4pt},
    querydot/.style={circle, fill=black!85, inner sep=1.5pt},
    qlabel/.style={font=\footnotesize\bfseries},
    beamtag/.style={font=\tiny, text=blue!55!black},
    panelttl/.style={font=\small},
    panelsub/.style={font=\scriptsize, text=black!65, align=left},
    legtxt/.style={font=\scriptsize, text=black!70},
    callout/.style={font=\scriptsize, align=left, fill=white, draw=black!35,
                    rounded corners=2pt, inner sep=3pt},
]

\begin{scope}
  \draw[corpusframe] (0,0) rectangle (7.4,4.9);

  \begin{scope}
  \clip (0,0) rectangle (7.4,4.9);
  \fill[beamdeep] (0.55,2.45) -- ++(-21.0:5.65) arc[start angle=-21.0, end angle=21.0, radius=5.65cm] -- cycle;
  \fill[beamnear] (0.55,2.45) -- ++(-21.0:3.05) arc[start angle=-21.0, end angle=21.0, radius=3.05cm] -- cycle;
  \draw[beamedge] (0.55,2.45) -- ++(-21.0:5.65) arc[start angle=-21.0, end angle=21.0, radius=5.65cm] -- cycle;
  \end{scope}

  \fill[corpusdot] (3.39,2.71) circle (0.045);
  \fill[corpusdot] (6.64,2.30) circle (0.045);
  \fill[corpusdot] (3.77,2.83) circle (0.045);
  \fill[corpusdot] (1.55,2.50) circle (0.045);
  \fill[corpusdot] (4.61,3.73) circle (0.045);
  \fill[corpusdot] (0.93,1.59) circle (0.045);
  \fill[corpusdot] (0.90,3.81) circle (0.045);
  \fill[corpusdot] (5.05,0.44) circle (0.045);
  \fill[corpusdot] (7.04,4.49) circle (0.045);
  \fill[corpusdot] (4.78,2.96) circle (0.045);
  \fill[corpusdot] (1.36,0.33) circle (0.045);
  \fill[corpusdot] (3.92,0.52) circle (0.045);
  \fill[corpusdot] (1.59,1.32) circle (0.045);
  \fill[corpusdot] (0.49,2.29) circle (0.045);
  \fill[corpusdot] (3.31,3.95) circle (0.045);
  \fill[corpusdot] (3.85,3.06) circle (0.045);
  \fill[corpusdot] (3.43,1.48) circle (0.045);
  \fill[corpusdot] (6.06,3.36) circle (0.045);
  \fill[corpusdot] (2.45,1.27) circle (0.045);
  \fill[corpusdot] (2.27,0.57) circle (0.045);
  \fill[corpusdot] (5.55,2.01) circle (0.045);
  \fill[corpusdot] (6.10,1.95) circle (0.045);
  \fill[corpusdot] (6.87,3.97) circle (0.045);
  \fill[corpusdot] (0.28,1.18) circle (0.045);
  \fill[corpusdot] (7.02,2.00) circle (0.045);
  \fill[corpusdot] (0.78,3.02) circle (0.045);
  \fill[corpusdot] (5.64,1.44) circle (0.045);
  \fill[corpusdot] (6.91,3.58) circle (0.045);
  \fill[corpusdot] (1.09,1.34) circle (0.045);
  \fill[corpusdot] (0.98,0.52) circle (0.045);
  \fill[corpusdot] (5.76,1.04) circle (0.045);
  \fill[corpusdot] (4.13,2.22) circle (0.045);
  \fill[corpusdot] (1.59,3.47) circle (0.045);
  \fill[corpusdot] (1.18,3.08) circle (0.045);
  \fill[corpusdot] (1.08,2.10) circle (0.045);
  \fill[corpusdot] (6.96,3.78) circle (0.045);
  \fill[corpusdot] (2.37,4.14) circle (0.045);
  \fill[corpusdot] (1.73,1.99) circle (0.045);
  \fill[corpusdot] (6.16,3.07) circle (0.045);
  \fill[corpusdot] (0.97,4.59) circle (0.045);
  \fill[corpusdot] (5.60,1.70) circle (0.045);
  \fill[corpusdot] (0.90,2.81) circle (0.045);
  \fill[corpusdot] (1.95,2.89) circle (0.045);
  \fill[corpusdot] (2.84,2.25) circle (0.045);
  \fill[corpusdot] (6.88,2.38) circle (0.045);
  \fill[corpusdot] (4.23,4.06) circle (0.045);
  \fill[corpusdot] (1.54,0.94) circle (0.045);
  \fill[corpusdot] (6.53,3.84) circle (0.045);
  \fill[corpusdot] (2.00,1.09) circle (0.045);
  \fill[corpusdot] (5.37,4.38) circle (0.045);
  \fill[corpusdot] (1.63,4.42) circle (0.045);
  \fill[corpusdot] (6.35,2.90) circle (0.045);
  \fill[corpusdot] (3.18,0.71) circle (0.045);
  \fill[corpusdot] (0.55,4.48) circle (0.045);
  \fill[corpusdot] (1.92,3.35) circle (0.045);
  \fill[corpusdot] (2.05,3.87) circle (0.045);
  \fill[corpusdot] (4.38,1.55) circle (0.045);
  \fill[corpusdot] (0.75,1.26) circle (0.045);
  \fill[corpusdot] (6.59,1.15) circle (0.045);
  \fill[corpusdot] (0.39,1.44) circle (0.045);
  \fill[corpusdot] (3.35,0.52) circle (0.045);
  \fill[corpusdot] (1.49,1.88) circle (0.045);
  \fill[corpusdot] (4.22,0.84) circle (0.045);
  \fill[corpusdot] (2.77,4.16) circle (0.045);
  \fill[corpusdot] (7.03,3.14) circle (0.045);
  \fill[corpusdot] (5.04,2.82) circle (0.045);
  \fill[corpusdot] (0.40,4.25) circle (0.045);
  \fill[corpusdot] (5.10,4.48) circle (0.045);
  \fill[corpusdot] (0.43,3.05) circle (0.045);
  \fill[corpusdot] (3.60,3.46) circle (0.045);
  \fill[corpusdot] (2.47,4.64) circle (0.045);
  \fill[corpusdot] (5.35,3.34) circle (0.045);
  \fill[corpusdot] (5.74,4.27) circle (0.045);
  \fill[corpusdot] (2.70,3.26) circle (0.045);
  \fill[corpusdot] (6.48,4.08) circle (0.045);
  \fill[corpusdot] (3.15,3.72) circle (0.045);
  \fill[corpusdot] (4.58,1.93) circle (0.045);
  \fill[corpusdot] (4.29,2.93) circle (0.045);
  \fill[corpusdot] (7.11,4.11) circle (0.045);
  \fill[corpusdot] (5.29,1.96) circle (0.045);
  \fill[corpusdot] (5.34,2.80) circle (0.045);
  \fill[corpusdot] (0.86,3.55) circle (0.045);
  \fill[corpusdot] (3.59,1.27) circle (0.045);
  \fill[corpusdot] (5.08,2.44) circle (0.045);
  \fill[corpusdot] (4.51,4.29) circle (0.045);
  \fill[corpusdot] (2.04,0.31) circle (0.045);
  \fill[corpusdot] (2.35,3.23) circle (0.045);
  \fill[corpusdot] (6.51,3.15) circle (0.045);
  \fill[corpusdot] (3.32,4.17) circle (0.045);
  \fill[corpusdot] (6.34,1.94) circle (0.045);
  \fill[corpusdot] (1.22,2.43) circle (0.045);
  \fill[corpusdot] (6.04,3.98) circle (0.045);
  \fill[corpusdot] (2.18,1.00) circle (0.045);
  \fill[corpusdot] (4.59,2.42) circle (0.045);
  \fill[corpusdot] (2.45,3.94) circle (0.045);
  \fill[corpusdot] (7.04,2.24) circle (0.045);
  \fill[corpusdot] (0.79,0.40) circle (0.045);
  \fill[corpusdot] (6.29,0.44) circle (0.045);
  \fill[corpusdot] (2.41,3.73) circle (0.045);
  \fill[corpusdot] (0.41,0.86) circle (0.045);
  \fill[corpusdot] (5.99,1.30) circle (0.045);
  \fill[corpusdot] (4.61,2.22) circle (0.045);
  \fill[corpusdot] (4.61,3.13) circle (0.045);
  \fill[corpusdot] (5.83,4.46) circle (0.045);
  \fill[corpusdot] (4.99,1.13) circle (0.045);
  \fill[corpusdot] (3.55,1.04) circle (0.045);
  \fill[corpusdot] (5.19,1.04) circle (0.045);
  \fill[corpusdot] (2.15,1.77) circle (0.045);
  \fill[corpusdot] (6.51,0.64) circle (0.045);
  \fill[corpusdot] (6.70,3.42) circle (0.045);
  \fill[corpusdot] (2.87,2.75) circle (0.045);
  \fill[corpusdot] (6.33,3.75) circle (0.045);
  \fill[corpusdot] (4.76,1.16) circle (0.045);
  \fill[corpusdot] (5.25,3.84) circle (0.045);
  \fill[corpusdot] (4.69,3.40) circle (0.045);
  \fill[corpusdot] (1.75,4.20) circle (0.045);
  \fill[corpusdot] (3.97,3.72) circle (0.045);
  \fill[corpusdot] (0.85,2.19) circle (0.045);
  \fill[corpusdot] (3.63,0.38) circle (0.045);
  \fill[corpusdot] (5.85,0.54) circle (0.045);
  \fill[corpusdot] (1.25,0.91) circle (0.045);
  \fill[corpusdot] (3.83,3.43) circle (0.045);
  \fill[corpusdot] (6.81,0.64) circle (0.045);
  \fill[corpusdot] (1.80,2.57) circle (0.045);
  \fill[corpusdot] (2.28,3.45) circle (0.045);
  \fill[corpusdot] (6.08,2.71) circle (0.045);
  \fill[corpusdot] (2.42,1.93) circle (0.045);
  \fill[corpusdot] (6.10,4.20) circle (0.045);
  \fill[corpusdot] (1.71,3.99) circle (0.045);
  \fill[corpusdot] (4.22,1.14) circle (0.045);
  \fill[corpusdot] (3.97,2.46) circle (0.045);
  \fill[corpusdot] (4.44,0.38) circle (0.045);
  \fill[corpusdot] (3.99,2.07) circle (0.045);
  \fill[corpusdot] (6.86,4.30) circle (0.045);
  \fill[corpusdot] (2.13,2.33) circle (0.045);
  \fill[corpusdot] (3.56,1.65) circle (0.045);
  \fill[corpusdot] (1.60,2.97) circle (0.045);
  \fill[corpusdot] (6.65,0.83) circle (0.045);
  \fill[corpusdot] (5.64,0.36) circle (0.045);
  \fill[corpusdot] (5.01,1.67) circle (0.045);
  \fill[corpusdot] (2.72,2.97) circle (0.045);
  \fill[corpusdot] (3.92,0.96) circle (0.045);
  \fill[corpusdot] (6.49,1.64) circle (0.045);
  \fill[corpusdot] (1.78,4.63) circle (0.045);
  \fill[corpusdot] (6.39,0.85) circle (0.045);
  \fill[corpusdot] (2.07,0.68) circle (0.045);
  \fill[corpusdot] (5.71,0.81) circle (0.045);
  \fill[corpusdot] (3.05,3.26) circle (0.045);
  \fill[corpusdot] (4.97,4.25) circle (0.045);
  \fill[corpusdot] (1.58,0.57) circle (0.045);
  \fill[corpusdot] (6.06,4.60) circle (0.045);
  \fill[corpusdot] (3.46,3.61) circle (0.045);
  \fill[corpusdot] (2.53,2.31) circle (0.045);
  \fill[corpusdot] (1.53,2.25) circle (0.045);
  \fill[corpusdot] (5.13,4.03) circle (0.045);
  \fill[corpusdot] (2.06,4.25) circle (0.045);
  \fill[corpusdot] (5.40,3.67) circle (0.045);
  \fill[corpusdot] (5.88,2.04) circle (0.045);
  \fill[corpusdot] (5.06,3.62) circle (0.045);
  \fill[corpusdot] (5.25,0.57) circle (0.045);
  \fill[corpusdot] (0.35,1.82) circle (0.045);
  \fill[corpusdot] (1.88,4.40) circle (0.045);
  \fill[corpusdot] (4.82,2.75) circle (0.045);
  \fill[corpusdot] (7.02,0.36) circle (0.045);
  \fill[corpusdot] (4.51,3.50) circle (0.045);
  \fill[corpusdot] (2.05,2.02) circle (0.045);
  \fill[corpusdot] (2.86,0.69) circle (0.045);
  \fill[corpusdot] (6.93,2.75) circle (0.045);
  \fill[corpusdot] (6.78,2.10) circle (0.045);
  \fill[corpusdot] (2.80,1.51) circle (0.045);
  \fill[corpusdot] (2.32,0.32) circle (0.045);
  \fill[corpusdot] (1.36,3.57) circle (0.045);
  \fill[corpusdot] (0.79,4.23) circle (0.045);
  \fill[corpusdot] (3.24,2.35) circle (0.045);
  \fill[corpusdot] (6.98,1.33) circle (0.045);
  \fill[corpusdot] (3.88,4.37) circle (0.045);
  \fill[corpusdot] (5.25,2.31) circle (0.045);
  \fill[corpusdot] (4.43,0.76) circle (0.045);
  \fill[corpusdot] (3.33,3.19) circle (0.045);
  \fill[corpusdot] (5.63,2.58) circle (0.045);
  \fill[corpusdot] (5.33,1.30) circle (0.045);
  \fill[corpusdot] (1.06,4.17) circle (0.045);
  \fill[corpusdot] (5.68,3.00) circle (0.045);
  \fill[corpusdot] (5.46,1.09) circle (0.045);
  \fill[corpusdot] (0.55,0.53) circle (0.045);
  \fill[corpusdot] (0.87,3.22) circle (0.045);
  \fill[corpusdot] (4.07,3.02) circle (0.045);
  \fill[corpusdot] (1.73,1.77) circle (0.045);
  \fill[corpusdot] (0.79,0.78) circle (0.045);
  \fill[corpusdot] (3.20,1.39) circle (0.045);
  \fill[corpusdot] (4.78,4.59) circle (0.045);
  \fill[corpusdot] (2.52,2.66) circle (0.045);
  \fill[corpusdot] (3.22,1.88) circle (0.045);
  \fill[corpusdot] (5.62,0.59) circle (0.045);
  \fill[corpusdot] (1.75,3.16) circle (0.045);
  \fill[corpusdot] (6.60,4.38) circle (0.045);
  \fill[corpusdot] (5.80,1.59) circle (0.045);
  \fill[corpusdot] (2.80,1.85) circle (0.045);
  \fill[corpusdot] (3.00,1.96) circle (0.045);
  \fill[corpusdot] (1.62,2.73) circle (0.045);
  \fill[corpusdot] (6.34,1.49) circle (0.045);
  \fill[corpusdot] (0.43,2.52) circle (0.045);
  \fill[corpusdot] (4.02,2.75) circle (0.045);
  \fill[corpusdot] (1.80,3.61) circle (0.045);
  \fill[corpusdot] (2.99,1.74) circle (0.045);
  \fill[corpusdot] (3.68,2.61) circle (0.045);
  \fill[corpusdot] (6.58,2.06) circle (0.045);
  \fill[corpusdot] (5.96,3.18) circle (0.045);
  \fill[corpusdot] (5.38,4.60) circle (0.045);
  \fill[corpusdot] (2.86,0.99) circle (0.045);

  \node[star, star points=5, star point ratio=2.2, inner sep=1.35pt, goldfound] at (5.59,1.08) {};
  \node[star, star points=5, star point ratio=2.2, inner sep=1.35pt, goldfound] at (4.36,1.27) {};
  \node[star, star points=5, star point ratio=2.2, inner sep=1.35pt, goldfound] at (2.45,2.66) {};
  \node[star, star points=5, star point ratio=2.2, inner sep=1.35pt, goldfound] at (5.87,2.36) {};
  \node[star, star points=5, star point ratio=2.2, inner sep=1.35pt, goldfound] at (3.91,2.52) {};
  \node[star, star points=5, star point ratio=2.2, inner sep=1.35pt, goldfound] at (4.79,1.33) {};
  \node[star, star points=5, star point ratio=2.2, inner sep=1.35pt, goldfound] at (5.16,2.90) {};
  \node[star, star points=5, star point ratio=2.2, inner sep=1.35pt, goldfound] at (2.89,2.62) {};
  \node[star, star points=5, star point ratio=2.2, inner sep=1.35pt, goldfound] at (5.29,1.72) {};
  \node[star, star points=5, star point ratio=2.2, inner sep=1.35pt, goldfound] at (4.92,3.34) {};
  \node[star, star points=5, star point ratio=2.2, inner sep=1.35pt, goldfound] at (3.80,3.08) {};
  \node[star, star points=5, star point ratio=2.2, inner sep=1.35pt, goldfound] at (5.53,3.82) {};
  \node[star, star points=5, star point ratio=2.2, inner sep=1.35pt, goldfound] at (5.92,2.92) {};
  \node[star, star points=5, star point ratio=2.2, inner sep=1.35pt, goldfound] at (4.24,3.29) {};
  \node[star, star points=5, star point ratio=2.2, inner sep=1.35pt, goldfound] at (3.86,1.44) {};
  \node[star, star points=5, star point ratio=2.2, inner sep=1.35pt, goldmissed] at (2.51,0.55) {};
  \node[star, star points=5, star point ratio=2.2, inner sep=1.35pt, goldmissed] at (2.95,4.28) {};
  \node[star, star points=5, star point ratio=2.2, inner sep=1.35pt, goldmissed] at (1.82,4.54) {};
  \node[star, star points=5, star point ratio=2.2, inner sep=1.35pt, goldmissed] at (3.14,0.37) {};
  \node[star, star points=5, star point ratio=2.2, inner sep=1.35pt, goldmissed] at (2.49,3.81) {};
  \node[star, star points=5, star point ratio=2.2, inner sep=1.35pt, goldmissed] at (1.32,3.64) {};
  \node[star, star points=5, star point ratio=2.2, inner sep=1.35pt, goldmissed] at (1.09,0.48) {};

  \node[querydot] (qa) at (0.55,2.45) {};
  \node[qlabel, left=0.5mm of qa] {$Q$};

  \node[beamtag, anchor=west] at (1.28,2.75) {top-$k$};
  \node[beamtag, anchor=west] at (3.96,2.75) {top-$500$};

  \node[panelsub, anchor=south west, text width=7.20cm] (subA)
       at (0.06,4.90)
       {Raising $k$ extends the \emph{same} region; reranking only reorders
         within it. Gold outside stays unretrieved at any affordable depth.};
  \node[panelttl, anchor=south west] at ([yshift=0.15cm]subA.north west)
       {\textbf{One query reaches one region}};
\end{scope}

\begin{scope}[xshift=8.25cm]
  \draw[corpusframe] (0,0) rectangle (7.4,4.9);

  \begin{scope}
  \clip (0,0) rectangle (7.4,4.9);
  \fill[beamspent] (0.55,2.45) -- ++(-21.0:5.65) arc[start angle=-21.0, end angle=21.0, radius=5.65cm] -- cycle;
  \fill[probefill] (0.55,2.45) -- ++(-52.0:5.15) arc[start angle=-52.0, end angle=-34.0, radius=5.15cm] -- cycle;
  \draw[probeedge] (0.55,2.45) -- ++(-52.0:5.15) arc[start angle=-52.0, end angle=-34.0, radius=5.15cm] -- cycle;
  \fill[probefill] (0.55,2.45) -- ++(25.0:5.15) arc[start angle=25.0, end angle=43.0, radius=5.15cm] -- cycle;
  \draw[probeedge] (0.55,2.45) -- ++(25.0:5.15) arc[start angle=25.0, end angle=43.0, radius=5.15cm] -- cycle;
  \fill[probefill] (0.55,2.45) -- ++(48.0:5.15) arc[start angle=48.0, end angle=66.0, radius=5.15cm] -- cycle;
  \draw[probeedge] (0.55,2.45) -- ++(48.0:5.15) arc[start angle=48.0, end angle=66.0, radius=5.15cm] -- cycle;
  \end{scope}

  \fill[corpusdot] (3.39,2.71) circle (0.045);
  \fill[corpusdot] (6.64,2.30) circle (0.045);
  \fill[corpusdot] (3.77,2.83) circle (0.045);
  \fill[corpusdot] (1.55,2.50) circle (0.045);
  \fill[corpusdot] (4.61,3.73) circle (0.045);
  \fill[corpusdot] (0.93,1.59) circle (0.045);
  \fill[corpusdot] (0.90,3.81) circle (0.045);
  \fill[corpusdot] (5.05,0.44) circle (0.045);
  \fill[corpusdot] (7.04,4.49) circle (0.045);
  \fill[corpusdot] (4.78,2.96) circle (0.045);
  \fill[corpusdot] (1.36,0.33) circle (0.045);
  \fill[corpusdot] (3.92,0.52) circle (0.045);
  \fill[corpusdot] (1.59,1.32) circle (0.045);
  \fill[corpusdot] (0.49,2.29) circle (0.045);
  \fill[corpusdot] (3.31,3.95) circle (0.045);
  \fill[corpusdot] (3.85,3.06) circle (0.045);
  \fill[corpusdot] (3.43,1.48) circle (0.045);
  \fill[corpusdot] (6.06,3.36) circle (0.045);
  \fill[corpusdot] (2.45,1.27) circle (0.045);
  \fill[corpusdot] (2.27,0.57) circle (0.045);
  \fill[corpusdot] (5.55,2.01) circle (0.045);
  \fill[corpusdot] (6.10,1.95) circle (0.045);
  \fill[corpusdot] (6.87,3.97) circle (0.045);
  \fill[corpusdot] (0.28,1.18) circle (0.045);
  \fill[corpusdot] (7.02,2.00) circle (0.045);
  \fill[corpusdot] (0.78,3.02) circle (0.045);
  \fill[corpusdot] (5.64,1.44) circle (0.045);
  \fill[corpusdot] (6.91,3.58) circle (0.045);
  \fill[corpusdot] (1.09,1.34) circle (0.045);
  \fill[corpusdot] (0.98,0.52) circle (0.045);
  \fill[corpusdot] (5.76,1.04) circle (0.045);
  \fill[corpusdot] (4.13,2.22) circle (0.045);
  \fill[corpusdot] (1.59,3.47) circle (0.045);
  \fill[corpusdot] (1.18,3.08) circle (0.045);
  \fill[corpusdot] (1.08,2.10) circle (0.045);
  \fill[corpusdot] (6.96,3.78) circle (0.045);
  \fill[corpusdot] (2.37,4.14) circle (0.045);
  \fill[corpusdot] (1.73,1.99) circle (0.045);
  \fill[corpusdot] (6.16,3.07) circle (0.045);
  \fill[corpusdot] (0.97,4.59) circle (0.045);
  \fill[corpusdot] (5.60,1.70) circle (0.045);
  \fill[corpusdot] (0.90,2.81) circle (0.045);
  \fill[corpusdot] (1.95,2.89) circle (0.045);
  \fill[corpusdot] (2.84,2.25) circle (0.045);
  \fill[corpusdot] (6.88,2.38) circle (0.045);
  \fill[corpusdot] (4.23,4.06) circle (0.045);
  \fill[corpusdot] (1.54,0.94) circle (0.045);
  \fill[corpusdot] (6.53,3.84) circle (0.045);
  \fill[corpusdot] (2.00,1.09) circle (0.045);
  \fill[corpusdot] (5.37,4.38) circle (0.045);
  \fill[corpusdot] (1.63,4.42) circle (0.045);
  \fill[corpusdot] (6.35,2.90) circle (0.045);
  \fill[corpusdot] (3.18,0.71) circle (0.045);
  \fill[corpusdot] (0.55,4.48) circle (0.045);
  \fill[corpusdot] (1.92,3.35) circle (0.045);
  \fill[corpusdot] (2.05,3.87) circle (0.045);
  \fill[corpusdot] (4.38,1.55) circle (0.045);
  \fill[corpusdot] (0.75,1.26) circle (0.045);
  \fill[corpusdot] (6.59,1.15) circle (0.045);
  \fill[corpusdot] (0.39,1.44) circle (0.045);
  \fill[corpusdot] (3.35,0.52) circle (0.045);
  \fill[corpusdot] (1.49,1.88) circle (0.045);
  \fill[corpusdot] (4.22,0.84) circle (0.045);
  \fill[corpusdot] (2.77,4.16) circle (0.045);
  \fill[corpusdot] (7.03,3.14) circle (0.045);
  \fill[corpusdot] (5.04,2.82) circle (0.045);
  \fill[corpusdot] (0.40,4.25) circle (0.045);
  \fill[corpusdot] (5.10,4.48) circle (0.045);
  \fill[corpusdot] (0.43,3.05) circle (0.045);
  \fill[corpusdot] (3.60,3.46) circle (0.045);
  \fill[corpusdot] (2.47,4.64) circle (0.045);
  \fill[corpusdot] (5.35,3.34) circle (0.045);
  \fill[corpusdot] (5.74,4.27) circle (0.045);
  \fill[corpusdot] (2.70,3.26) circle (0.045);
  \fill[corpusdot] (6.48,4.08) circle (0.045);
  \fill[corpusdot] (3.15,3.72) circle (0.045);
  \fill[corpusdot] (4.58,1.93) circle (0.045);
  \fill[corpusdot] (4.29,2.93) circle (0.045);
  \fill[corpusdot] (7.11,4.11) circle (0.045);
  \fill[corpusdot] (5.29,1.96) circle (0.045);
  \fill[corpusdot] (5.34,2.80) circle (0.045);
  \fill[corpusdot] (0.86,3.55) circle (0.045);
  \fill[corpusdot] (3.59,1.27) circle (0.045);
  \fill[corpusdot] (5.08,2.44) circle (0.045);
  \fill[corpusdot] (4.51,4.29) circle (0.045);
  \fill[corpusdot] (2.04,0.31) circle (0.045);
  \fill[corpusdot] (2.35,3.23) circle (0.045);
  \fill[corpusdot] (6.51,3.15) circle (0.045);
  \fill[corpusdot] (3.32,4.17) circle (0.045);
  \fill[corpusdot] (6.34,1.94) circle (0.045);
  \fill[corpusdot] (1.22,2.43) circle (0.045);
  \fill[corpusdot] (6.04,3.98) circle (0.045);
  \fill[corpusdot] (2.18,1.00) circle (0.045);
  \fill[corpusdot] (4.59,2.42) circle (0.045);
  \fill[corpusdot] (2.45,3.94) circle (0.045);
  \fill[corpusdot] (7.04,2.24) circle (0.045);
  \fill[corpusdot] (0.79,0.40) circle (0.045);
  \fill[corpusdot] (6.29,0.44) circle (0.045);
  \fill[corpusdot] (2.41,3.73) circle (0.045);
  \fill[corpusdot] (0.41,0.86) circle (0.045);
  \fill[corpusdot] (5.99,1.30) circle (0.045);
  \fill[corpusdot] (4.61,2.22) circle (0.045);
  \fill[corpusdot] (4.61,3.13) circle (0.045);
  \fill[corpusdot] (5.83,4.46) circle (0.045);
  \fill[corpusdot] (4.99,1.13) circle (0.045);
  \fill[corpusdot] (3.55,1.04) circle (0.045);
  \fill[corpusdot] (5.19,1.04) circle (0.045);
  \fill[corpusdot] (2.15,1.77) circle (0.045);
  \fill[corpusdot] (6.51,0.64) circle (0.045);
  \fill[corpusdot] (6.70,3.42) circle (0.045);
  \fill[corpusdot] (2.87,2.75) circle (0.045);
  \fill[corpusdot] (6.33,3.75) circle (0.045);
  \fill[corpusdot] (4.76,1.16) circle (0.045);
  \fill[corpusdot] (5.25,3.84) circle (0.045);
  \fill[corpusdot] (4.69,3.40) circle (0.045);
  \fill[corpusdot] (1.75,4.20) circle (0.045);
  \fill[corpusdot] (3.97,3.72) circle (0.045);
  \fill[corpusdot] (0.85,2.19) circle (0.045);
  \fill[corpusdot] (3.63,0.38) circle (0.045);
  \fill[corpusdot] (5.85,0.54) circle (0.045);
  \fill[corpusdot] (1.25,0.91) circle (0.045);
  \fill[corpusdot] (3.83,3.43) circle (0.045);
  \fill[corpusdot] (6.81,0.64) circle (0.045);
  \fill[corpusdot] (1.80,2.57) circle (0.045);
  \fill[corpusdot] (2.28,3.45) circle (0.045);
  \fill[corpusdot] (6.08,2.71) circle (0.045);
  \fill[corpusdot] (2.42,1.93) circle (0.045);
  \fill[corpusdot] (6.10,4.20) circle (0.045);
  \fill[corpusdot] (1.71,3.99) circle (0.045);
  \fill[corpusdot] (4.22,1.14) circle (0.045);
  \fill[corpusdot] (3.97,2.46) circle (0.045);
  \fill[corpusdot] (4.44,0.38) circle (0.045);
  \fill[corpusdot] (3.99,2.07) circle (0.045);
  \fill[corpusdot] (6.86,4.30) circle (0.045);
  \fill[corpusdot] (2.13,2.33) circle (0.045);
  \fill[corpusdot] (3.56,1.65) circle (0.045);
  \fill[corpusdot] (1.60,2.97) circle (0.045);
  \fill[corpusdot] (6.65,0.83) circle (0.045);
  \fill[corpusdot] (5.64,0.36) circle (0.045);
  \fill[corpusdot] (5.01,1.67) circle (0.045);
  \fill[corpusdot] (2.72,2.97) circle (0.045);
  \fill[corpusdot] (3.92,0.96) circle (0.045);
  \fill[corpusdot] (6.49,1.64) circle (0.045);
  \fill[corpusdot] (1.78,4.63) circle (0.045);
  \fill[corpusdot] (6.39,0.85) circle (0.045);
  \fill[corpusdot] (2.07,0.68) circle (0.045);
  \fill[corpusdot] (5.71,0.81) circle (0.045);
  \fill[corpusdot] (3.05,3.26) circle (0.045);
  \fill[corpusdot] (4.97,4.25) circle (0.045);
  \fill[corpusdot] (1.58,0.57) circle (0.045);
  \fill[corpusdot] (6.06,4.60) circle (0.045);
  \fill[corpusdot] (3.46,3.61) circle (0.045);
  \fill[corpusdot] (2.53,2.31) circle (0.045);
  \fill[corpusdot] (1.53,2.25) circle (0.045);
  \fill[corpusdot] (5.13,4.03) circle (0.045);
  \fill[corpusdot] (2.06,4.25) circle (0.045);
  \fill[corpusdot] (5.40,3.67) circle (0.045);
  \fill[corpusdot] (5.88,2.04) circle (0.045);
  \fill[corpusdot] (5.06,3.62) circle (0.045);
  \fill[corpusdot] (5.25,0.57) circle (0.045);
  \fill[corpusdot] (0.35,1.82) circle (0.045);
  \fill[corpusdot] (1.88,4.40) circle (0.045);
  \fill[corpusdot] (4.82,2.75) circle (0.045);
  \fill[corpusdot] (7.02,0.36) circle (0.045);
  \fill[corpusdot] (4.51,3.50) circle (0.045);
  \fill[corpusdot] (2.05,2.02) circle (0.045);
  \fill[corpusdot] (2.86,0.69) circle (0.045);
  \fill[corpusdot] (6.93,2.75) circle (0.045);
  \fill[corpusdot] (6.78,2.10) circle (0.045);
  \fill[corpusdot] (2.80,1.51) circle (0.045);
  \fill[corpusdot] (2.32,0.32) circle (0.045);
  \fill[corpusdot] (1.36,3.57) circle (0.045);
  \fill[corpusdot] (0.79,4.23) circle (0.045);
  \fill[corpusdot] (3.24,2.35) circle (0.045);
  \fill[corpusdot] (6.98,1.33) circle (0.045);
  \fill[corpusdot] (3.88,4.37) circle (0.045);
  \fill[corpusdot] (5.25,2.31) circle (0.045);
  \fill[corpusdot] (4.43,0.76) circle (0.045);
  \fill[corpusdot] (3.33,3.19) circle (0.045);
  \fill[corpusdot] (5.63,2.58) circle (0.045);
  \fill[corpusdot] (5.33,1.30) circle (0.045);
  \fill[corpusdot] (1.06,4.17) circle (0.045);
  \fill[corpusdot] (5.68,3.00) circle (0.045);
  \fill[corpusdot] (5.46,1.09) circle (0.045);
  \fill[corpusdot] (0.55,0.53) circle (0.045);
  \fill[corpusdot] (0.87,3.22) circle (0.045);
  \fill[corpusdot] (4.07,3.02) circle (0.045);
  \fill[corpusdot] (1.73,1.77) circle (0.045);
  \fill[corpusdot] (0.79,0.78) circle (0.045);
  \fill[corpusdot] (3.20,1.39) circle (0.045);
  \fill[corpusdot] (4.78,4.59) circle (0.045);
  \fill[corpusdot] (2.52,2.66) circle (0.045);
  \fill[corpusdot] (3.22,1.88) circle (0.045);
  \fill[corpusdot] (5.62,0.59) circle (0.045);
  \fill[corpusdot] (1.75,3.16) circle (0.045);
  \fill[corpusdot] (6.60,4.38) circle (0.045);
  \fill[corpusdot] (5.80,1.59) circle (0.045);
  \fill[corpusdot] (2.80,1.85) circle (0.045);
  \fill[corpusdot] (3.00,1.96) circle (0.045);
  \fill[corpusdot] (1.62,2.73) circle (0.045);
  \fill[corpusdot] (6.34,1.49) circle (0.045);
  \fill[corpusdot] (0.43,2.52) circle (0.045);
  \fill[corpusdot] (4.02,2.75) circle (0.045);
  \fill[corpusdot] (1.80,3.61) circle (0.045);
  \fill[corpusdot] (2.99,1.74) circle (0.045);
  \fill[corpusdot] (3.68,2.61) circle (0.045);
  \fill[corpusdot] (6.58,2.06) circle (0.045);
  \fill[corpusdot] (5.96,3.18) circle (0.045);
  \fill[corpusdot] (5.38,4.60) circle (0.045);
  \fill[corpusdot] (2.86,0.99) circle (0.045);

  \node[star, star points=5, star point ratio=2.2, inner sep=1.35pt, goldfound] at (5.59,1.08) {};
  \node[star, star points=5, star point ratio=2.2, inner sep=1.35pt, goldfound] at (4.36,1.27) {};
  \node[star, star points=5, star point ratio=2.2, inner sep=1.35pt, goldfound] at (2.45,2.66) {};
  \node[star, star points=5, star point ratio=2.2, inner sep=1.35pt, goldfound] at (5.87,2.36) {};
  \node[star, star points=5, star point ratio=2.2, inner sep=1.35pt, goldfound] at (3.91,2.52) {};
  \node[star, star points=5, star point ratio=2.2, inner sep=1.35pt, goldfound] at (4.79,1.33) {};
  \node[star, star points=5, star point ratio=2.2, inner sep=1.35pt, goldfound] at (5.16,2.90) {};
  \node[star, star points=5, star point ratio=2.2, inner sep=1.35pt, goldfound] at (2.89,2.62) {};
  \node[star, star points=5, star point ratio=2.2, inner sep=1.35pt, goldfound] at (5.29,1.72) {};
  \node[star, star points=5, star point ratio=2.2, inner sep=1.35pt, goldfound] at (4.92,3.34) {};
  \node[star, star points=5, star point ratio=2.2, inner sep=1.35pt, goldfound] at (3.80,3.08) {};
  \node[star, star points=5, star point ratio=2.2, inner sep=1.35pt, goldfound] at (5.53,3.82) {};
  \node[star, star points=5, star point ratio=2.2, inner sep=1.35pt, goldfound] at (5.92,2.92) {};
  \node[star, star points=5, star point ratio=2.2, inner sep=1.35pt, goldfound] at (4.24,3.29) {};
  \node[star, star points=5, star point ratio=2.2, inner sep=1.35pt, goldfound] at (3.86,1.44) {};
  \node[star, star points=5, star point ratio=2.2, inner sep=1.35pt, goldrecovered] at (2.51,0.55) {};
  \node[star, star points=5, star point ratio=2.2, inner sep=1.35pt, goldrecovered] at (2.95,4.28) {};
  \node[star, star points=5, star point ratio=2.2, inner sep=1.35pt, goldrecovered] at (1.82,4.54) {};
  \node[star, star points=5, star point ratio=2.2, inner sep=1.35pt, goldrecovered] at (3.14,0.37) {};
  \node[star, star points=5, star point ratio=2.2, inner sep=1.35pt, goldrecovered] at (2.49,3.81) {};
  \node[star, star points=5, star point ratio=2.2, inner sep=1.35pt, goldrecovered] at (1.32,3.64) {};
  \node[star, star points=5, star point ratio=2.2, inner sep=1.35pt, goldmissed] at (1.09,0.48) {};

  \node[querydot] (qb) at (0.55,2.45) {};
  \node[qlabel, left=0.5mm of qb] {$Q$};

  \node[panelsub, anchor=south west, text width=7.20cm] (subB)
       at (0.06,4.90)
       {The topic registry names facets the answer does not cover; each becomes
         a probe issued to the \emph{same} retriever.};
  \node[panelttl, anchor=south west] at ([yshift=0.15cm]subB.north west)
       {\textbf{Re:CAP probes for what is missing}};

  \node[callout, anchor=south east, text width=3.5cm] at (7.28,0.12)
       {Re:CAP recovers \textbf{9--29\%} unreachable gold
         (\textbf{48\%} on TREC-COVID)};
\end{scope}

\begin{scope}[yshift=-0.40cm]
  \fill[corpusdot] (0.35,0) circle (0.045);
  \node[legtxt, anchor=west] at (0.52,0) {corpus document};

  \node[star, star points=5, star point ratio=2.2, inner sep=1.35pt, goldfound] at (3.35,0) {};
  \node[legtxt, anchor=west] at (3.52,0) {gold, retrieved};

  \node[star, star points=5, star point ratio=2.2, inner sep=1.35pt, goldmissed] at (6.05,0) {};
  \node[legtxt, anchor=west] at (6.22,0) {gold, never retrieved};

  \fill[probefill] (9.45,-0.10) rectangle (9.79,0.10);
  \draw[probeedge] (9.45,-0.10) rectangle (9.79,0.10);
  \node[legtxt, anchor=west] at (9.92,0) {Re:CAP probe (gold inside was unreachable)};
\end{scope}
\end{tikzpicture}

\caption{\textbf{Schematic: retrieval coverage has a blind spot, and depth
does not close it.}  Gold ranked below any affordable top-$k$ by the original
query stays unretrieved, and answer-side evaluation cannot detect it.  Re:CAP
issues further queries from the topic registry; each induces a different
ranking over the same corpus and the same retriever
(\S\ref{sec:method}).  The cone geometry is illustrative, not a model of
the retriever.}
\label{fig:blindspot}
\end{figure*}
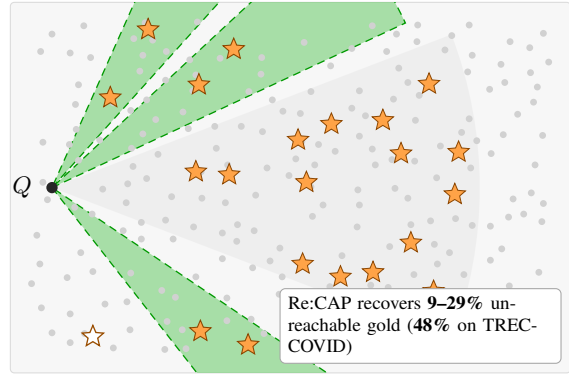

Retrieval-augmented generation (RAG) is a standard
component of enterprise knowledge assistants over proprietary
corpora such as news feeds, internal research, and regulatory
filings~\cite{lewis2020rag,gao2023ragsurvey}.
Popular RAG evaluation frameworks --- RAGAS~\cite{es2023ragas}
and ARES~\cite{saadfalcon2024ares} --- score answer-side
signals (faithfulness, answer relevance) and the relevance of
the retrieved context, but cannot detect what the retriever
failed to surface.  As a result, an answer that is fluent and
faithful to its retrieved context can still be substantively
incomplete, and these metrics do not register the omission.
For instance, a research assistant asked about a
recent central-bank policy announcement may
ground its answer in the policy statement and the lead wire
story while overlooking the updated economic projections, the
press-conference transcript, recorded dissents, and post-meeting
analyst commentary --- all present in the same corpus.
We refer to this gap between the documents the retriever
surfaces and the documents the corpus contains as the
\emph{retrieval coverage} problem, and we observe it most
acutely in non-stationary corpora with multiple sources of truth.
Figure~\ref{fig:blindspot} illustrates the problem: retrieving
more deeply extends the region a query already reaches, so gold that
lies outside it stays unreachable at any affordable depth.

Classical recall-based evaluation addresses retrieval coverage
in principle but is impractical for large production deployments.  Modern
production indices contain $10^8$ or more
passages~\cite{karpukhin2020dpr,bajaj2018msmarco}, and the
underlying corpus is re-indexed continuously as documents are
added, edited, or withdrawn.  Exhaustive per-query relevance
judgements are prohibitive at this scale, and TREC-style
pooling~\cite{voorhees2000variations,buckley2007bias} requires
multiple participating systems and fresh labels after every change
to the index, embedder, or ranker.  Proprietary deployments
typically have no per-query relevance labels at all, leaving
operators reliant on end-to-end answer signals that cannot
distinguish a retrieval failure from a reader failure.

We therefore reframe the problem.  Rather than enumerating every
relevant document (intractable), we probe each query for positive
evidence that the retriever missed relevant content (tractable
and automatable).  We instantiate this reframing in
\textbf{Re:CAP} (REtrieval Coverage Audit by iterative Probing), a
reference-free iterative loop that audits a deployed RAG pipeline
without consuming per-query gold labels.  Given a query $Q$, an
initial set of retrieved documents $D_0$, and the corresponding
answer $A_0$, Re:CAP
(Algorithm~\ref{alg:recap})
maintains a topic registry $T$ of the information facets already
covered, generates entity-anchored gap-probing questions targeting
aspects relevant to $Q$ but absent from $T$, expands retrieval with those
questions, and uses an LLM judge to label each new candidate document as
novel, redundant, or off-topic.  The loop iterates until $T$
stops growing, yielding a structured inventory of
retrieval gaps $G \subseteq T$ together with label-free monitoring
signals (gap count, gap rate) whose stability
and sensitivity to retrieval quality we characterise empirically
(\S\ref{sec:results-sensitivity}, \S\ref{sec:results-variance}).
At $250$--$500$ LLM calls per audited query, Re:CAP is a
\emph{periodic sampling auditor} rather than a per-query monitor: it
runs over a representative sample on an audit cycle --- typically
before promoting an index, embedder, or ranker change --- not inline
on live traffic.  Cost is operator-tunable: an all-mini pipeline is
$4.6\times$ cheaper for $-0.51$\,pp paired recall, at the edge
of the run-to-run noise floor (App.~\ref{sec:appendix-ablations}, E3.5).

Our contributions are as follows:

\begin{itemize}\itemsep0pt\parsep0pt
    \item We propose a \textbf{topic-based gap-discovery
    framework} and instantiate it as the \textbf{Re:CAP
    iterative protocol}, whose gap-question generator combines
    \textbf{one dominant anchoring mechanism} --- the entity
    ledger, which accounts for nearly all of the average-recall
    lift --- with \textbf{four supporting guards}: probe-role
    diversity, coverage-aware topic serialisation,
    anti-collapse termination, and failure memory.  The four
    are not recall levers.  Each suppresses a specific failure
    mode of a minimal $(Q{+}T)$-only generator --- topic-coverage
    skew, low probe diversity, cold-start collapse, and wasted
    iterations --- that leave-one-out on \emph{average} recall
    over well-behaved benchmarks does not surface, and their
    effect is concentrated in worst-case behaviour
    (\S\ref{sec:method}, App.~\ref{sec:appendix-failure},
    Table~\ref{tab:lever-loo}).
    \item We provide \textbf{empirical evidence} across four
    public benchmarks that Re:CAP recovers $9$--$29\%$ of
    gold unreachable by flat BM25 top-$500$ ($+12.9$\,pp
    over flat hybrid top-$500$ on MuSiQue at less than half
    the document budget), moves predictably with retrieval
    quality (\S\ref{sec:results-sensitivity}), and reproduces
    to within $1\%$ across runs (\S\ref{sec:results-variance}).
    \item We \textbf{validate Re:CAP end-to-end on two
    unrelated live production deployments} over proprietary
    corpora ($400$ queries total;
    App.~\ref{sec:appendix-production}).
    \item We \textbf{validate Re:CAP-discovered gaps against
    blinded human judgement}: on the ensemble-missed stratum
    that no single-retriever upgrade closes, $78.9\%$ of
    recovered documents add information the baseline answer
    lacks (three blinded annotators, Fleiss $\kappa = 0.79$,
    $n = 123$; \S\ref{sec:probe-quality}).
\end{itemize}

\section{Method}
\label{sec:method}

\subsection{Setting and topic-based gaps}
\label{sec:method-setting}

A RAG pipeline has a corpus $\mathcal{C}$, retriever $R$, and LLM reader
$M$.  Given query $Q$, it produces an initial set of retrieved documents
$D_0 = R(Q, k)$ and an answer $A_0 = M(Q, D_0)$.  Without exhaustive
labels for $\mathcal{C}$ we audit the pipeline by discovering
\emph{retrieval gaps}: aspects of $Q$ that are relevant and
evidenced in $\mathcal{C}$ but absent from $D_0$.

We define gaps at the \emph{topic} level rather than the document
level.  What counts as a single topic is fixed by three rules: a \textbf{paragraph test} (each topic
warrants a distinct, non-overlapping paragraph), a
\textbf{type-not-instance} rule\footnote{A topic describes the
\emph{kind} of information sought, not a particular value of it
--- analogous to the class/instance distinction in object-oriented
design.}, and a hard cap (default $50$ topics per query).  A topic
$t$ is a \emph{retrieval gap} iff (i)~$t$ is relevant to $Q$,
(ii)~some $d \in \mathcal{C}$ covers $t$, and (iii)~no
$d' \in D_0$ covers $t$.  Topic-level counting provides built-in
deduplication across redundant evidence documents, interpretable
per-query reports, and a normalised gap rate
$|G|/|T_{\text{final}}|$ comparable across queries of varying
complexity.

\begin{algorithm}[t]
\caption{Re:CAP gap-discovery loop.
$\text{ColdStart}(D_0, A_0, i) := i\!=\!1 \wedge (D_0\!=\!\emptyset \vee A_0 \text{ is `insufficient context'})$ vetoes early termination (\S\ref{sec:method-loop}).}
\label{alg:recap}
\begin{algorithmic}[1]
\footnotesize
\Require query $Q$, retriever $R$, reader $M$, judge $\mathcal{J}$
\Require $k$, $m$, MAX\_ITER, MAX\_TOPICS
\State $D_0 \gets R(Q,k)$;\quad $A_0 \gets M(Q,D_0)$ \Comment{S1}
\State $T \gets \text{ExtractTopics}(Q, A_0)$ \Comment{S2}
\State $T \gets \text{ReconcileDocs}(Q, T, D_0)$ \Comment{S2}
\State $L \gets \text{ExtractEntityLedger}(Q, A_0)$ \Comment{S2}
\State $G \gets \emptyset$;\quad $F \gets \emptyset$;\quad $\text{seen} \gets \emptyset$
\For{$i = 1$ to MAX\_ITER}
    \State $\mathcal{Q}_g \gets \text{GenGapQs}(Q, T, L, F)$ \Comment{S3}
    \State $C \gets \bigcup_{q \in \mathcal{Q}_g} R(q, m) \setminus (D_0 \cup \text{seen})$ \Comment{S4}
    \State $\text{seen} \gets \text{seen} \cup C$
    \State $N \gets \emptyset$
    \ForAll{$c \in C$ \textbf{in parallel}} \Comment{S5}
        \State $v, \ell, t_{\text{id}} \gets \mathcal{J}(Q, T, A_0, c)$
        \If{$v \in \{\textsc{NewTopic}, \textsc{SubTopic}\}$}
            \State $N \gets N \cup \{(\ell, c)\}$
        \ElsIf{$v = \textsc{Redundant}$}
            \State $T[t_{\text{id}}].\text{evid} \mathrel{+{=}} \{c\}$
        \EndIf
    \EndFor
    \State $N \gets \text{Dedup}(N, T)$ \Comment{S5}
    \State $T \gets T \cup N$;\ \ $G \gets G \cup N$ \Comment{S5}
    \State $F \gets \mathcal{Q}_g \setminus \{q : q\text{ yielded a new topic}\}$ \Comment{S5}
    \If{$N = \emptyset \wedge \neg \text{ColdStart}(D_0, A_0, i)$}\ \textbf{break} \Comment{S6}
    \EndIf
    \If{$|T| \geq \text{MAX\_TOPICS}$}\ \textbf{break} \EndIf \Comment{S6}
\EndFor
\State \Return $T,\ G,\ \text{ComputeMetrics}(T, G)$
\end{algorithmic}
\end{algorithm}

\subsection{The Re:CAP loop}
\label{sec:method-loop}

Re:CAP runs as a six-step loop
(Algorithm~\ref{alg:recap}) layered on
top of an unchanged host RAG pipeline.

\begin{enumerate}\itemsep1pt\parsep0pt\topsep2pt\leftmargin1.4em
    \item \textbf{Initial pass (S1).} The host RAG pipeline runs
    unchanged: $D_0 = R(Q,k)$, $A_0 = M(Q, D_0)$.
    \item \textbf{Topic extraction and ledger init (S2).} An LLM
    extracts the distinct topics from $A_0$ as short labels; each
    $d \in D_0$ is then reconciled against the registry $T$ in a
    batched LLM call.  Reconciliation recovers topics from $D_0$ the
    reader omitted, so the rest of the loop measures only
    retriever-side gaps.  An \emph{entity ledger} $L$ (named
    entities, numeric and temporal anchors from $Q$ and $A_0$)
    is also extracted at this step and reused by S3 in every
    iteration.
    \item \textbf{Gap-Q generation (S3).} The generator takes
    $Q$, the registry $T$, and the entity ledger $L$, and emits
    five gap questions targeting aspects absent from $T$.  Each
    question is generated and tagged with one role from a fixed
    five-role taxonomy (\emph{entity-anchored},
    \emph{concept-anchored}, \emph{constraint-relaxed},
    \emph{constraint-tightened}, \emph{inverse-negation};
    definitions in App.~\ref{sec:appendix-probe-roles}).
    Orthogonal to these five roles, one anchoring
    \emph{mechanism} and four supporting \emph{guards} drive
    coverage and diversity:
    (A)~every question must contain a literal anchor
    from $L$ verbatim, preventing the generator from paraphrasing
    away bridge entities; (B)~topics are serialised with their
    evidence count so the generator targets under-supported
    topics; (C)~the role taxonomy enforces probe diversity;
    (D)~an \emph{anti-collapse} guard requires $\geq 2$
    iterations when $D_0$ is empty or $A_0$ is an
    ``insufficient context'' answer; (E)~\emph{failure memory}
    passes the previous iteration's unproductive gap-Qs back as
    negative examples.  As an ablation reference we additionally
    evaluate a minimal $(Q+T)$-only generator that drops
    A--E.
    \item \textbf{Expanded retrieval (S4).} For each gap question
    $g_i$, $R$ returns the top-$m$ documents, and
    $C = \bigcup_i R(g_i, m)$ is deduplicated against $D_0$ and
    previously seen candidate documents.  The same retriever as the host
    system is used, so discovered gaps reflect that retriever's
    coverage limitations.
    \item \textbf{Novelty judging (S5).} For each $c \in C$, the
    judge assigns one of five verdicts (\textsc{NewTopic},
    \textsc{SubTopic}, \textsc{Redundant}, \textsc{Irrelevant},
    \textsc{Contradictory}); the first two add the candidate to
    $T$ and to the gap inventory $G$, \textsc{Redundant} attaches
    the candidate as additional evidence to its existing topic,
    and the last two are discarded.  Unproductive gap-Qs (those
    yielding no new topic) are recorded in the failure memory $F$
    and passed back to S3 at the next iteration.  Overlapping new
    topics across parallel judges are resolved by two-stage
    post-hoc deduplication (App.~\ref{sec:appendix-dedup}).
    \item \textbf{Termination (S6).} The loop ends on natural
    convergence (no novel candidates), an anti-collapse veto on
    cold-start iter $1$, the topic cap, or the MAX\_ITER budget
    (Algorithm~\ref{alg:recap}).
\end{enumerate}

The loop returns (i) the topic registry $T$ with per-topic
evidence, (ii) the gap inventory $G \subseteq T$, and (iii)
monitoring metrics: gap count, gap rate, and recall when labels are available.

\section{Experimental Setup}
\label{sec:setup}

\subsection{Datasets}
\label{sec:setup-datasets}
We evaluate on four primary datasets spanning bounded-evidence multi-hop
QA and pooled-graded IR, plus MS MARCO TREC-DL 2019/2020 as an
auxiliary BM25-only sensitivity ladder (Table~\ref{tab:datasets}).
MuSiQue and HotPotQA are the primary bounded-evidence
regime; MultiHop-RAG, on a small 609-article corpus, serves as a
saturation/ceiling sanity check.  TREC-COVID is included to
delimit where Re:CAP applies and where it does not
(\S\ref{sec:limitations}).  Full per-dataset descriptions
and rationale are in Appendix~\ref{sec:appendix-datasets}.

\begin{table}[t]
\centering
\footnotesize
\setlength{\tabcolsep}{3pt}
\resizebox{\columnwidth}{!}{%
\begin{tabular}{lccccc}
\toprule
             & \textbf{MuSiQue} & \textbf{HotPot} & \textbf{MHopRAG} & \textbf{TC} & \textbf{MS MARCO} \\
\midrule
Corpus       & 21K     & 5.2M    & 609 articles & 171K          & 8.8M \\
Queries used & 100     & 98      & 98           & 50            & 97 \\
Evidence/q   & 2--4    & 2       & 2--4         & $\sim$1{,}327 & $\sim$213 \\
Judgements   & binary  & binary  & binary       & graded (0--2) & graded (0--3) \\
\bottomrule
\end{tabular}}
\caption{Dataset characteristics.  MHopRAG\,$=$\,MultiHop-RAG;
TC\,$=$\,TREC-COVID.  Evidence/q is gold-supporting-evidence count
for binary-judgement sets, mean judged-pool size for graded sets.
MS MARCO is BM25-only (corpus size precludes embedding); all others
use hybrid BM25\,+\,dense.}
\label{tab:datasets}
\end{table}

\subsection{Retrieval and models}
\label{sec:setup-models}
We evaluate Re:CAP with sparse BM25,
dense MiniLM-L12v2~\cite{reimers2019sbert,wang2020minilm}, and a
hybrid of the two via reciprocal-rank
fusion~\cite{cormack2009rrf}.  On MuSiQue we additionally evaluate
a stronger dense back-end, OpenAI
\texttt{text-embedding-3-large}.  We adopt hybrid as the reference
back-end; BM25-only and dense-only act as ablations.
The QA reader is held fixed at GPT-5.2 across all configurations to
remove reader quality as a confounding factor: the object of measurement
is \emph{retrieval} gaps, not reader capability.  The remaining Re:CAP
components (topic extractor, reconciler, gap-Q generator, judge)
default to GPT-4.1 with temperature $0$ (gap-Q generator temperature = $0.3$ for
diversity).  Unless noted otherwise, every main result uses hybrid
retrieval, the five-mechanism gap-Q generator, $k = 10$, $m = 50$,
five gap-Qs per iteration, MAX\_ITER $= 3$, and MAX\_TOPICS $= 50$.
The full model$\times$component table and a per-knob justification
are in Appendix~\ref{sec:appendix-defaults}.

\subsection{Baselines}
\label{sec:setup-baselines}
We compare Re:CAP against (i) \textbf{flat top-$N$ retrieval} at
budgets matched to Re:CAP's per-query unique-candidate count $N_q$
--- the central budget-matched control that isolates the audit's
contribution from raw exposure to more documents --- under three
retriever back-ends (BM25, MiniLM hybrid, and OpenAI dense
\texttt{text-embedding-3-large}; OpenAI on MuSiQue only);
(ii) \textbf{single-pass probing} (MAX\_ITER\,=\,1) to test whether
iteration adds value; (iii) \textbf{RM3 pseudo-relevance feedback
(PRF)}~\cite{lavrenko2001rm,abduljaleel2004umass}
(App.~\ref{sec:appendix-prf}); and (iv) the \textbf{minimal
$(Q+T)$-only} generator ablation on both BM25 and hybrid retrievers.

\paragraph{Re:CAP is an audit layer, not a competing retriever.}
These comparisons are budget-matched controls; none of them claims
Re:CAP is a better retrieval system.  Re:CAP runs \emph{on top of}
whatever retriever a deployment already operates
(\S\ref{sec:method}), so the evaluation question is not ``does the
loop beat a stronger retriever?'' but ``does the loop surface gold
that the host retrieval missed?''  Two standard upgrades do not
change that answer.  Cross-encoder \textbf{reranking} reorders the
flat top-$K$ already retrieved and cannot surface any document outside
it --- the axis the headline results are anchored on
(Figure~\ref{fig:blindspot}).  One-shot \textbf{query rewriting} is a
strict subset of what the loop does across iterations
(\S\ref{sec:method}).  Applying either to both sides raises the floor
symmetrically, which is why we report unreachable-gold share alongside
recall throughout.

\subsection{Evaluation protocol}
\label{sec:setup-protocol}
For each Re:CAP run we record per-query (i) gold recall against
dataset qrels (for validation only, Re:CAP itself does not consume
labels), (ii) Re:CAP gap count and gap rate, (iii)
unreachable-gold share vs.\ flat top-$500$ (the fraction of
Re:CAP-recovered gold not in flat top-$500$ under each retriever), and
(iv) telemetry rolled up to dollar cost using published Azure OpenAI rates.
Confidence intervals on run-level metrics are $1{,}000$-sample
query-bootstrap $95\%$, and paired differences use the same bootstrap on
per-query deltas; the human evaluations state their interval method with
their tables.  The
variance protocol re-runs Re:CAP with the default settings three times under the same
query slice and seed to estimate end-to-end stability under inherent
LLM non-determinism.

\section{Results}
\label{sec:results}

\subsection{Re:CAP vs.\ flat retrieval}
\label{sec:results-headline}
Table~\ref{tab:headline} reports the cross-dataset main result
(Pareto visualization in App.~Figure~\ref{fig:recall-pareto}):
Re:CAP
at the default configuration vs.\ three flat baselines at matched
docs-seen budget.  On MuSiQue and HotPotQA, Re:CAP
beats flat BM25 at matched-$N_q$ budget by
$+6.1$ to $+29.1$\,pp; on MultiHop-RAG, where flat top-$500$ is
near saturation, the matched-$N_q$ gain narrows to $+2.6$\,pp.
It also beats stronger flat
baselines at less than half the document budget: on MuSiQue,
$+12.9$\,pp vs.\ flat MiniLM hybrid top-$500$ and $+4.7$\,pp vs.\
flat OpenAI dense top-$500$; on HotPotQA, $+11.5$\,pp vs.\ flat
MiniLM dense top-$500$.  Across the six comparisons in
Table~\ref{tab:headline} the CIs are disjoint on the three largest
margins ($+11.5$ to $+29.1$\,pp) and overlap on the three smallest
($+2.4$ to $+6.1$\,pp): on the recall axis a stronger baseline
narrows the margin, which is why the unreachable-gold share
(\S\ref{sec:results-complementarity}) rather than recall is the
retriever-independent signal.  On
MultiHop-RAG the 609-article corpus is near-saturated by flat
top-$500$ across retriever back-ends ($0.88$--$1.00$), so it serves
as a ceiling sanity check rather than a discriminative benchmark;
Re:CAP converges within $N_q \approx 45$--$90$ depending on back-end.
On TREC-COVID Re:CAP shows the strongest structural complementarity:
$48\%$ of its recovered gold is absent from flat BM25 top-$500$ and
\textbf{$21.2\%$ remains absent from the ensemble of flat BM25,
dense, and hybrid top-$500$ combined} -- a $1{,}500$-document,
three-retriever budget -- appearing in $98\%$ of queries (49 of 50).
By majority vote of three blinded annotators, $78.9\%$
[$71.5$--$86.2\%$] of $n = 123$ such documents add information beyond
the $D_0$-only baseline answer (Fleiss $\kappa = 0.79$; $78.3\%$ on
MuSiQue; \S\ref{sec:probe-quality}).  On the recall
metric Re:CAP trails flat BM25 top-$500$ by a marginal $-2.0$\,pp
($0.221$ vs.\ $0.241$) at $\sim$$24\%$ smaller document budget:
flat top-$500$ itself reaches only $24\%$ on
this pooled-graded gold set of $\sim$$493$ relevant docs per
query, and binary-novelty judging at a $\sim$$378$-doc Re:CAP budget is
structurally mismatched with the recall ceiling
(\S\ref{sec:limitations}), thus 
complementarity, not recall@$k$, is the audit-relevant signal here.

\begin{table}[t]
\centering
\scriptsize
\setlength{\tabcolsep}{2.5pt}
\begin{tabular}{llrrrr}
\toprule
Dataset & Method & Docs & Recall [95\% CI] & Unr. & $|G|$ \\
\midrule
\multirow{4}{*}{\textbf{MuSiQue}}
 & Flat BM25 $N_q$            & 238 & 0.610 [.558,.664] & --- & --- \\
 & Flat hyb.\ top-500         & 500 & 0.772 [.723,.818] & --- & --- \\
 & Flat OpenAI top-500        & 500 & 0.854 [.815,.896] & --- & --- \\
 & \textbf{RR hyb.}           & \textbf{238} & \textbf{0.901 [.854,.938]} & \textbf{29.4\%} & \textbf{3.6} \\
\midrule
\multirow{4}{*}{\textbf{HotPot}}
 & Flat BM25 $N_q$            & 249 & 0.827 [.775,.878] & --- & --- \\
 & Flat hyb.\ top-500         & 500 & 0.864 [.813,.909] & --- & --- \\
 & Flat dense top-500$^{\ddagger}$ & 500 & 0.773 [.717,.828] & --- & --- \\
 & \textbf{RR hyb.}           & \textbf{249} & \textbf{0.888 [.837,.939]} & \textbf{9.2\%} & \textbf{2.3} \\
\midrule
\multirow{2}{*}{\textbf{TC}$^{\dagger}$}
 & Flat BM25 top-500          & 500 & 0.241 [.211,.271] & --- & --- \\
 & \textbf{RR hyb.}           & \textbf{378} & \textbf{0.221 [.195,.248]} & \textbf{48.0\%}$^{\S}$ & \textbf{36.9} \\
\bottomrule
\end{tabular}
\caption{Re:CAP vs.\ flat retrieval at matched docs-seen budget.
\emph{RR hyb.}\,=\,Re:CAP with hybrid retrieval;
\emph{Unr.}\,=\,share of recovered gold not in flat BM25 top-500
(see $^{\S}$ for the ensemble variant);
$|G|$\,=\,mean Re:CAP gap count per query.
MultiHop-RAG omitted (saturates at flat top-$500$).
$^{\dagger}$pooled-graded scope boundary;
$^{\ddagger}$MiniLM-L12 dense;
$^{\S}$$21.2\%$ also unreachable by the ensemble of flat BM25,
dense, and hybrid top-$500$.  Full matrix:
App.~\ref{sec:appendix-headline-full}.}
\label{tab:headline}
\end{table}
\subsection{Structural complementarity}
\label{sec:results-complementarity}

Re:CAP samples a meaningfully different slice of the relevance pool
than flat retrieval --- an audit signal flat top-$k$ cannot produce
by construction.  Against flat BM25 top-$500$, $9.2$--$29.4\%$ of
Re:CAP's recovered gold is unreachable on bounded-evidence (column
\emph{Unr.}\ of Table~\ref{tab:headline}), rising to $48\%$ on
TREC-COVID under the default hybrid back-end ($45$--$61\%$ across
alternative Re:CAP retriever back-ends; App.~\ref{sec:appendix-headline-full}).
Against the ensemble of BM25, dense, and hybrid top-$500$, the share
tracks gold-pool
diversity: $2.5\%$ on HotPotQA ($4\%$ of queries; 2-doc pools),
$10.0\%$ on MuSiQue ($21\%$ of queries; 2--4 hops), and $21.2\%$
on TREC-COVID ($98\%$ of queries; $\sim$$493$ gold docs/q;
Table~\ref{tab:unreachable}).  Re:CAP's structural contribution is
largest exactly where exhaustive recall labels are most intractable
to obtain --- the production setting the method is designed for.

\subsection{Sensitivity to retrieval quality}
\label{sec:results-sensitivity}

A monitoring metric is only useful if it \emph{moves} when the
underlying retrieval changes --- whether from a corpus refresh, an
embedder upgrade, or a reranker change.  Table~\ref{tab:ladder} reports a
within-dataset retrieval-quality ladder: on MuSiQue, sweeping
retriever type (BM25 / hybrid) and top-$k$ ($10$ / $20$ / $50$) with
generator fixed; on MS MARCO TREC-DL 2019/2020 ($n=97$ NIST-judged
queries, $\text{rel} \geq 2$), sweeping BM25 top-$k$ over the same
ladder.  Two observations.  (i) Retriever \emph{type} dominates where
multiple are available: on MuSiQue, BM25 vs.\ hybrid moves Re:CAP
recall by $\sim$$6$\,pp at matched $k$.  (ii) Top-$k$ within a
single retriever is approximately flat because Re:CAP's expansion
step saturates the candidate pool regardless of $D_0$ depth
($\leq 1.9$\,pp paired across top-$10$/$20$/$50$ on every ladder;
$\leq 0.75$\,pp on MS MARCO BM25).
Together these are consistent with the design intent: Re:CAP metrics are
\emph{discriminative} where retrieval truly differs and \emph{stable}
where it does not.

The two label-free monitoring signals behave differently along this
ladder, and the distinction matters for anyone deploying them.  Mean
gap count $|G|$ separates the two retriever types in the expected
direction --- every BM25 rung sits above every hybrid rung
($4.00$--$4.20$ vs.\ $3.61$--$3.97$) --- so a weaker host retriever
leaves Re:CAP more gaps to open.  The magnitude, however, should be
read with care: the BM25--hybrid mean separation is $0.28$, only
$1.5\times$ the run-to-run standard deviation of $|G|$ ($0.19$,
Table~\ref{tab:variance}), and at matched $k{=}20$ the two retrievers
differ by $0.03$.  The type separation is consistent on this ladder,
but within a retriever top-$k$ does not order $|G|$ reliably, and the
per-rung differences are not resolvable within a single audit cycle.  $|G|$ is therefore a directional indicator here rather than a
calibrated one, and separating adjacent configurations needs repeated
cycles or a larger degradation than this ladder spans.

Gap rate is the more stable signal (CV $1.46\%$ against $5.04\%$) but
is near saturation here ($0.81$--$0.86$ on MuSiQue, $0.96$--$0.98$ on
MS MARCO): on MS MARCO almost every query yields a gap, so it acts as
a coverage \emph{floor} rather than a fine-grained sensitivity
signal.  Neither signal
is a drop-in alarm on its own; the threshold recipe is in
App.~\ref{sec:appendix-deployment}.

\begin{table}[t]
\centering
\footnotesize
\setlength{\tabcolsep}{4pt}
\resizebox{\columnwidth}{!}{%
\begin{tabular}{lrrrrrr}
\toprule
Retriever & recall & $\Delta_{\text{paired}}$ & $|G|$ & gap rate & \$/q & $\overline{\text{it}}$ \\
\midrule
\multicolumn{7}{l}{\emph{MuSiQue ($n=100$, hybrid top-10 ref.)}} \\
BM25 top-10 & 0.842 & $+5.92$ & 4.20 & 0.831 & 0.65 & 1.93 \\
BM25 top-20 & 0.861 & $+4.00$ & 4.00 & 0.859 & 0.65 & 1.88 \\
BM25 top-50 & 0.857 & $+4.42$ & 4.15 & 0.825 & 0.61 & 1.92 \\
\textbf{Hybrid top-10 (ref)} & \textbf{0.901} & --- & \textbf{3.61} & \textbf{0.845} & \textbf{0.59} & \textbf{1.79} \\
Hybrid top-20 & 0.918 & $-1.67$ & 3.97 & 0.814 & 0.62 & 1.88 \\
Hybrid top-50 & 0.915 & $-1.01$ & 3.93 & 0.831 & 0.57 & 1.77 \\
\midrule
\multicolumn{7}{l}{\emph{MS MARCO TREC-DL ($n=97$, BM25 top-10 ref.)}} \\
\textbf{BM25 top-10 (ref)} & \textbf{0.683} & --- & \textbf{12.58} & \textbf{0.976} & \textbf{0.94} & \textbf{2.43} \\
BM25 top-20 & 0.679 & $+0.42$ & 12.32 & 0.955 & 0.96 & 2.51 \\
BM25 top-50 & 0.676 & $+0.75$ & 12.03 & 0.956 & 0.87 & 2.34 \\
\bottomrule
\end{tabular}}
\caption{Retrieval-quality ladder.  Columns: Re:CAP recall; paired delta (positive $\Delta_{\text{paired}}$ implies the cell is \emph{worse} than its reference); mean gap count $|G|$ per query; gap rate; cost per query; and mean iterations per query till convergence.  $|G|$ separates the two retriever types in the expected direction, though by margins comparable to its own run-to-run noise; gap rate saturates on these workloads and acts as a coverage floor (\S\ref{sec:results-sensitivity}).}
\label{tab:ladder}
\end{table}

\subsection{Controlled gold-deletion check}
\label{sec:results-deletion}

We additionally validate sensitivity under controlled deletion:
removing $K \in \{1, 2, \text{all}\}$ gold passages from $D_0$
($n=91$ MuSiQue queries with gold in $D_0$), Re:CAP re-discovers
$98.9\%/98.4\%/100\%$ of the removed gold IDs respectively.  Paired
recall still falls $11$--$13$\,pp, attributable to secondary gold
rather than to failed recovery.  Full E2.2 table and discussion:
Appendix~\ref{sec:appendix-e22}.

\subsection{Operational characteristics}
\label{sec:results-ops}

\paragraph{Cost.}\label{sec:results-cost}
The judge dominates per-query cost ($\sim$$97\%$ across all datasets;
judge $\to$ GPT-4.1-mini saves $75\%$ of HotPotQA cost, the generator
swap alone saves $8\%$).  Per-query cost varies $\sim$$3.5\times$ across
corpora (\$$0.59$--\$$2.05$, Table~\ref{tab:cost}).  A $4.6\times$
Pareto improvement is available for $-0.51$\,pp paired recall on
HotPotQA by swapping all pipeline components to GPT-4.1-mini (at the
edge of the $0.42$\,pp noise floor below; full grid in
Appendix~\ref{sec:appendix-ablations}).

\begin{table}[t]
\centering
\footnotesize
\setlength{\tabcolsep}{2.5pt}
\begin{tabular}{lrrrr}
\toprule
Metric & MuSiQue & HotPot & MHopRAG & TC \\
\midrule
LLM calls (total)   & 28.6k & 25.0k & 25.8k & 30.8k \\
\quad of which judge& 28.1k & 24.5k & 25.3k & 30.2k \\
Judge share         & 97.5\% & 97\%  & 97\%  & 97\% \\
\textbf{\$ / query} & \textbf{0.59} & \textbf{0.65} & \textbf{0.69} & \textbf{2.05}$^{\ddagger}$ \\
\bottomrule
\end{tabular}
\caption{Cost breakdown (Re:CAP default on hybrid retrieval).
$^{\ddagger}$TREC-COVID cell is the BM25\,+\,$(Q{+}T)$ variant
(only TC configuration with full token telemetry).}
\label{tab:cost}
\end{table}

\paragraph{Reproducibility.}\label{sec:results-variance}
Three independent runs of Re:CAP on MuSiQue
(Table~\ref{tab:variance}; same slice and seed; variance is inherent
LLM non-determinism) give Re:CAP-recall a Coefficient-of-variation  $= 0.47\%$
($\pm 0.42$\,pp), so Re:CAP returns a stable recall estimate for a
fixed (pipeline, corpus) pair.  Combined with the retriever-type sensitivity demonstrated in
\S\ref{sec:results-sensitivity}, this licenses inter-run comparisons
at that granularity: recall deltas that exceed the noise floor and
accompany a change of retriever can be read as real changes rather
than measurement noise.  Adjacent top-$k$ settings are not separable
this way.

\begin{table}[t]
\centering
\footnotesize
\setlength{\tabcolsep}{4pt}
\begin{tabular}{lrrrr}
\toprule
Metric & Mean & Std & CV & Range \\
\midrule
\textbf{Re:CAP recall}    & \textbf{0.8972} & 0.0043 & \textbf{0.47\%} & 0.0083 \\
Re:CAP gap count $|G|$    & 3.77           & 0.19   & 5.04\%          & 0.37 \\
\textbf{Re:CAP gap rate}  & \textbf{0.832} & 0.012  & \textbf{1.46\%} & 0.024 \\
$N_q$ (unique cands.)     & 246.5 & 7.4    & 2.99\%          & 15 \\
Cost / run (\$)           & 62.47 & 3.18   & 5.09\%          & 6.01 \\
\bottomrule
\end{tabular}
\caption{Run-to-run variance, three independent runs
(MuSiQue, default, $n=100$). Recall CV $0.47\%$ and gap-rate CV
$1.46\%$ both underpin the deployable-monitoring claim.}
\label{tab:variance}
\end{table}

\paragraph{Attribution of residual losses.}
On bounded-evidence primaries the dominant failure is gap-question
generation, not judging ($95.3\%$ of missing docs never surface as
candidates): early convergence, bridge-entity erasure, or entity-ledger
anchoring on a hallucinated name from $A_0$.  On pooled-graded
TREC-COVID the failure is structural: binary novelty judging
over-aggregates sub-mechanisms and the Re:CAP budget is dwarfed by
the graded pool.  Full taxonomy, counts, and worked examples are in
Appendix~\ref{sec:appendix-failure}.

\paragraph{Production deployment.}\label{sec:results-production}
The default configuration also audits two live
deployments on proprietary corpora ($\approx$$123$M and
$\approx$$70$M passages; $200$ queries each), differing
chiefly in $D_0$ width: ``AI Search'' ($D_0$ fixed at $10$)
and ``NEWS QA'' ($D_0$ mean $47.2$, max $110$).  Both run
$200/200$ with no production-specific code path: gap rate
$0.991$ / $0.955$, gap-iteration topic share $77.4\%$ / $78.2\%$,
judge share of calls $\geq 97.6\%$, median cost
\$$1.68$ / \$$1.79$ per query.  Full two-cohort breakdown in
App.~\ref{sec:appendix-production} (Table~\ref{tab:production-cohorts}).

\section{Human evaluation of recovered gap documents}
\label{sec:probe-quality}

The $21.2\%$ TREC-COVID and $10.0\%$ MuSiQue ensemble-unreachable
shares in \S\ref{sec:results-headline} are qrels-mediated.  To
test whether these structurally distinct documents add information
the reader actually lacks --- independent of the qrels --- we ran
a blinded human read on the ensemble-unreachable stratum.

\paragraph{Setup.}
We sampled $n = 123$ ensemble-unreachable gap documents from the
deployed default: $100$ TREC-COVID documents stratified by query
(target $2$ docs per query among the TREC-COVID queries with
$\geq 1$ ensemble-unreachable doc; $48$ queries represented,
$1$--$3$ docs each) plus the $23$-document
MuSiQue census; HotPotQA, MultiHop-RAG, and MS MARCO are
ineligible (App.~\ref{sec:appendix-probe-quality}).  Three
annotators saw the query, the $D_0$-only baseline
answer, and the document text, blinded to judge verdict, qrels
status, and dataset.  The binary rubric labels a document
\emph{new\_info} if it adds a substantive query-relevant fact
the baseline answer lacks, and \emph{covered} otherwise.

\paragraph{Result.}
By majority vote, $78.9\%$ [$71.5$--$86.2$, $10{,}000$-resample
bootstrap] of ensemble-missed gap documents are judged
\emph{new\_info} at Fleiss $\kappa = 0.79$
(Table~\ref{tab:probe-quality}); per-dataset rates are
substantively comparable.  This bounds Re:CAP's operational value on
the strictest stratum: the documents it recovers that no
$1{,}500$-document single-retriever combination surfaces carry
information the baseline answer lacks.

\begin{table}[h]
\centering
\scriptsize
\setlength{\tabcolsep}{3pt}
\begin{tabular}{lrrrc}
\toprule
Slice & $n$ & new & Rate \% [95\% CI] & $\kappa$ \\
\midrule
TREC-COVID (stratified) & 100 & 79  & $79.0$ [$71.0$--$87.0$] & $0.82$ \\
MuSiQue (census)        &  23 & 18  & $78.3$ [$60.9$--$95.7$] & $0.69$ \\
\textbf{Overall}        & \textbf{123} & \textbf{97} & \textbf{$78.9$ [$71.5$--$86.2$]} & \textbf{$0.79$} \\
\bottomrule
\end{tabular}
\caption{Human-evaluation verdicts on $123$ ensemble-unreachable
gap documents from the deployed default.  \emph{new} is the
majority-vote \emph{new\_info} count; $95\%$ CIs are
$10{,}000$-resample bootstrap; $\kappa$ is Fleiss across three
annotators on binary verdicts.  $110/123$ ($89.4\%$) unanimous.}
\label{tab:probe-quality}
\end{table}

\section{Conclusion}
\label{sec:conclusion}
Re:CAP reformulates the retrieval evaluation problem in Production
RAG pipelines from that of \emph{enumeration} to one of
\emph{probing} by utilizing a topic-aware iterative gap-discovery
loop that audits a deployed pipeline without per-query gold
labels.  Across four publicly available benchmarks we demonstrate
that Re:CAP recovers gold that baselines cannot discover
($9$--$29\%$, rising to $48\%$ on TREC-COVID) and does so at half
the document budget (on MuSiQue).  Results are reproducible to
$\sim$$1\%$ recall across runs.  Human assessment on Re:CAP
recovered gold documents shows that $78.9\%$ of the recovered
documents have novel information missing from the baseline answer
($\kappa = 0.79$).  We share Re:CAP statistics and examples from
2 live large-scale production RAG pipelines with $200$-query
audits each (App.~\ref{sec:appendix-production}) along with a deployment recipe
(App.~\ref{sec:appendix-deployment}).

\section*{Ethical Considerations}
\label{sec:ethics}

\paragraph{Computational and environmental cost.}
The iterative probing loop has significant energy and carbon
implications at scale (\$$0.59$--\$$2.05$/query at default;
\S\ref{sec:results-cost}); we report full per-corpus cost accounting
to enable informed deployment decisions and document a steep
Pareto improvement (judge $\to$ GPT-4.1-mini) for cost-sensitive
deployments.  Teams running Re:CAP continuously should sample
queries rather than audit every request.

\paragraph{Bias in gap discovery.}
The LLM-as-judge may have systematic blind spots --- topics it
consistently fails to recognise as novel --- that could correlate with
sensitive attributes and provide false assurance of completeness.
Re:CAP cannot fully audit itself; periodic gold-label spot-checks of
judge outputs against held-out qrels
(Appendix~\ref{sec:appendix-component}) are recommended before relying
on Re:CAP signals for compliance-grade reporting.  Re:CAP's gap-Q
generator also inherits any entity biases of the source LLM
through the entity ledger; we have not characterised this with respect to
demographic or geographic biases.

\paragraph{Human annotation.}
The human evaluations in \S\ref{sec:probe-quality} and
App.~\ref{sec:appendix-humanread-prod} were performed by three
in-house annotators, not by the authors and not by
crowdworkers.  The
production cohorts are internal corpora: annotation took place
inside that organisation, and no document text from them is
reproduced here --- the worked examples in
App.~\ref{sec:appendix-production} report topic labels and
generated probe questions only.  We collected no personal data
from annotators and report no demographic characteristics of the
annotator population.

\paragraph{Privacy and access control.}
Expanded retrieval issues additional probes against the corpus with
LLM-generated questions.  In deployed systems with row-level access
controls, those controls must be honoured during gap probing ---
otherwise the audit can surface relevant documents the original user
would not be entitled to retrieve.  Our reference
implementation passes the original retrieval ACL context through the
loop; we recommend the same for any deployment.

\paragraph{Surface area for prompt injection.}
Step 2 (reconciliation) and Step 5 (judging) ingest retrieved
document text as input to LLM calls and are therefore exposed to
prompt-injection attempts embedded in corpus documents.  We use
structured-output decoding and a narrowly scoped classification task
to reduce this surface, but for adversarial corpora additional
input sanitisation is warranted.

\section*{Limitations}
\label{sec:limitations}

\paragraph{LLM cost.} Re:CAP requires multiple LLM calls per
query (topic extraction, doc reconciliation, gap-Q generation,
novelty judging per candidate, post-hoc deduplication). At the
default this totals $250$--$500$ LLM calls per query
(\$$0.59$--\$$2.05$ at GPT-4.1 pricing; \S\ref{sec:results-cost}).
This is materially more expensive than single-pass metrics like
RAGAS, and may not suit tight monitoring budgets.  Swapping
all pipeline components to GPT-4.1-mini cuts this $4.6\times$
($-0.51$\,pp paired recall on HotPotQA;
App.~\ref{sec:appendix-ablations}, E3.5).

\paragraph{Judge reliability ceiling.} Re:CAP is in principle
bounded by the LLM judge's accuracy. The gold-label failure-mode
analysis (Appendix~\ref{sec:appendix-failure}) bounds this
empirically: $1.4\%$ of missing-gold docs are judge-rejected
overall ($4.7\%$ on the bounded-evidence primaries), so $> 95\%$ of
missing-gold losses on the regime Re:CAP is designed for are
upstream of judging.  We rely on the LLM-as-judge literature
\cite{faggioli2023perspectives,thomas2024large,upadhyay2024llms,
upadhyay2024largescale} as support for narrow,
structured-output tasks, and follow cautions against substituting
LLMs for full human qrels~\cite{soboroff2025dont} by validating
Re:CAP against held-out labels.

\paragraph{Corpus coverage assumption.} Re:CAP discovers gaps
only for topics that exist in the corpus.  If the corpus itself
lacks coverage of an aspect, no probing will find it.  Re:CAP
measures \emph{retrieval} gaps, not \emph{corpus} gaps.

\paragraph{English only.} All experiments are on
English-language datasets with English-language LLMs. The
approach is language-agnostic in principle but unvalidated
multilingually.

\paragraph{Gap-Q generator dependence.} Gap-probing depends on
the LLM-generated questions; vague or off-target questions
under-surface relevant documents and under-count gaps. The
five-mechanism generator reduces but does not eliminate this risk.

\paragraph{Topic granularity.} We have no formal definition of
an ``atomic topic''. We rely on three prompt-engineering rules
(paragraph test, type-not-instance, different-dimension sub-topic)
plus a configurable hard cap. Broad or ambiguous queries can
still trigger instance enumeration, inflating topic counts.

\paragraph{Instance-level coverage holes can be masked.}
By design, the type-not-instance rule collapses instance
enumeration into a single topic.  A retriever that surfaces
\emph{some} instances of a topic but misses others along the
same dimension is scored as covering that topic, even though
instance-level recall is incomplete.  The \textsc{SubTopic}
verdict mitigates this only when missed instances reveal a
different \emph{dimension}, not when they are missing along the
existing one.  Deployments that care about instance-level
enumeration recall (e.g., fact verification, audit trails) should
track Re:CAP at a finer granularity, or pair it with
instance-level metrics.

\paragraph{Recall@$k$ undercounts complementarity on
pooled-graded corpora.}
On pooled-graded corpora such as TREC-COVID --- where a query
has hundreds--thousands of partially-relevant documents, flat
BM25 top-$500$ itself reaches only $24\%$ recall, and any
$\sim$$378$-doc method cannot hold the $\sim$$493$ relevant
docs per query --- recall@$k$
compresses complementarity into a small negative delta
(Re:CAP best cell $\Delta = -2.0$\,pp;
\S\ref{sec:results-headline}).  The audit-relevant signal
survives: $21.2\%$ of Re:CAP's recovered gold remains absent
from the ensemble of flat BM25, dense, and hybrid top-$500$
combined, present in $98\%$ of queries
(Table~\ref{tab:unreachable}).  On such corpora,
recall@$k$ should be paired with a structural complementarity
metric to capture Re:CAP's audit contribution.

\paragraph{Initial-retriever choice can cause cell-level
regressions.} The HotPotQA dense-$D_0$ cell ($-3.6$\,pp
matched-$N_q$) shows the initial retriever interacts with the
dataset's evidence structure.  The initial retriever should be
selected based on dataset-level recall-at-$k$ diagnostics before
Re:CAP is layered on top.

\paragraph{Provider-side content filtering depresses
recall.} All experiments run against Azure OpenAI, whose
content-management policy rejects a small share of
prompts as \texttt{ResponsibleAIPolicyViolation}.  Rejected
candidates produce no verdict, so any that would have been
\textsc{NewTopic} are silently dropped and reported recall is a
conservative lower bound.  The effect is small per-query and
well within the variance noise floor of
\S\ref{sec:results-variance}, but a non-filtered backend would
likely yield slightly higher recall than reported here.

\paragraph{Calibrated metrics rest on benchmarks; production
audit is a single snapshot.}  The recall, paired-$\Delta$,
unreachable-gold, variance, and cost ladders all rest on four
English-language academic benchmarks
(\S\S\ref{sec:results-headline}--\ref{sec:results-variance});
those metrics require per-query gold and so cannot be reported
on production traffic.  We do also report end-to-end audits on
two unrelated live deployments over proprietary corpora ($400$
queries total, App.~\ref{sec:appendix-production}), which
confirm the gap rate, topic-source distribution, convergence
behaviour, and judge-share-of-cost pattern transfer to
production traffic across deployments differing in $D_0$ width
by $\sim$$5\times$.

\section*{Acknowledgments}

We thank the Annotation Center of Excellence (ACoE) at JPMorgan Chase \& Co.\ for the human annotation work reported in this paper. The annotators are salaried employees of JPMorgan Chase \& Co.\ and performed this work as part of their regular duties.

\section*{Disclaimer}

This paper was prepared for informational purposes in part by the Machine Learning Center of Excellence group of JPMorgan Chase \& Co.\ and its affiliates (``JP Morgan'') and is not a product of the Research Department of JP Morgan. JP Morgan makes no representation and warranty whatsoever and disclaims all liability, for the completeness, accuracy or reliability of the information contained herein. This document is not intended as investment research or investment advice, or a recommendation, offer or solicitation for the purchase or sale of any security, financial instrument, financial product or service, or to be used in any way for evaluating the merits of participating in any transaction, and shall not constitute a solicitation under any jurisdiction or to any person, if such solicitation under such jurisdiction or to such person would be unlawful.

\appendix

\section*{Appendix}

\section{Datasets and Rationale}
\label{sec:appendix-datasets}

We evaluate on four datasets, chosen to span (i) bounded-evidence
multi-hop QA --- the operational regime Re:CAP is designed for ---
and (ii) pooled-graded IR as a contrasting regime
(Table~\ref{tab:datasets} in the main body).  The MS MARCO TREC-DL
2019/2020 ladder is an additional BM25-only validation surface for
the sensitivity result (\S\ref{sec:results-sensitivity}).

\textbf{MuSiQue}~\cite{trivedi2022musique}: 21K-paragraph deduped
corpus, 2--4-hop chained reasoning, shortcut-resistant by design;
$2{,}417$ dev queries, of which we sample $100$ with seed $42$. The
strongest signal in our sensitivity ladder.

\textbf{HotPotQA}~\cite{yang2018hotpotqa}: 5.2M-passage Wikipedia
corpus, canonical multi-hop benchmark; $5{,}447$ dev / $7{,}405$ test
queries, of which we sample $98$ with seed $42$. Provides the
largest-corpus comparison point.

\textbf{MultiHop-RAG}~\cite{tang2024multihoprag}: 609-article news
corpus, $2{,}556$ test queries, 2--4 documents per query; an
EMNLP 2024 RAG-native benchmark.  We use the published test set
($n=98$ for our slice).  The small corpus means flat BM25 top-$500$
reaches recall $1.000$, so this dataset is best read
as a saturation/ceiling sanity check rather than a discrimination
result.

\textbf{TREC-COVID} (BEIR)~\cite{voorhees2021treccovid}: 171K biomedical
passages, 50 round-3 topics with pooled graded judgements
($\sim$$493$ rel docs per query at $\text{rel} \geq 1$; $\sim$$1{,}327$ judged).  We
include it to characterise where Re:CAP's binary-novelty
machinery breaks down (see Limitations, \emph{Recall@$k$
undercounts complementarity on pooled-graded corpora}).

\textbf{MS MARCO} TREC-DL 2019/2020~\cite{bajaj2018msmarco,craswell2020trecdl}:
8.8M passages, 97 NIST-judged queries with graded qrels
thresholded at $\text{rel} \geq 2$.  We use this as a BM25-only
sensitivity ladder (Table~\ref{tab:ladder}, second panel) on an
independently judged, IR-standard corpus.  We deliberately do not
embed MS MARCO for dense or hybrid retrieval (project-budget
choice); it is used only in BM25-only sensitivity configurations.

\section{Related Work}
\label{sec:appendix-related}
\label{sec:related}

\paragraph{Retrieval evaluation and coverage.}
Recall, precision, nDCG, and MAP require relevance judgements.  TREC
pooling~\cite{voorhees2000variations} aggregates top-ranked documents
across systems and judges only the pool, but is biased toward in-pool
systems, treats unjudged documents as irrelevant, and does not scale
to multi-million-passage corpora that re-index
frequently~\cite{zobel1998reliable,buckley2007bias}.  Novelty,
diversity, and nugget-based test collections move the unit of
evaluation from documents toward subtopics or information
nuggets~\cite{clarke2008novelty,voorhees2003trec,pavlu2012nugget}.
Re:CAP adopts this topic-level unit, but does not build reusable
qrels (query relevance judgements) or estimate absolute recall: it
probes a single deployed pipeline for positive evidence of missed
topics.

\paragraph{RAG evaluation frameworks.}
RAGAS~\cite{es2023ragas}, ARES~\cite{saadfalcon2024ares},
RGB~\cite{chen2024benchmarking}, and
RAGChecker~\cite{ru2024ragchecker} score supplied retrieval and
generation for faithfulness, answer relevance, context
relevance/recall, and module-level diagnostics.  They are
complementary to Re:CAP: they ask whether the answer is supported by
the retrieved context; Re:CAP asks whether relevant facets exist
\emph{outside} that context and returns an actionable gap inventory.

\paragraph{Coverage-oriented RAG evaluation.}
Closest to Re:CAP are recent coverage-oriented RAG evaluations.  Xie
et~al.~\cite{xie2025subquestion} decompose open-ended questions into
core, background, and follow-up sub-questions; CRUX evaluates whether
retrieved contexts cover human-grounded information needed for
long-form generation~\cite{ju2025crux}; CoverageBench assembles
coverage-oriented test collections~\cite{samuel2026coveragebench};
and LANCER optimises reranking for nugget coverage~\cite{ju2026lancer}.
These works assume pre-specified sub-questions, summaries, nuggets, or
coverage-aware training/evaluation targets.  Re:CAP instead induces a
topic registry from $D_0$ and $A_0$, then actively probes for missing
topics without per-query qrels or pre-authored facets.

\paragraph{LLM-as-judge and iterative retrieval.}
LLM relevance judging shows mixed results: studies report
strong agreement with preferences or TREC-style assessments
\cite{faggioli2023perspectives,thomas2024large,upadhyay2024llms,
upadhyay2024largescale}, while others warn against replacing human
qrels with LLM labels~\cite{wang2023fair,soboroff2025dont}.  Re:CAP's
judge performs novelty classification against a per-query
topic registry, not absolute relevance scoring; we validate it
against held-out qrels ($1.4\%$ judge-rejection on missing-gold
candidates).
Self-RAG~\cite{asai2023selfrag}, FLARE~\cite{jiang2023flare}, and
IRCoT~\cite{trivedi2023ircot} use iterative retrieval to
\emph{improve generation}; Re:CAP repurposes related mechanics for
\emph{evaluation}, auditing a fixed pipeline after the fact.
Pseudo-relevance feedback~\cite{lavrenko2001rm,abduljaleel2004umass}
and LLM query expansion~\cite{nogueira2019doc2query} provide our
baseline references.  Table~\ref{tab:positioning} below summarises
the positioning.

\subsection{Positioning against closest prior work}
\label{sec:appendix-positioning}

Table~\ref{tab:positioning} compares Re:CAP to four families of prior
work along five capabilities.

\paragraph{Reading the columns.}
\emph{RAG metrics} (RAGAS, ARES, RGB, RAGChecker) score retrieval
indirectly through answer faithfulness and answer/context relevance,
and typically require reference answers or per-query reference
contexts. \emph{Coverage evaluation} (sub-question decomposition,
CRUX, CoverageBench, LANCER, nugget-based test collections) audits
coverage directly, but assumes pre-specified sub-questions, summaries,
or nuggets authored offline. \emph{Iterative RAG} (Self-RAG, FLARE,
IRCoT) issues follow-up queries to \emph{improve the generated answer},
not to evaluate the retriever, and does not surface a gap inventory.
\emph{TREC-style pooling} produces reusable qrels by aggregating
top-ranked documents across many systems, but is expensive to mount
per deployment and does not target a single pipeline.

\paragraph{Conceptual precedent.}
The closest conceptual precedent for Re:CAP is nugget-based
evaluation~\cite{voorhees2003trec}: a fixed inventory of atomic
information units against which a system is scored. Re:CAP is
inspired by this style of decomposition -- measuring coverage in
terms of discrete information units rather than whole-document
relevance -- but differs in two ways. First, the units (topics)
are \emph{induced} from $(D_0, A_0)$ and grown across iterations,
rather than pre-authored as gold. Second, Re:CAP does not score
against a fixed nugget set; gap probing extends the registry by
generating questions that may surface novel topics, and novelty
judging determines whether the retriever could have reached them.
The audit signal is the \emph{growth} of the registry under probing,
not its overlap with a held-out list. Re:CAP is the only entry in
Table~\ref{tab:positioning} that audits coverage on a single deployed
pipeline, needs no per-query labels, and returns an actionable
per-query gap inventory.

\begin{table}[h]
\centering
\footnotesize
\setlength{\tabcolsep}{2.5pt}
\resizebox{\columnwidth}{!}{%
\begin{tabular}{lccccc}
\toprule
Capability & RAG metrics & Cov. eval. & Iter. RAG & TREC pool & \textbf{Re:CAP} \\
\midrule
Audits retrieval coverage   & \textit{indir.} & \checkmark & --- & \checkmark & \textbf{\checkmark} \\
Probes for missing content  & --- & offline & \textit{gen.} & --- & \textbf{\checkmark} \\
Needs no per-query labels    & \textit{some} & --- & \checkmark & --- & \textbf{\checkmark} \\
Works on one deployed system & \checkmark & --- & \checkmark & --- & \textbf{\checkmark} \\
Output: gap inventory        & --- & facets & --- & qrels & \textbf{\checkmark} \\
\bottomrule
\end{tabular}
}
\caption{Re:CAP positioning. RAG metrics include RAGAS/ARES/RGB/
RAGChecker; Cov. eval. includes sub-question, CRUX, CoverageBench,
and nugget-coverage work. \textit{indir.}\,=\,indirectly (via answer
quality); \textit{gen.}\,=\,for generation, not for evaluation.}
\label{tab:positioning}
\end{table}

\section{Default Re:CAP Configuration}
\label{sec:appendix-defaults}

Re:CAP exposes about a dozen knobs (loop control, generator design,
LLM choice per component); the deployed default sets each to a
specific value, most of them backed by an ablation reported
elsewhere in this appendix.  Table~\ref{tab:defaults} lists every default with a
pointer to its justifying ablation, and Table~\ref{tab:llm-config}
breaks down the model and sampling temperature used at each
pipeline step.

\begin{table}[h]
\centering
\footnotesize
\setlength{\tabcolsep}{3pt}
\resizebox{\columnwidth}{!}{%
\begin{tabular}{lll}
\toprule
Choice & Default & Ablated in \\
\midrule
Gap-Qs / iteration       & 5         & App.\,\ref{sec:appendix-ablations} \\
MAX\_ITER                & 3         & App.\,\ref{sec:appendix-ablations} \\
MAX\_TOPICS              & 50        & --- \\
Gap-Q gen.\ input        & $Q + T + L$ & App.\,\ref{sec:appendix-ablations} \\
Probe roles (Step 3)     & 5         & App.\,\ref{sec:appendix-ablations} (LOO) \\
Anti-collapse guard      & on        & App.\,\ref{sec:appendix-ablations-loo} (LOO) \\
Failure memory           & last iter & App.\,\ref{sec:appendix-ablations-loo} (LOO) \\
Shared pipeline LLM      & yes       & App.\,\ref{sec:appendix-ablations} \\
Expansion top-$m$        & 50        & App.\,\ref{sec:appendix-ablations} \\
Judge temperature        & 0         & --- \\
Gap-Q gen.\ temperature  & 0.3       & --- \\
\textbf{Retriever}       & \textbf{hybrid BM25\,+\,dense} & \S\ref{sec:results-headline} \\
\bottomrule
\end{tabular}}
\caption{Default Re:CAP configuration. Hybrid retrieval uses RRF over
BM25 and MiniLM-L12 dense embeddings.  \emph{LOO}\,=\,leave-one-out.}
\label{tab:defaults}
\end{table}

The three remaining ``---'' rows are not ablated.  \textbf{MAX\_TOPICS}
($=\!50$) is a safety cap on topic-memory size that bounds prompt
growth across iterations; it sits above the per-query topic count on every benchmark run, so
on those workloads it acts as a guardrail rather than an active
parameter.  It is not inert in production: it binds on $8.0\%$ of
AI~Search and $40.5\%$ of NEWS~QA queries
(Table~\ref{tab:production-cohorts}), truncating the registry and
bounding $|G|$ and gap rate on those queries, so wide-$D_0$
deployments should raise it before reading either signal.  The two \textbf{temperature} rows
are deterministic-by-design conventions (judge / extractors at $0$;
generator at $0.3$ to give the role taxonomy room to diversify),
documented in the ``Why'' column of Table~\ref{tab:llm-config}
rather than ablated.

\begin{table}[h]
\centering
\footnotesize
\setlength{\tabcolsep}{4pt}
\begin{tabular}{lllc}
\toprule
Component & Model & T. & Why \\
\midrule
QA reader (Step 1)        & GPT-5.2  & 0   & strongest \\
Topic extract.\ (Step 2)  & GPT-4.1  & 0   & deterministic \\
Doc reconcile (Step 2)    & GPT-4.1  & 0   & deterministic \\
Gap-Q gen.\ (Step 3)      & GPT-4.1  & 0.3 & diversity \\
Gap judge (Step 5)        & GPT-4.1  & 0   & deterministic \\
\bottomrule
\end{tabular}
\caption{Per-component LLM configuration. Non-reader components share
a single \emph{pipeline model} (GPT-4.1) and are ablated jointly in
the pipeline-model swap (App.~\ref{sec:appendix-ablations}, E3.5).
T.\,=\,sampling temperature.}
\label{tab:llm-config}
\end{table}

\subsection{Post-hoc topic deduplication (Step 5b)}
\label{sec:appendix-dedup}

Parallel judging in Step 5 emits one verdict per candidate document,
so the same underlying topic is frequently surfaced by several
candidates within a single iteration under slightly different labels
(``Madonna referred to as the Queen of Pop'' vs.\ ``Madonna's Queen
of Pop title'').  Without deduplication these inflate the topic
count and the gap inventory $G$.  Re:CAP collapses them in two
stages before topics enter the registry $T$.

\paragraph{Stage A --- fuzzy string clustering (deterministic).}
The per-iteration novel verdicts (\textsc{NewTopic} and
\textsc{SubTopic}) are normalised (lowercased, whitespace collapsed)
and greedily clustered using Python's
\texttt{difflib.SequenceMatcher} ratio with a threshold of $0.85$.
For each cluster, the first label is taken as canonical, evidence
document IDs are merged across cluster members, and a
\textsc{SubTopic} verdict (with its parent) dominates if any member
produced one.  Stage A is deterministic and runs without an LLM
call.

\paragraph{Stage B --- semantic merge against $T$ (LLM).}
The Stage A survivors are passed to a single LLM call (GPT-4.1,
$T = 0$, batches of $\leq 30$ labels) together with the current
registry $T$.  This deduplication call --- distinct from the Step-5
novelty judge --- performs two jobs at once:
(i)~it clusters semantically equivalent new labels --- including
instance-of-the-same-category collapses (e.g.\ ``India's World Cup
wins'' and ``Australia's World Cup wins'' both fold under ``Cricket
World Cup winning countries'') --- and picks the most
category-level label as canonical; (ii)~for each cluster, it marks
\texttt{overlaps\_existing = true} together with the
\texttt{existing\_topic\_id} when the cluster duplicates a topic
already in $T$.  Clusters with \texttt{overlaps\_existing}
attach their evidence to the existing topic and are \emph{not}
counted as new gaps; the remaining clusters are added to $T$ and
to $G$ with the merged evidence set.

\section{Deployment Recipe}
\label{sec:appendix-deployment}

For teams operating a RAG system on a proprietary corpus, Re:CAP
plugs into the existing pipeline as an out-of-band auditor:
(1) on a representative query sample, run Re:CAP against the current
production retriever to establish a baseline (gap inventory, gap
rate, $N_q$ envelope, per-corpus cost);
(2) before promoting an index, embedder, or ranker change, re-run on
the same sample and compare;
(3) a drop of $\geq 5$\,pp in mean recall exceeds $10\times$ the
$0.47\%$ run-to-run noise floor of \S\ref{sec:results-variance} and
lies in the
discriminative range exercised by the BM25 sensitivity ladder of
\S\ref{sec:results-sensitivity}, so it can be read as a real
degradation rather than noise; we also watch unreachable-gold share
but report no threshold for it, having measured no run-to-run floor
for that quantity;
(4) feed the per-query gap inventory into operations dashboards so
on-call engineers see \emph{specifically which topics} were missed
rather than only an end-to-end faithfulness score.

\paragraph{Why hybrid retrieval.}
Re:CAP's tighter, entity-anchored probes can under-explore when
paired with a single-channel retriever; the dense channel in hybrid
supplies the breadth that BM25 alone cannot.  Hybrid is therefore
the recommended default.  The choice of initial retriever is not a
free parameter --- on HotPotQA with \emph{dense} $D_0$, Re:CAP
regresses by $3.6$\,pp matched-$N_q$ because the gap-Q generator
inherits dense-side biases and wastes early iterations on
semantically related but non-gold documents that BM25 would have
surfaced via bridge-entity lexical match (see Limitations,
\emph{Initial-retriever choice can cause cell-level regressions}).
The initial retriever should be selected via standard
recall-at-$k$ diagnostics, with Re:CAP then layered on top.
\paragraph{Attributing a metric move: retrieval drift vs.\ judge or
reader drift.}
Because Re:CAP has no per-query gold in production, a rise in its
monitoring signals is only actionable if a retrieval-side cause can be
separated from drift in the LLM components themselves.  Two properties
make that separation possible.  First, the components are pinned: the
novelty judge runs at temperature $0$ and the gap-Q generator at
$0.3$ (Table~\ref{tab:llm-config}), which removes deliberate sampling
variance once the model version is held fixed.  Pinning does not make
the components deterministic --- re-running the judge on identical
prompts reproduces its own novel/not verdict on $94.0\%$ of candidates
(App.~\ref{sec:appendix-rejudge}).  The operative threshold is therefore
the magnitude of a move relative to that noise, not its presence.  Second, that
residual non-determinism is bounded and measured end-to-end --- across
three independent runs (\S\ref{sec:results-variance},
Table~\ref{tab:variance}) Re:CAP recall has CV $0.47\%$ and gap rate CV
$1.46\%$, i.e.\ per-verdict disagreement largely averages out at the
level of the reported metrics.  Under a fixed
pipeline, a move exceeding those bounds cannot be produced by LLM
non-determinism alone and is therefore attributable to a change on the
retrieval side --- an index refresh, an embedder swap, or a corpus
shift.

Two caveats govern how the thresholds should be set.  Gap count $|G|$
responds to retriever quality in the expected direction
(\S\ref{sec:results-sensitivity}) but is the noisiest of the three
signals, at CV $5.04\%$; on the MuSiQue ladder the BM25--hybrid
separation is only $1.5\times$ that run-to-run standard deviation, so
$|G|$ resolves large regressions and not the adjacent-configuration
differences that ladder spans.  Gap rate is far more stable
($1.46\%$) but saturates on high-hop workloads, where it is best read
as a coverage-floor breach.  In practice, alert on gap rate for
gradual movement, use $|G|$ for magnitude once a breach fires, and
treat any $|G|$ move under roughly $2\times$ its noise floor as
unresolved rather than as evidence of stability.  Separately,
none of these bounds survive a \emph{model version} change: upgrading
the judge or generator re-baselines every signal, so the pre-change
sample must be re-run to re-establish the envelope before the new
version is trusted.  This is the same re-baselining discipline step
(2) above prescribes for retriever changes.

\paragraph{Cost-sensitive deployments.}
At \$$0.59$--\$$2.05$/q, auditing a thousand queries weekly costs
roughly \$$600$--\$$2{,}100$ per week at default settings.  Tighter monitoring budgets
can swap all pipeline components to GPT-4.1-mini for a $4.6\times$
cost reduction at $-0.51$\,pp paired recall on HotPotQA --- at the
edge of the $\pm 0.42$\,pp run-to-run noise floor of
\S\ref{sec:results-variance} (full ablation:
App.~\ref{sec:appendix-ablations}, E3.5).

\section{Production Deployment Case Studies}
\label{sec:appendix-production}

This appendix gives the full case-study description for the live
production audit summarised in
\S\ref{sec:results-production}.  We report two unrelated production
deployments --- ``AI Search'' (the original case study) and
``NEWS QA'' (a second cohort added to test transfer across deployments
with very different $D_0$ widths).  Both use the paper's
default configuration unchanged.
Table~\ref{tab:production-cohorts} summarises the two cohorts side
by side; the rest of this appendix details AI Search first and then
the NEWS QA deltas.

\begin{table}[h]
\centering
\footnotesize
\setlength{\tabcolsep}{4pt}
\resizebox{\columnwidth}{!}{%
\begin{tabular}{lrr}
\toprule
                                              & AI Search & NEWS QA \\
\midrule
Queries (success / total)                     & $200/200$ & $200/200$ \\
$D_0$ width (mean / median / max)             & $10$ (fixed) & $47.2$ / $33$ / $110$ \\
Topics/q: mean / median                       & $25.1$ / $20$       & $31.2$ / $34$ \\
$\mathrm{MAX\_TOPICS}{=}50$ saturated         & $16/200$ ($8.0\%$)  & $81/200$ ($40.5\%$) \\
Gap rate: mean / median / @$1.0$              & $0.991$ / $1.00$ / $94.5\%$ & $0.955$ / $1.00$ / $83.5\%$ \\
Topic source share: Step 1 / 2 / 5            & $21.9$ / $0.6$ / $77.4\%$    & $17.3$ / $4.4$ / $78.2\%$ \\
Natural convergence                           & $59/200$ ($29.5\%$) & $55/200$ ($27.5\%$) \\
Mean iterations (median)                      & $2.55$ ($3$)        & $2.13$ ($2$) \\
Per-q cost: median / mean                     & \$$1.68$ / \$$2.73$ & \$$1.79$ / \$$2.03$ \\
Per-q LLM calls: mean / max                   & $410$ / $1{,}703$   & $333$ / $1{,}114$ \\
Judge share of calls                          & $98.0\%$            & $97.6\%$ \\
\bottomrule
\end{tabular}}
\caption{Two production cohorts, same default Re:CAP
configuration.  Gap rate and judge dominance reproduce across both
deployments; NEWS QA's $\sim$$5\times$ wider $D_0$ shifts more
coverage to doc-reconciliation (Step 2) and pushes more queries to
the topic cap, but the iterative loop (Step 5) still contributes
$\sim$$78\%$ of topics on both.}
\label{tab:production-cohorts}
\end{table}

\paragraph{Deployment.}  ``AI Search'' is an enterprise research
assistant deployed over a proprietary, frequently re-indexed
knowledge corpus ($\approx$$123$M passages) that mixes news
feeds, internal research, and regulatory filings --- exactly the
non-stationary-with-multiple-sources-of-truth setting motivating the
paper.  Live queries span short factoid lookups (e.g.\ entity name
disambiguation), multi-document analyst briefs (e.g.\
``what's the Street saying about \emph{X}?''), and multi-paragraph
country / sector intelligence summaries (the long-form tail of the
distribution).  Per-query gold qrels do not exist in this setting
(corpus churn precludes exhaustive offline labelling), which is why
Re:CAP was developed in the first place; consequently, we report the
audit's intrinsic signals --- gap rate, topic-source provenance,
convergence, cost --- rather than recall against gold.

\paragraph{Configuration.}  The audit uses the paper's recommended
default (Table~\ref{tab:defaults}) without modification:
hybrid retrieval combining OpenSearch BM25 and an OpenAI
\texttt{text-embedding-3-large} 1024-d dense index; GPT-5.2 reader;
GPT-4.1 for the topic extractor, doc-topic reconciler, gap-Q
generator (at $T = 0.3$), and novelty judge (at $T = 0$);
$\mathrm{MAX\_ITER} = 3$; $5$ gap-Qs per iteration; expansion top-$m
= 50$ per gap-Q; $\mathrm{MAX\_TOPICS} = 50$.

\paragraph{Sample.}  $200$ representative production queries drawn
from the live query log without filtering on query type, length, or
expected complexity.  We deliberately included multi-paragraph
briefing queries that drive the cost long-tail rather than excluding
them as outliers.

\paragraph{Stability.}  All queries executed without any noticeable LLM failures. The only
implementation issue encountered were OpenSearch concurrency related. Mean iteration count is
$2.55$ (median $3$, $\sigma 0.71$, range $1$--$3$); $59/200 = 29.5\%$
of queries converge naturally before the iteration cap, with mean
convergence iteration $2.20$ implying that audits for internal data might benefit from an increased MAX\_ITERS cap.

\paragraph{Coverage signal.}  Mean total topics per query is $25.1$
(median $20$, $\sigma 16.1$, $\min 4$, $\max 50$); $16$ queries
saturate the $\mathrm{MAX\_TOPICS} = 50$ cap (long-form briefs).
The per-query gap rate (fraction of topics surfaced beyond what the
initial answer + reconciliation already cover; equivalently,
$|G|/|T_{\text{final}}|$) has mean $0.991$;
$189/200 = 94.5\%$ of queries have $\text{gap rate} = 1.0$, and the
lowest single-query gap rate is $0.60$ (a short factoid query with
substantial topic overlap between the initial retrieval and gap
expansion).  In aggregate, Re:CAP surfaces at least one
previously-unretrieved topic on effectively every production query sampled.

\paragraph{Where topics come from.}  Table~\ref{tab:production-sources}
breaks down the $5{,}017$ total topics across $200$ queries by their
source step in the loop.  $77.4\%$ of the per-query coverage map is
contributed by Step 5 (the gap judge's \textsc{NewTopic} verdicts on
documents fetched in gap iterations), with the remaining $21.9\%$
coming from Step 1 (the topic extractor on the initial answer) and
$0.6\%$ from Step 2 (doc-topic reconciliation on the initial
retrieval). The doc-reconciliation share is much lower than on the
benchmark runs -- production reader answers are comprehensive
enough that the reconciliation step rarely surfaces topics the
answer omitted, leaving virtually all coverage discovery to the
iterative loop.

\begin{table}[h]
\centering
\footnotesize
\setlength{\tabcolsep}{4pt}
\resizebox{\columnwidth}{!}{%
\begin{tabular}{lrr}
\toprule
Source step & Topics (200 q) & Share \\
\midrule
Step 1 --- Topic extractor on $A_0$         & 1{,}101 & 21.9\% \\
Step 2 --- Doc-topic reconciliation         &    32   &  0.6\% \\
Step 5 --- Gap judge (\textsc{NewTopic})    & 3{,}884 & 77.4\% \\
\midrule
\textbf{Total}                              & \textbf{5{,}017} & \textbf{100.0\%} \\
\bottomrule
\end{tabular}}
\caption{Topic provenance on the $200$-query production audit.
Step 5's $77.4\%$ share is the iterative loop's added value: more
than three-quarters of the per-query coverage map is surfaced only
because Re:CAP probes beyond the initial answer.}
\label{tab:production-sources}
\end{table}

\paragraph{LLM usage and cost.}  Mean LLM calls per query is $410$
(median $294$, $\max 1{,}703$); the judge accounts for $98.0\%$ of
calls ($80{,}422$ of $82{,}036$ across the audit), matching the
judge-dominance pattern of Table~\ref{tab:cost}.  Mean input tokens
per query is $1.18$\,M (median $699$\,k); mean output tokens $28.4$\,k
(median $19.8$\,k).  Per-query cost from the LLM-trace rollup (over
the $190$ queries with complete telemetry; $10$ queries lost their
\texttt{llm\_trace} payload during the resume that closed the
deterministic-failure tail) has median \$$1.68$, mean \$$2.73$
($Q_1 = \$0.56$, $Q_3 = \$3.96$, $\max = \$24.86$).  The median
sits within the benchmark envelope of \S\ref{sec:results-cost}
(\$$0.59$--\$$2.05$/q); the mean sits $\sim$$33\%$ above the upper
bound, driven by the long-form multi-paragraph briefing queries in
the long tail (the single \$$24.86$ query issued $1{,}643$ LLM calls
across $3$ iterations to map a $19$-topic intelligence brief).

\paragraph{NEWS QA cohort: deltas from AI Search.}  ``NEWS QA'' is
a second enterprise deployment over a multi-vendor news QA
corpus ($\approx$$70$M passages, $256$-d
\texttt{text-embedding-3-large} dense index).
We re-used the production pipeline's own retrieved set as $D_0$
(mean $47.2$ docs/q, median $33$, $\max 110$, vs $\approx$10 for
AI Search) and its $A_0$ as the reader output, so the audit
reflects exactly what the deployed system surfaced.  All other
defaults match Table~\ref{tab:defaults}.
Gap rate has mean $0.955$, $167/200$ ($83.5\%$) at
$\text{rate}=1.0$, $197/200$ ($98.5\%$) with at least one gap.
The $\sim$$5\times$ wider $D_0$ shifts the topic-source mix in
two ways: doc-topic reconciliation (Step 2) climbs from $0.6\%$ to
$4.4\%$ of topics, and the $\mathrm{MAX\_TOPICS}{=}50$ cap is
saturated on $81/200$ ($40.5\%$) of queries (vs $8.0\%$ on AI
Search), so reported gap counts on NEWS QA are right-censored more
often.  Mean iteration count is $2.13$ and natural convergence is
$27.5\%$, both within a percentage point of AI Search.
Per-query cost from the LLM-trace rollup has median \$$1.79$, mean \$$2.03$
($Q_1 = \$0.51$, $Q_3 = \$3.16$, $\max = \$6.87$).
The judge dominates calls at $97.6\%$ (vs $98.0\%$ on
AI Search), matching the benchmark cost pattern.

\paragraph{Worked example (i) --- AI Search (\texttt{idr news}).}
\textbf{Query.} \emph{``idr news''}.\quad
\textbf{$D_0$.} $10$ documents from the deployed hybrid retriever
(JPM Global Markets research; April $2026$ slice).\quad
\textbf{$A_0$ topics.} Bank Indonesia monetary policy and hawkish
stance on IDR; FX pressures and market dynamics; analyst forecasts
for $2026$; macro/fiscal headwinds; SRBI liquidity tightening.\quad
\textbf{Iter-1 gap-Qs (sample).} \emph{What recent developments
have occurred regarding SRBI and its role in stabilizing the IDR?};
\emph{What analyst forecasts contradict the prevailing outlook for
the IDR exchange rate in $2026$?}\quad
\textbf{New topics added.}  Iter-1; $4$ \textsc{SubTopic} and $1$
\textsc{NewTopic} after dedup.
SRBI-driven liquidity tightening on deposit and credit growth;
IDR depreciation on property developers' costs and debt exposure;
Middle-East de-escalation on Indo CDS and IDR outlook;
MSCI market-classification reforms on Indonesian equities;
equity-market sentiment under macro risk.\quad
\textbf{Iter-2.} $5$ further probes, no novel verdicts; the loop
converges.\quad
\textbf{Interpretation.} The deployed reader returned a coherent
top-line briefing; the loop doubled topic coverage by surfacing
five distinct second-order dimensions the initial answer left
implicit, without leaving the production retriever's reachable
candidate pool.

\paragraph{Worked example (ii) --- NEWS QA (Tyler Technologies).}
\textbf{Query.} \emph{``Why has Tyler Technologies' stock price been
declining and what are the issues affecting its performance?''}\quad
\textbf{$D_0$.} $7$ documents from production (Reuters and Benzinga,
February $2026$).\quad
\textbf{$A_0$ topics.} Government budget cuts on revenue; Q$4$
earnings and revenue miss; slower cloud migration and extended
procurement cycles; analyst downgrades and reduced price targets;
financial metrics indicating weak capital efficiency.\quad
\textbf{Iter-1 gap-Qs (sample).} \emph{What evidence exists that
contradicts the claim that Tyler Technologies has missed analyst
expectations or delivered disappointing financial results?};
\emph{Why has Tyler Technologies' stock price been declining,
regardless of specific economic conditions or government budget
changes?}\quad
\textbf{New topics added.}  Iter-1, all \textsc{NewTopic} after dedup.
Share-repurchase plan as a capital-allocation response to perceived
undervaluation; acquisition of \emph{For The Record} expanding the
court-technology portfolio; tail risks from cyber-attacks, AI
vulnerabilities, and regulatory changes.\quad
\textbf{Iter-2.} Two additional candidates clear the judge but
fold into existing topics in post-iteration dedup; the loop
converges with $\mathrm{new\_topics\_found} = 0$.\quad
\textbf{Interpretation.} The deployed pipeline correctly identified
the headline drivers of decline; the loop additionally surfaced the
company's response (buybacks, M\&A) and an unstated risk register,
both directly relevant to the \emph{why is the stock declining}
query and neither present in $A_0$.

\paragraph{Scope of these case studies.}  These audits confirm that
Re:CAP runs end-to-end on two unrelated live production deployments
with no production-specific code path, that the gap discovery rate
observed on benchmarks transfers to live traffic, and that the
iterative loop --- not the initial answer --- contributes the bulk
of the per-query coverage map across both cohorts despite a
$\sim$$5\times$ difference in $D_0$ width.  They do not provide
calibrated recall numbers (no production gold labels) or variance
estimates (single run per cohort).  Those roles remain with the four
benchmark datasets reported in
\S\S\ref{sec:results-headline}--\ref{sec:results-variance}.  Judge
accuracy on production traffic is bounded directly by the human read of
App.~\ref{sec:appendix-humanread-prod}, and indirectly by the $1.4\%$
judge-rejection rate of App.~\ref{sec:appendix-failure}.  That human
read samples only documents the judge called gap-filling, so it bounds
the judge's precision on this traffic and not the gaps it missed.

\section{Human Evaluation on the Production Cohorts}
\label{sec:appendix-humanread-prod}

\S\ref{sec:probe-quality} evaluates recovered documents on public
benchmarks, where relevance labels exist.  This appendix repeats the
exercise on the two production cohorts of
App.~\ref{sec:appendix-production}, where no relevance labels exist
and the corpora are proprietary.

\paragraph{Setup.}
We sampled $180$ documents that Re:CAP's novelty judge marked
as gap-filling, drawn from $180$ distinct queries and split
evenly between the two cohorts.  Three annotators saw the query, the
deployed reader's $D_0$-only baseline answer, and the document text,
blinded to the judge's verdict and to the cohort.  The rubric is the
binary \emph{new\_info} / \emph{covered} decision of
\S\ref{sec:probe-quality}.

\paragraph{Result.}
$75.6\%$ of judge-identified gap documents on
AI Search and $72.2\%$ on NEWS QA add information the
baseline answer lacks ($73.9\%$ overall, $[67.0$--$79.8]$;
Table~\ref{tab:humanread-prod}).

\begin{table}[h]
\centering
\scriptsize
\setlength{\tabcolsep}{3pt}
\begin{tabular}{lrrrc}
\toprule
Cohort & $n$ & new & Rate \% [95\% CI] & $\kappa$ \\
\midrule
AI Search & 90 & 68 & $75.6$ [$65.8$--$83.3$] & $0.84$ \\
NEWS QA & 90 & 65 & $72.2$ [$62.2$--$80.4$] & $0.88$ \\
\textbf{Overall} & \textbf{180} & \textbf{133} & \textbf{$73.9$ [$67.0$--$79.8$]} & $0.86$ \\
\bottomrule
\end{tabular}
\caption{Human-evaluation verdicts on judge-identified gap documents
from the two production cohorts.  \emph{new} is the majority-vote
\emph{new\_info} count; $95\%$ CIs are Wilson.}
\label{tab:humanread-prod}
\end{table}

\section{Full Cross-Dataset Results (Including TREC-COVID Matrix)}
\label{sec:appendix-headline-full}

Table~\ref{tab:headline-full} repeats the cross-dataset
comparison of \S\ref{sec:results-headline} and includes the full
$4$-cell retriever$\times$generator matrix on TREC-COVID, which we
elided from the main body for compactness.  The TREC-COVID matrix
illustrates the structural-complementarity finding of
\S\ref{sec:results-headline}: no combination of retriever (BM25 /
dense / hybrid) and generator ($(Q{+}T)$-only / five-mechanism)
closes the small negative recall delta against flat BM25
top-$500$, yet every cell maintains a $\geq 45\%$ vs-BM25-500
unreachable-gold share.  The default Re:CAP cell goes further: $21.2\%$
of its recovered gold remains absent even from the \emph{ensemble}
of flat BM25, dense, and hybrid top-$500$ combined
(Table~\ref{tab:unreachable}).  The negative recall delta is a
regime-level metric mismatch on pooled-graded corpora (see
Limitations, \emph{Recall@$k$ undercounts complementarity on
pooled-graded corpora}), not a cell-specific failure; the
audit-relevant signal is the complementarity.
  Appendix~\ref{sec:appendix-recall-vs-k} sweeps flat
BM25 depth $K$ on HotPotQA and TREC-COVID and locates the depth
at which flat retrieval catches up.

\begin{table*}[h]
\centering
\footnotesize
\setlength{\tabcolsep}{4pt}
\resizebox{\textwidth}{!}{%
\begin{tabular}{llrrrr}
\toprule
Dataset & Method & Docs seen & Recall [95\% CI] & $\Delta$ vs.\,RR & Notes \\
\midrule
\multirow{5}{*}{\textbf{MuSiQue}}
 & Flat BM25 matched-$N_q$         & 238 & 0.610 [0.558, 0.664] & $-0.291$ & matched-budget ref. \\
 & Flat MiniLM hybrid top-500       & 500 & 0.772 [0.723, 0.818] & $-0.129$ & RR wins at 48\% budget \\
 & Flat OpenAI dense top-500        & 500 & 0.854 [0.815, 0.896] & $-0.047$ & strongest dense baseline \\
 & \textbf{Re:CAP (hybrid)}         & \textbf{238} & \textbf{0.901 [0.854, 0.938]} & --- & $29.4\%$ unreach.\ vs.\,BM25-500 \\
 & \textbf{Re:CAP (dense)}          & \textbf{191} & \textbf{0.898 [0.862, 0.932]} & $-0.003$ & largest abs.\ lift vs.\,BM25 ($+30.2$\,pp) \\
\midrule
\multirow{4}{*}{\textbf{HotPotQA}}
 & Flat BM25 matched-$N_q$         & 249 & 0.827 [0.775, 0.878] & $-0.061$ &  \\
 & Flat MiniLM hybrid top-500       & 500 & 0.864 [0.813, 0.909] & $-0.024$ & only flat cell beating BM25-matched \\
 & Flat MiniLM dense top-500        & 500 & 0.773 [0.717, 0.828] & $-0.115$ & supports $+11.5$\,pp claim (\S\ref{sec:results-headline}) \\
 & \textbf{Re:CAP (hybrid)}         & \textbf{249} & \textbf{0.888 [0.837, 0.939]} & --- & $9.2\%$ unreach.\ vs.\,BM25-500 \\
\midrule
\multirow{3}{*}{\textbf{MultiHop-RAG}}
 & Flat BM25 matched-$N_q$         &  90 & 0.948 [0.913, 0.980] & $-0.030$ & ceiling test \\
 & Flat BM25 top-500                & 500 & 1.000 [1.000, 1.000] & $+0.022$ & corpus saturates \\
 & \textbf{Re:CAP (hybrid)}         & \textbf{90}  & \textbf{0.978 [0.955, 0.996]} & --- & ceiling reached at $5.5\times$ fewer documents \\
\midrule
\multirow{6}{*}{\textbf{TREC-COVID}$^{\dagger}$}
 & Flat BM25 matched-$N_q$         & 493 & 0.233 [0.201, 0.267] & $+0.068$ &  \\
 & Flat BM25 top-500                & 500 & 0.241 [0.211, 0.271] & $+0.076$ &  \\
 & Re:CAP (BM25, $(Q{+}T)$ gen.)    & 493 & 0.165 [0.138, 0.194] & --- & $52\%$ unreach.\ vs.\,BM25-500 \\
 & Re:CAP (BM25)                    & 383 & 0.209 [0.190, 0.230] & $+0.044$ & $45\%$ unreach. \\
 & Re:CAP (dense)                   & 298 & 0.178 [0.153, 0.208] & $+0.013$ & $61\%$ unreach. \\
 & \textbf{Re:CAP (hybrid)}         & \textbf{378} & \textbf{0.221 [0.195, 0.248]} & \textbf{best cell} & $-0.020$ vs.\,BM25-500; $48\%$ unreach. \\
\bottomrule
\end{tabular}}
\caption{Full cross-dataset matrix including the 4-cell TREC-COVID
retriever$\times$generator sweep.  Default $=$ hybrid retriever with
the five-mechanism generator (bold rows); the $(Q{+}T)$ row marks the
$(Q{+}T)$-only generator ablation.  $^{\dagger}$Every TREC-COVID cell
maintains $\geq 45\%$ vs-BM25-500 unreachable-gold share; the
default cell adds $21.2\%$ ensemble-unreachable on $98\%$ of
queries (Table~\ref{tab:unreachable}).  The small negative recall
delta is a metric mismatch on pooled-graded corpora, not a
cell-specific failure (see Limitations).}
\label{tab:headline-full}
\end{table*}

\section{Recall vs.\ Retrieval Depth}
\label{sec:appendix-recall-vs-k}

\begin{figure*}[h]
\centering
\includegraphics[width=\textwidth]{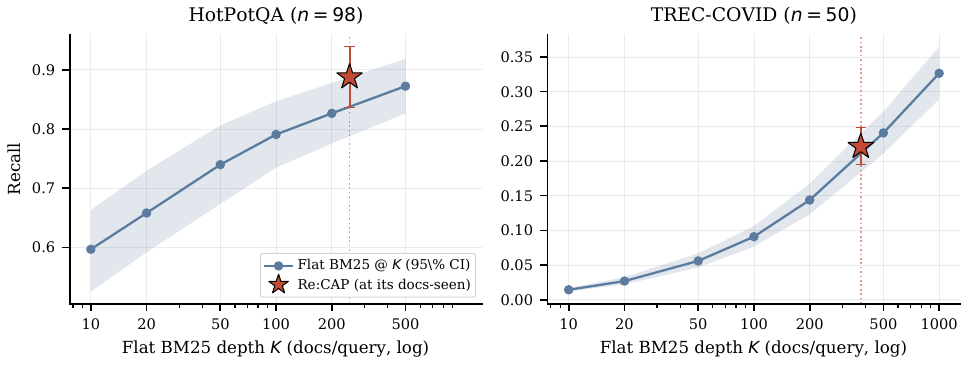}
\caption{Recall vs.\ flat BM25 retrieval depth $K$, HotPotQA and
TREC-COVID.  Blue: flat BM25 recall at depth $K$, shaded band is the
$95\%$ bootstrap CI.  Red star: Re:CAP at its own mean docs-seen
budget (dotted guide), with $95\%$ CI.  The $y$-axes are independent
--- the two corpora sit in very different recall regimes --- while the
$x$-axes are shared; HotPotQA's sweep stops at $K{=}500$.  Re:CAP
leads by a wide margin at shallow $K$ on both corpora and is caught
between $K{=}500$ and $K{=}1000$, earlier on TREC-COVID, the
scope-boundary case of \S\ref{sec:results-headline}.  Paired
per-query deltas with CIs in Table~\ref{tab:recall-vs-k}.}
\label{fig:recall-vs-k}
\end{figure*}

Table~\ref{tab:ladder} sweeps retrieval depth on MuSiQue and MS MARCO
TREC-DL.  This appendix completes that sweep on the two remaining
benchmarks, HotPotQA and TREC-COVID, over the full range of $K$ for
which we hold per-query judgements
(Figure~\ref{fig:recall-vs-k}, Table~\ref{tab:recall-vs-k}).
The curves are recomputed from the per-query records of the same runs
that back Table~\ref{tab:headline} rather than transcribed from it, so
the two cannot drift apart; the recomputation reproduces every
published flat-BM25 confidence interval exactly.  Deltas are
paired per query and then bootstrapped ($1000$ resamples, seed $42$),
the same estimator used throughout the paper.

Two observations.  First, Re:CAP's advantage is monotone decreasing in
$K$: it is largest where retrieval is shallow ($+29.1$\,pp on
HotPotQA and $+20.6$\,pp on TREC-COVID at $K{=}10$) and shrinks as
the flat baseline is permitted to see more of the corpus.  This is the
expected shape --- Re:CAP spends its budget on \emph{which} regions of
the corpus to probe, an advantage that necessarily erodes once a flat
run is allowed to read the corpus exhaustively.

Second, the depth at which the flat baseline catches up separates the
two corpora, and the separation is the scope boundary reported in
\S\ref{sec:results-headline}.  On HotPotQA, Re:CAP at 249
documents per query still leads flat BM25 at $K{=}200$ by
$6.1$\,pp and is statistically indistinguishable from $K{=}500$
while reading half as many documents.  On TREC-COVID, Re:CAP at
378 documents matches flat $K{=}500$ within noise and is then
overtaken at $K{=}1000$, by $10.6$\,pp at $2.6\times$ its
document budget.

The crossover is reported in full because on a pooled-graded corpus
Recall@$k$ rewards depth mechanically, since
deeper pools intersect more of the judged set (see Limitations,
\emph{Recall@$k$ undercounts complementarity on pooled-graded
corpora}).  The audit signal Re:CAP is built to produce --- gold that
no flat run surfaces at any depth we can afford in production --- is
unchanged by it: $21.2\%$ of Re:CAP's recovered gold on TREC-COVID is
absent from the \emph{union} of flat BM25, dense, and hybrid top-$500$
(Table~\ref{tab:unreachable}).  Aggregate recall at $K{=}1000$ is
therefore not the quantity the TREC-COVID result rests on.

\begin{table}[h]
\centering
\footnotesize
\setlength{\tabcolsep}{4pt}
\resizebox{\columnwidth}{!}{%
\begin{tabular}{llll}
\toprule
Data & $K$ & Flat BM25 [$95\%$ CI] & $\Delta$\,pp [$95\%$ CI] \\
\midrule
\multirow{6}{*}{\textbf{HotPotQA}} & 10 & 0.597 [0.526, 0.663] & $+29.1$ [$+22.4$, $+35.7$] \\
 & 20 & 0.658 [0.592, 0.730] & $+23.0$ [$+16.3$, $+29.6$] \\
 & 50 & 0.740 [0.673, 0.806] & $+14.8$ [$+8.7$, $+21.4$] \\
 & 100 & 0.791 [0.735, 0.847] & $+9.7$ [$+4.1$, $+15.8$] \\
 & 200 & 0.827 [0.776, 0.878] & $+6.1$ [$+0.5$, $+11.7$] \\
 & 500 & 0.872 [0.827, 0.918] & $+1.5$ [$-4.1$, $+7.7$]\,\emph{n.s.} \\
\midrule
\multirow{7}{*}{\textbf{TREC-COVID}} & 10 & 0.014 [0.012, 0.017] & $+20.6$ [$+18.3$, $+23.2$] \\
 & 20 & 0.027 [0.022, 0.032] & $+19.4$ [$+17.2$, $+21.8$] \\
 & 50 & 0.056 [0.046, 0.067] & $+16.5$ [$+14.4$, $+18.7$] \\
 & 100 & 0.091 [0.077, 0.106] & $+13.0$ [$+11.0$, $+15.2$] \\
 & 200 & 0.144 [0.123, 0.167] & $+7.7$ [$+5.4$, $+10.1$] \\
 & 500 & 0.241 [0.211, 0.271] & $-2.0$ [$-5.0$, $+1.1$]\,\emph{n.s.} \\
 & 1000 & 0.326 [0.289, 0.365] & $-10.6$ [$-14.2$, $-6.6$] \\
\bottomrule
\end{tabular}}
\caption{Flat BM25 recall at depth $K$, and the paired per-query delta
$\Delta$ (Re:CAP $-$ flat@$K$) in percentage points, for the curves of
Figure~\ref{fig:recall-vs-k}.  Re:CAP reads 249 documents per
query on HotPotQA and 378 on TREC-COVID, so rows below those
depths favour Re:CAP on budget as well as on recall.  \emph{n.s.}
marks a delta whose CI contains zero.}
\label{tab:recall-vs-k}
\end{table}

\section{Cross-model re-judgement}
\label{sec:appendix-rejudge}

The gap judge and the answer generator share a model family, so the
judge's accept/reject decisions are re-run against a newer generation to
test whether they are family-specific.  We draw a stratified sample of
$n = 1000$ judged candidates from the runs whose judge calls are
stored verbatim: every \textsc{new\_topic} and \textsc{sub\_topic} candidate in the pool, with \textsc{irrelevant} and \textsc{redundant} subsampled to fill the remainder.  We
replay {\em the judge prompt} against gpt-5.4-mini-2026-03-17 and gpt-5.5-2026-04-24.  Each model judges
every candidate three times and votes; agreement is between a
challenger's majority verdict and gpt-4.1-2025-04-14's own majority verdict on the
same prompts, so all three panels are measured the same way.

\begin{table}[h]
\centering
\small
\setlength{\tabcolsep}{4pt}
\resizebox{\columnwidth}{!}{%
\begin{tabular}{lrrr}
\toprule
Judge & Self-agreement & vs.\ gpt-4.1 & 95\% CI \\
\midrule
gpt-4.1 \emph{(reference)} & 94.0\% & --- & --- \\
gpt-5.4-mini & 89.5\% & 77.4\% & [74, 83] \\
gpt-5.5 & 95.7\% & 84.4\% & [81, 89] \\
\bottomrule
\end{tabular}}
\caption{Novel vs.\ not-novel agreement ($\kappa = 0.410$ (gpt-5.4-mini), $\kappa = 0.543$ (gpt-5.5)).  A candidate is
\emph{novel} if the judge assigns \textsc{new\_topic} or
\textsc{sub\_topic}; the topic registry counts both identically when
forming $|G|$, so this is the only verdict distinction any reported
quantity depends on.  On the full four-way taxonomy, which organises
the topic tree rather than producing a number, agreement is 54.1\% (gpt-5.4-mini), 59.9\% (gpt-5.5).
\emph{Self-agreement} is the mean pairwise agreement between
replicate runs of the same model on the same prompts, and bounds the
second column: gpt-4.1 reproduces its own verdicts on
94.0\% of candidates, so agreement above that level is not
attainable.  \mbox{gpt-5.5} is the most self-consistent model
measured, exceeding the deployed judge.
Candidates span 166 distinct
queries; intervals are bootstrapped over queries rather than
candidates, since candidates from one query share a topic registry and
a baseline answer.}
\label{tab:rejudge}
\end{table}

Models are run at temperature $0$ where the deployment permits it.  gpt-5.5-2026-04-24 serves only the API default of $1$, so that model is run at that setting; the resulting sampling variance is quantified by the self-agreement column.  Reasoning is disabled on every deployment that supports
it, matching the configuration of the deployed \mbox{gpt-4.1} judge, so
that model generation rather than deliberation budget is the variable
under test.

The judge prompt was written for the deployed model and may therefore disadvantage a newer one.  To test this, the relevance step was rewritten to admit intermediate (``bridge'') evidence, which multi-hop queries require and which the newer models were disproportionately rejecting; both challenger models were then re-run unchanged in every other respect.  Agreement moved by $+1.0$ and $+5.0$ points and the multi-hop rejection rate was unchanged, indicating that the disagreement reported here is not an artefact of prompt wording.

Our production infrastructure is Azure-OpenAI-only, so a truly cross-family (Anthropic / Google) audit is feasible on benchmark artefacts but not on production cohorts.  The sample above is therefore drawn entirely from
benchmark artefacts.

\section{Pseudo-Relevance Feedback Baseline (RM3)}
\label{sec:appendix-prf}

We compare Re:CAP against RM3 pseudo-relevance feedback
\cite{lavrenko2001rm,abduljaleel2004umass} as the closest
classical query-expansion control.  RM3 uses canonical defaults (no
per-dataset tuning): $k_{\text{fb}}=10$ feedback documents,
$n_{\text{exp}}=20$ expansion terms, $\lambda_{\text{orig}}=0.5$,
$\text{min\_df}=5$, maximum document-frequency ratio $0.5$;
stopwords, numerics, and original-query terms are excluded from the
expansion candidates.

\begin{figure*}[h]
\centering
\includegraphics[width=0.95\textwidth]{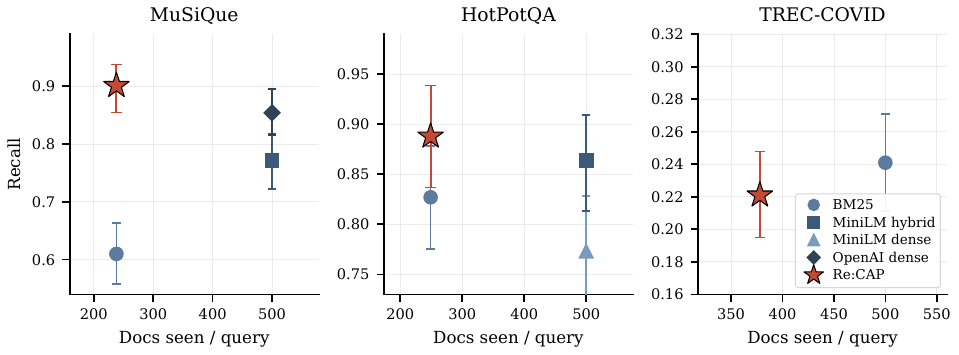}
\caption{Recall vs.\ documents seen per query, three datasets.
Flat baselines (blue circles) at their native top-$N$ budgets;
Re:CAP at the deployed default (red star).  On MuSiQue and
HotPotQA, Re:CAP attains the highest recall at less than half the
document budget of the strongest flat baseline.  TREC-COVID is the
scope-boundary case: Re:CAP recall is within noise of flat BM25
top-$500$, but $21.2\%$ of its recovered gold remains unreachable
by the ensemble of flat BM25, dense, and hybrid top-$500$ combined,
so the audit signal here is structural complementarity
rather than aggregate recall.  Error bars are $95\%$ bootstrap CIs;
MultiHop-RAG omitted (corpus saturates at flat top-$500$).
Numbers in Table~\ref{tab:headline}.}
\label{fig:recall-pareto}
\end{figure*}

\begin{table}[h]
\centering
\scriptsize
\setlength{\tabcolsep}{3pt}
\resizebox{\columnwidth}{!}{%
\begin{tabular}{llrrrr}
\toprule
Dataset & $D_0$ & Re:CAP & PRF $N_q$ & PRF 500 & $\Delta$ vs PRF 500 \\
\midrule
MuSiQue   & hybrid & \textbf{0.901} & 0.636 & 0.704 & \textbf{$+0.197$} \\
MuSiQue   & dense  & \textbf{0.898} & 0.624 & 0.704 & \textbf{$+0.194$} \\
HotPot    & hybrid & \textbf{0.888} & 0.832 & 0.888 & $0.000$ \\
HotPot    & dense  & 0.786          & 0.827 & 0.888 & $-0.102^{\dagger}$ \\
MHopRAG   & hybrid & 0.978          & 0.953 & 1.000 & $-0.022^{\star}$ \\
MHopRAG   & dense  & 0.974          & 0.895 & 1.000 & $-0.026^{\star}$ \\
\bottomrule
\end{tabular}}
\caption{Re:CAP vs.\ RM3 (PRF) across the six
main cells ($n=98$--$100$ per row).  PRF $N_q$ matches Re:CAP's
unique-candidate budget; PRF $500$ is the unbounded baseline.  Across
all six cells, RM3 lifts top-$500$ recall by at most $1.6$\,pp over
flat BM25 top-$500$.
$^{\dagger}$The HotPot-dense regression is the cell flagged under
Limitations (\emph{Initial-retriever choice can cause cell-level
regressions}): dense $D_0$ misses the bridge-entity lexical
signal that RM3 inherits from its BM25 backbone.
$^{\star}$MultiHop-RAG saturates at PRF top-$500$ on its $3.8$k-doc
corpus; Re:CAP wins at matched-$N_q$ by $+2.5$ to $+7.9$\,pp.}
\label{tab:prf}
\end{table}

On the bounded-evidence primary where flat retrieval leaves the most
room (MuSiQue), Re:CAP beats RM3 by $+19.4$ to $+19.7$\,pp at
top-$500$ and by $+26.5$ to $+27.4$\,pp at matched-$N_q$.  On
HotPotQA-hybrid Re:CAP matches PRF top-$500$ (both $0.888$) using
$249$ documents rather than $500$, and recovers $7.5\%$ of gold
absent from PRF top-$500$; on MuSiQue this unreachable share rises
to $28$--$30\%$.  The one Re:CAP regression cell (HotPot-dense) is
the dataset$\times$retriever combination already noted under
Limitations.  We do not report Bo1
\cite{amati2003probability} separately: on our datasets it tracks
RM3 within $\pm 1$\,pp and the conclusions are identical.

\section{Unreachable-Gold Diversification}
\label{sec:appendix-unreachable}

Table~\ref{tab:unreachable} reports per-dataset shares of gold
documents Re:CAP surfaces via gap probing that \emph{individual}
flat top-$500$ baselines (BM25, dense, hybrid) cannot reach, plus
the strictest cell: the \emph{ensemble}-unreachable share, i.e.,
gold absent from all three baselines combined ($1{,}500$ docs
total).  This operationalises the diversification claim
(\S\ref{sec:results-headline}): Re:CAP retrieves a structurally
distinct slice of the corpus that no single-retriever upgrade
recovers.  The ensemble shares are non-zero on every dataset
shown, peaking at $21.2\%$ on pooled-graded
TREC-COVID where the gold pool is largest and most diverse.

\begin{table}[h]
\centering
\footnotesize
\setlength{\tabcolsep}{4pt}
\resizebox{\columnwidth}{!}{%
\begin{tabular}{lrrrrrr}
\toprule
        & RR    & \multicolumn{4}{c}{Unreach.\ share vs.\ flat top-$500$} & Qs $\geq 1$ \\
\cmidrule(lr){3-6}
Dataset & gold  & BM25   & dense  & hyb.\  & \textbf{ens.}   & (ens.) \\
\midrule
MuSiQue              &    231 & 29.4\% & 14.8\% & 22.6\% & \textbf{10.0\%} & 21\%  \\
HotPotQA             &    174 &  9.2\% & 13.7\% &  3.7\% & \textbf{ 2.5\%} &  4\%  \\
TREC-COVID           & 4\,871 & 48.0\% & 48.0\% & 30.0\% & \textbf{21.2\%} & 98\%  \\
\bottomrule
\end{tabular}}
\caption{Per-baseline and ensemble unreachable shares: Re:CAP-recovered
gold absent from the indicated flat top-$500$ baseline, or from the
\emph{ensemble} of all three baselines combined ($1{,}500$ docs
total).  All Re:CAP runs use the deployed default (hybrid $D_0$,
five-mechanism generator).  Dense-$D_0$ Re:CAP gives shares within
$\pm 2$\,pp of the hybrid rows (not tabulated).  \emph{RR gold}
\,=\,total Re:CAP-recovered $(q, \text{doc})$ pairs summed across
queries.  \emph{Qs $\geq 1$ (ens.)}\,=\,share of queries with at
least one ensemble-unreachable doc.  MultiHop-RAG is omitted: flat BM25 top-$500$ reaches recall
$1.000$ on that corpus (App.~\ref{sec:appendix-headline-full}), so
no gold is unreachable and the shares are undefined.}
\label{tab:unreachable}
\end{table}

\section{Controlled Gold Deletion (E2.2)}
\label{sec:appendix-e22}

\paragraph{Protocol.}
For each MuSiQue query in the $M_0$ slice ($n=100$, hybrid retriever,
$N_q=5$, MAX\_ITER\,$=3$), we remove $K \in \{1, 2, \text{all}\}$
\emph{gold} passages from $D_0$ (drop-all removes every gold doc that
naturally surfaced in the top-$10$) and ask whether Re:CAP
re-discovers them through gap probing.  Nine queries have no gold
passage in their top-$10$ $D_0$ and are excluded from the drop
test, leaving $n=91$ per drop cell.  The reference is the no-drop
run restricted to the same $91$ queries, at recall $0.919$
($0.901$ over the full $100$; the excluded nine are precisely the
queries whose gold never surfaced).

\paragraph{Result.}
For drop-$1$, drop-$2$, and drop-all, Re:CAP recovers $98.9\%$,
$98.4\%$, and $100.0\%$ of the deleted gold IDs respectively, with
$97.8$--$100\%$ of queries achieving full recovery in each cell.
Paired recall on drop-$1$, drop-$2$, and drop-all drops by $11.9$,
$13.5$, and $10.9$\,pp respectively (vs.\ reference $0.919$).
The recall loss is \emph{not} attributable to failed recovery
($98$--$100\%$ recovery shown above); it is the \emph{secondary} gold
that no longer surfaces.  In the no-drop run, iterative expansion
picks up $\sim$$10$\,pp of gold beyond $D_0 \cap \text{qrels}$ by
triangulating from the answer; when $D_0$ is poorer, the seed answer
is impoverished and downstream gap-Qs find fewer new probe directions.  The
$11$--$13$\,pp paired loss is therefore a lower bound on the value of
a non-empty $D_0$, not a failure of probe-driven recovery.

\paragraph{The three cells are near-replicates.}
MuSiQue rarely places more than one gold passage in $D_0$: of the
$91$ queries, $62$ have exactly one, $26$ have two and $3$ have
three.  All three cells therefore delete identical documents on
$62/91$ queries, and drop-$2$ and drop-all coincide on $88/91$.
The spread across the three cells ($2.6$\,pp) is also smaller than
run-to-run non-determinism: on the $88$ queries where drop-$2$ and
drop-all delete the same documents, mean recall still differs by
$2.7$\,pp between the two runs.  E2.2 should therefore be read as
one recovery result replicated three times, not as a severity
ordering; the ordering of the cells is not resolvable here.

\begin{table}[h]
\centering
\footnotesize
\setlength{\tabcolsep}{3pt}
\begin{tabular}{lrrrrr}
\toprule
Drop & $n$ & recovery (\%) & recall & $\Delta$ pp & $|G|$ \\
\midrule
None (ref) &  91 & --- & $0.919$ & --- & $3.6$ \\
$K = 1$    &  91 & $98.9$  & $0.800$ & $+11.9$ & $4.9$ \\
$K = 2$    &  91 & $98.4$  & $0.785$ & $+13.5$ & $5.1$ \\
$K =$ all  &  91 & $100.0$ & $0.810$ & $+10.9$ & $5.4$ \\
\bottomrule
\end{tabular}
\caption{Controlled deletion (E2.2). ``recovery'' is per-query mean
(recovered $\cap$ dropped)/(dropped). $\Delta$\,pp $=$ paired
(ref $-$ cell) $\times$ 100. Gap count $|G|$ rises monotonically with
$K$ ($3.6 \to 5.4$, $+51\%$).  All columns are over the $91$
queries that had gold to delete.}
\label{tab:controlled-deletion}
\end{table}

\section{Ablations}
\label{sec:appendix-ablations}

\begin{figure}[t]
\centering
\includegraphics[width=\columnwidth]{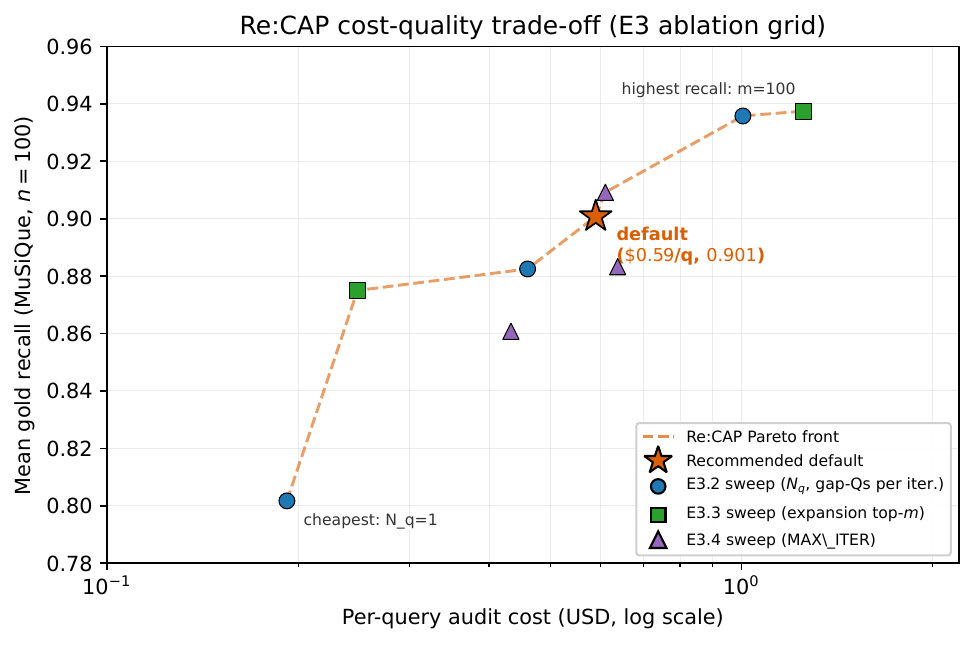}
\caption{Re:CAP cost-quality trade-off on MuSiQue
($n=100$, hybrid retrieval + redesigned generator). Star $=$
recommended default ($\$0.59$/q, recall $0.901$); dashed line $=$
Re:CAP Pareto front across the $E3$ ablation grid; markers indicate
which hyper-parameter sweep each point comes from (E3.2 $N_q$,
E3.3 expansion top-$m$, E3.4 MAX\_ITER).}
\label{fig:cost-pareto}
\end{figure}

We ablate Re:CAP along four dimensions:
\textbf{(i)} gap-Q generator design --- the $(Q{+}T)$-only ablation
vs.\ the five-mechanism deployed default (\S\ref{sec:appendix-ablations-gen});
\textbf{(ii)} per-lever leave-one-out within the five-mechanism
generator, attributing its lift to each of its five mechanisms
(\S\ref{sec:appendix-ablations-loo});
\textbf{(iii)} the three primary loop hyper-parameters --- gap-Qs
per iteration $N_q$, expansion depth $m$, and iteration cap
(\S\ref{sec:appendix-ablations-sweep});
\textbf{(iv)} the pipeline-LLM swap and a generator-vs-judge cost
decomposition (\S\ref{sec:appendix-ablations-llm}).
Table~\ref{tab:abl-summary} summarises the sweeps;
Figure~\ref{fig:cost-pareto} places each cell on the
MuSiQue cost-quality Pareto front.

Figure~\ref{fig:cost-pareto} visualises Re:CAP's internal
cost-quality trade-off across the $E3$ ablation grid on MuSiQue.
The recommended default ($N_q\!=\!5$, $m\!=\!50$, MAX\_ITER$\!=\!3$,
gpt-4.1 pipeline) sits at the elbow of the Pareto front: spending
$2\times$ more (the $N_q\!=\!10$ or top-$m\!=\!100$ cells) buys
$+3$--$4$\,pp recall, while spending $3\times$ less (the $N_q\!=\!1$
cell) costs $-10$\,pp.  The front is monotone in recall up to
$\sim$\$1/q --- no ablation simultaneously reduces cost and lifts
recall.
The chart is intra-Re:CAP: flat-retrieval baselines return ranked
documents but produce no coverage audit, so they are not plotted on
this axis; matched-budget recall comparisons appear in
Table~\ref{tab:headline}.

\subsection{Gap-Q generator design (E3.1)}
\label{sec:appendix-ablations-gen}

Table~\ref{tab:gen-ablation} reports the full
$2 \times 2 \times 3$ matrix (retriever $\times$ generator $\times$
dataset).  The five-mechanism generator beats the $(Q{+}T)$-only
ablation on hybrid by $+1.1$\,pp on MuSiQue and $+4.9$\,pp on
HotPotQA; MultiHop-RAG is at ceiling.  Two cells regress.  The larger --- MuSiQue with BM25 +
five-mechanism, $-5.4$\,pp --- is a retriever-coupling effect:
tighter probes under-explore on a single-channel retriever, and the
regression vanishes on hybrid.  The smaller is MultiHop-RAG on
hybrid ($-1.0$\,pp), where the corpus is already at ceiling.

\begin{table}[h]
\centering
\footnotesize
\setlength{\tabcolsep}{3pt}
\begin{tabular}{lrrrr}
\toprule
Dataset & BM25 \scriptsize{QT} & BM25 \scriptsize{5m.} & hyb.\ \scriptsize{QT} & \textbf{hyb.\ \scriptsize{5m.}} \\
\midrule
MuSiQue       & 0.888 & 0.834 & 0.890 & \textbf{0.901} \\
HotPotQA      & 0.813 & 0.864 & 0.838 & \textbf{0.888} \\
MultiHop-RAG  & 0.978 & 0.984 & 0.988 & \textbf{0.978} \\
\bottomrule
\end{tabular}
\caption{Gap-Q generator ablation (E3.1): Re:CAP mean recall by
retriever $\times$ gap-Q generator.  \emph{QT}\,=\,$(Q{+}T)$-only
baseline; \emph{5m.}\,=\,five-mechanism generator.  Default
in bold.}
\label{tab:gen-ablation}
\end{table}

\subsection{Per-lever leave-one-out (E3.1.2)}
\label{sec:appendix-ablations-loo}

Table~\ref{tab:lever-loo} attributes the five-mechanism generator's
lift to each of its mechanisms via a 15-cell LOO sweep (5 levers
$\times$ 3 bounded-evidence datasets, $100$\,q each on hybrid).  Paired
$\Delta$ is mean per-query (reference recall $-$ leave-one-out recall)
in pp; positive $\Delta$ means \emph{dropping} the lever hurts.  Cells
with $|\Delta| > 0.42$\,pp (the variance noise floor of
\S\ref{sec:results-variance}) are bolded.

\begin{table}[h]
\centering
\footnotesize
\setlength{\tabcolsep}{3pt}
\begin{tabular}{llrrr}
\toprule
& & MuSiQue & HotPotQA & MultiHop \\
Lever & Mechanism & $\Delta$ & $\Delta$ & $\Delta$ \\
\midrule
A & entity ledger        & \textbf{+8.58} & \textbf{+4.17} & \textbf{$-$1.47} \\
B & coverage topics       & $-$0.42       & \textbf{$-$2.06} & \textbf{$-$1.26} \\
C & probe roles           & \textbf{$-$1.00} & \textbf{$-$1.03} & \textbf{$-$1.45} \\
D & anti-collapse        & +0.33          & \textbf{+1.55}   & \textbf{$-$0.88} \\
E & failure memory        & \textbf{+0.75} & 0.00             & \textbf{$-$0.48} \\
\bottomrule
\end{tabular}
\caption{Per-lever leave-one-out (E3.1.2). Paired $\Delta$ in pp;
positive $=$ lever helps recall.  Lever A (entity-anchored gap-Q
generation) accounts for nearly all of the five-mechanism generator's
lift on the two bounded-evidence chain datasets; on near-ceiling
MultiHop-RAG it mildly hurts because entities are already pinned by
the question's surface form.}
\label{tab:lever-loo}
\end{table}

Decomposition: the five-mechanism generator is \textbf{one
dominant anchoring mechanism plus four supporting refinements}.  Lever
A's $+8.58$\,pp on MuSiQue and $+4.17$\,pp on HotPotQA recover the
entirety of the five-mechanism generator's hybrid-cell lift from
Table~\ref{tab:gen-ablation}.  On MultiHop-RAG, dropping A
\emph{raises} recall by $1.47$\,pp at near-ceiling reference: the
entity-ledger prompt over-constrains when the question itself names
the relevant entities (``between article X and article Y\dots'').

\subsection{Number of gap questions (E3.2), expansion depth (E3.3),
iteration depth (E3.4)}
\label{sec:appendix-ablations-sweep}

We swept each of the three primary hyper-parameters individually on
the MuSiQue $M_0$ slice ($100$\,q, hybrid, five-mechanism generator,
other defaults held).  All three curves confirm the default Re:CAP config sits at the cost-recall
knee.

\textbf{E3.2 $N_q$.} Recall climbs monotonically:
$0.802 \to 0.883 \to 0.901 \to 0.936$ for
$N_q \in \{1, 3, 5, 10\}$ at per-query cost
\$$0.19 / 0.46 / 0.59 / 1.01$.  $N_q\!=\!10$ buys $+3.5$\,pp paired over the
default $N_q\!=\!5$ at $+71\%$ cost.

\textbf{E3.3 expansion top-$m$.} Recall
$0.875 \to 0.901 \to 0.938$ at cost \$$0.25 / 0.59 / 1.25$ for
$m \in \{20, 50, 100\}$.  $m\!=\!100$ buys $+3.7$\,pp at $+113\%$
cost.

\textbf{E3.4 MAX\_ITER.} Iter\,$=\!1$ drops $4.0$\,pp paired (outside
the noise floor); iter\,$\in \{2, 3, 5\}$ are within
$\pm 2$\,pp of each other with iter\,$=\!2$ nominally best
($-0.83$\,pp vs.\ ref, roughly $2\times$ the $0.42$\,pp variance
noise floor of \S\ref{sec:results-variance} and roughly five times
smaller than the iter\,$=\!1$ drop).  Mean iterations actually run is
$1.00 / 1.75 / 1.79 / 1.91$ respectively (anti-collapse termination
dominates for $\geq 2$), so on the public benchmarks reported here
MAX\_ITER\,$> \!3$ buys neither recall nor convergence depth.  We
default to $3$ for headroom: on internal production traffic with
broader, more open-ended queries we observe occasional cases where
the loop still surfaces novel topics at iter\,$=\!3$ that
iter\,$=\!2$ misses, suggesting the public benchmarks under-sample
the regime where deeper iteration helps.

\subsection{Pipeline-model swap (E3.5) and generator-vs-judge cost
decomposition (E3.6)}
\label{sec:appendix-ablations-llm}

Swapping every \emph{pipeline} component (generator, judge, topic
extractor, reconciler) from GPT-4.1 to GPT-4.1-mini on HotPotQA drops
recall by $0.51$\,pp paired ($0.888 \to 0.884$, on the edge of the
$0.42$\,pp noise floor) while cutting per-query cost from \$$0.586$ to
\$$0.128$ ($-78\%$).  The reader (GPT-5.2) is held fixed.  Pipeline
LLM quality is largely interchangeable against a stronger reader, and
this is the largest cost lift available in the ablation grid.

To attribute the $E3.5$ saving, $E3.6$ isolates each component on the
same HotPotQA slice with reader/extractor/reconciler held at GPT-4.1.
(i) Generator $=$ mini, judge $=$ 4.1: recall $0.884$ (paired
$\Delta = +0.51$\,pp), cost \$$0.537$/q ($-8\%$).
(ii) Generator $=$ 4.1, judge $=$ mini: recall $0.867$ (paired
$\Delta = +2.06$\,pp, outside the noise band), cost \$$0.145$/q
($-75\%$).  The judge dominates both cost and quality; the
generator is interchangeable across model tiers.  This is the direct empirical
attribution behind the $\sim$$97\%$-judge claim in
Table~\ref{tab:cost} and behind the recommendation to keep the judge
at GPT-4.1 unless the deployment can tolerate $\sim$$2$\,pp recall in
exchange for the additional $\sim$$70\%$ cost saving.

\begin{table}[h]
\centering
\footnotesize
\setlength{\tabcolsep}{4pt}
\begin{tabular}{ll}
\toprule
Ablation & Sweep \\
\midrule
E3.1 Generator design & $(Q{+}T)$ / 5m. \\
E3.2 Num.\ gap-Qs    & 1 / 3 / 5 / 10 \\
E3.3 Expansion top-$m$ & 20 / 50 / 100 \\
E3.4 Iterations      & 1 / 2 / 3 / 5 \\
E3.5 Pipeline model  & 4.1-m / 4.1 \\
E3.6 Cross-model gen./judge & same / cross \\
\bottomrule
\end{tabular}
\caption{Ablation summary: the parameter sweeps reported across
\S\ref{sec:appendix-ablations}.  Per-cell numbers are in the
corresponding subsection.}
\label{tab:abl-summary}
\end{table}

\section{Component Validation and Generality}
\label{sec:appendix-component}

\subsection{Cross-dataset generality (E5.2)}

Table~\ref{tab:generality} reports the default configuration applied
identically across four datasets with no per-dataset tuning.

\begin{table}[h]
\centering
\footnotesize
\setlength{\tabcolsep}{3pt}
\begin{tabular}{lrrrrr}
\toprule
Dataset & $n$ & Recall & $\Delta$ vs.\,$N_q$ & $|G|$ & Iters \\
\midrule
MuSiQue       & 100 & \textbf{0.901} & \textbf{+0.291} & 3.6 & 1.79 \\
HotPotQA      &  98 & \textbf{0.888} & \textbf{+0.061} & 2.3 & 1.4 \\
MultiHop-RAG  &  98 & \textbf{0.978} & \textbf{+0.030} & 9.7 & 1.0--1.3 \\
TREC-COVID$^{\dagger}$ &  50 & 0.221 & $+0.023$ & 36.9 & 2.6 \\
\bottomrule
\end{tabular}
\caption{Cross-dataset generality (E5.2): default config applied with
no per-dataset tuning.  $\Delta$ is vs.\ flat BM25 at matched-$N_q$;
$|G|$ is mean Re:CAP gap count per query.}
\label{tab:generality}
\end{table}

\subsection{Judge ceiling and gap-Q quality from gold labels}

\textbf{Judge ceiling.}  The primary component-level
question is how often the LLM judge rejects a candidate that the
gold qrels mark relevant.  Across all $86$ loss queries and
$20{,}146$ missing-gold documents
(\S\ref{sec:appendix-failure}, Table~\ref{tab:fm-attrib}), $1.4\%$
of missing gold reached the candidate pool but was rejected by the
judge; on the bounded-evidence primaries (HotPotQA, MuSiQue, MultiHop-RAG)
the rate is $4.7\%$ ($3/64$).  The judge is therefore an empirically
tight upper bound on the audit in the regime Re:CAP is designed for:
$> 95\%$ of missing-gold losses are upstream of judging.  This is
consistent with LLM-as-judge studies on narrow relevance tasks
\cite{faggioli2023perspectives,thomas2024large,upadhyay2024llms,
upadhyay2024largescale}, while respecting cautions against replacing
full human qrels with LLM labels~\cite{soboroff2025dont}.

\textbf{Gap-Q quality (extrinsic).}  We evaluate gap-Q quality by
the most direct operational signal: whether the questions
recover gold the host retriever missed.  Three converging pieces of
evidence.
(i) \emph{Diversity:} mean pairwise LLM-embedding cosine $0.599$
across batches on the MuSiQue $100$-query anchor (target $\leq 0.7$;
max $0.758 \leq 0.85$).
(ii) \emph{Role coverage:} the five-role taxonomy enforces
$\geq 3$ of $5$ probe roles per batch.
(iii) \emph{Causal contribution:} the per-lever LOO sweep
(\S\ref{sec:appendix-ablations}, Table~\ref{tab:lever-loo})
attributes $+8.58$\,pp on MuSiQue and $+4.17$\,pp on HotPotQA to the
entity-anchored gap-Q lever alone --- the main recall lift
\emph{is} the gap-Q quality measurement under the audit's own
objective.

\subsection{Human-evaluation protocol on ensemble-unreachable gap documents}
\label{sec:appendix-probe-quality}

This appendix expands the body \S\ref{sec:probe-quality} human
evaluation: sample design and item-and-rubric protocol.  Verdict
counts are reported in Table~\ref{tab:probe-quality} in the body.

\textbf{Sample.} $n = 123$ items from the deployed default
(hybrid retrieval + five-mechanism generator).  TREC-COVID:
$n = 100$ stratified-random sample from the $1{,}034$
ensemble-unreachable docs in the $50$-query run, target $2$
docs/query covering all $50$ queries with $\geq 1$
ensemble-unreachable doc.  MuSiQue: census of all $n = 23$
ensemble-unreachable docs from the $100$-query run.  HotPotQA is
omitted ($4$ ensemble-unreachable docs, insufficient for per-doc
inference); MultiHop-RAG is omitted ($0$, corpus saturates);
MS MARCO is omitted (BM25-only setup; no ensemble defined).

\textbf{Item and rubric.} Per item the annotator sees the query,
the baseline reader's answer generated from the deployed default's
$D_0$ alone, and the candidate document text (truncated at
$\sim$$500$ words).  Blinded to: which baseline(s) miss the
document, qrels status, dataset name.  Binary rubric:
\emph{new\_info} if the document contains a substantive
query-relevant fact not in the baseline answer (a new entity, a
numeric or temporal anchor, a correction, an unambiguating
clarification); \emph{covered} if the relevant content is
substantively present in the baseline answer.  All sampled docs
are gold-relevant by construction (NIST TREC pool / MuSiQue dataset
authors), so the rubric does \emph{not} adjudicate relevance, only
informational novelty.

\begin{figure}[h]
  \centering
  \includegraphics[width=\columnwidth]{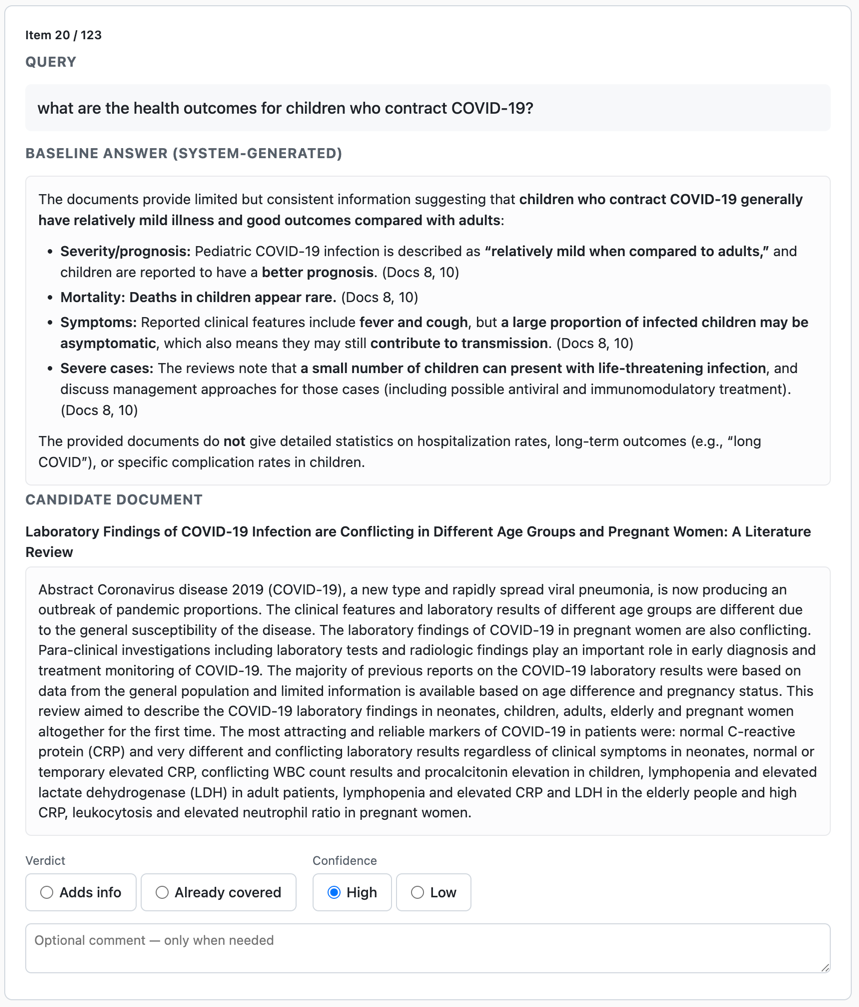}
  \caption{Annotation item TC-098 (\texttt{trec\_covid\_\_47\_\_qo3p62m4})
  as shown to annotator A1.  Top: query and $D_0$-only baseline
  answer.  Bottom: candidate document surfaced by Re:CAP's
  iteration-$3$ constraint-relaxed gap-question (text in main
  prose).  All three annotators (blinded to retriever, judge
  verdict, and dataset) selected \emph{new\_info}, while the
  loop's own novelty judge had marked the
  document \textsc{Redundant}.}
  \label{fig:annotation-example}
\end{figure}

\paragraph{Worked example.}
Figure~\ref{fig:annotation-example} shows one annotation item from
the TREC-COVID slice.  The deployed default's $D_0$ (hybrid
top-$10$) yields an $A_0$ on paediatric COVID-$19$ outcomes.
Re:CAP's gap-question generator produces an iteration-$3$
constraint-relaxed probe \textit{``What are the health outcomes
for children who contract viral respiratory infections, including
but not limited to COVID-$19$?''}; expanded retrieval on this
probe surfaces the candidate \textit{``Laboratory Findings of
COVID-$19$ Infection are Conflicting in Different Age
Groups\ldots''}, which adds paediatric-specific lab markers
(CRP, WBC, procalcitonin) absent from $A_0$.  The candidate is
qrels-positive yet missed by all three flat top-$500$ retrievers
(BM25, dense, hybrid).  Re:CAP's own novelty judge marks it
\textsc{Redundant} (collapsing the paediatric lab pattern into a
coarser parent topic --- the sub-mechanism over-aggregation
failure mode of App.~\ref{sec:appendix-failure}); all three
blinded annotators independently rate it \emph{new\_info}.

\section{Failure-Mode Analysis}
\label{sec:appendix-failure}

We classified $86$ \emph{loss queries} (where Re:CAP cumulative recall
fell below the matched-$N_q$ flat-retrieval floor) across four
datasets and $20{,}146$ missing-gold documents.
Tables~\ref{tab:fm-attrib} and~\ref{tab:fm-tags} report the
breakdown.  The aggregate $98.6\%$ no-surface /
$1.4\%$ judge-rejected split is dominated by TREC-COVID, which
contributes $20{,}082$ of $20{,}146$ missing-gold documents.

\paragraph{Two-regime interpretation.}
On bounded-evidence (HotPotQA, MuSiQue, MultiHop-RAG;
$n_{\text{loss}} = 43$, $64$ missing docs) the dominant failure modes
are early convergence ($41/43$), gap-Q off-topic from bridge-entity
erasure ($19/43$), and low topic yield ($30/43$) --- all addressable
on the generator side.  On pooled-graded (TREC-COVID,
$n_{\text{loss}} = 43$, $20{,}082$ missing docs) the dominant
failures are judge-rejected ($39/43$) compounded by iteration cap
($24/43$); both are structural consequences of binary-novelty judging
applied to pooled-graded relevance with $\sim$$493$ rel docs/query
at a $\sim$$378$-doc budget.

\paragraph{Targeted-recovery replay.}
We replayed the $33$ unique bounded-evidence loss queries under the
deployed default (hybrid + five-mechanism generator).  $22/33$
($67\%$) flipped from loss to tie or win; mean $\Delta$recall $+0.260$ per query.
Per dataset: MuSiQue-BM25 $7/11$ ($\Delta = +0.197$), HotPotQA-BM25
$11/18$ ($\Delta = +0.278$), MultiHop-RAG $4/4$ ($\Delta = +0.354$).
The residual $\sim$$33\%$ are largely cases where the entity ledger
anchors on a hallucinated entity from $A_0$.

\begin{table}[h]
\centering
\footnotesize
\setlength{\tabcolsep}{3pt}
\begin{tabular}{lrrrr}
\toprule
Dataset & $n_{\text{loss}}$ & Miss. & \%\,Judge & \%\,No-surf. \\
\midrule
HotPotQA (BM25)    & 18 &    26  &  0.0\%  & 100.0\% \\
MuSiQue (BM25)     & 11 &    17  &  5.9\%  &  94.1\% \\
MuSiQue (hybrid)   & 10 &    15  &  6.7\%  &  93.3\% \\
MultiHop-RAG       &  4 &     6  & 16.7\%  &  83.3\% \\
TREC-COVID         & 43 & 20\,082 &  1.4\%  &  98.6\% \\
\midrule
\textbf{Overall}   & \textbf{86} & \textbf{20\,146} & \textbf{1.4\%} & \textbf{98.6\%} \\
\textbf{AND only}  & \textbf{43} & \textbf{64} & \textbf{4.7\%} & \textbf{95.3\%} \\
\bottomrule
\end{tabular}
\caption{Missing-gold attribution per dataset.
\emph{Judge}\,=\,surfaced as candidate but judge-rejected;
\emph{No-surf.}\,=\,never reached the candidate pool.  The AND-only
aggregate ($95.3\%$ no-surface) is the relevant number for the
primary operational regime.}
\label{tab:fm-attrib}
\end{table}

\begin{table}[h]
\centering
\footnotesize
\setlength{\tabcolsep}{2pt}
\begin{tabular}{lrrrrrrr}
\toprule
Dataset & $n_l$ & EConv & OffT & ICap & Jdg & LowY & WkD \\
\midrule
HotPot (BM25)   & 18 & 18 & 6 & 0  &  0 & 15 & 10 \\
MuSiQue (BM25)  & 11 & 11 & 4 & 0  &  1 &  6 &  4 \\
MuSiQue (hyb.)  & 10 &  9 & 5 & 0  &  1 &  6 &  3 \\
MHopRAG         &  4 &  3 & 4 & 1  &  1 &  3 &  2 \\
TREC-COVID      & 43 &  1 & 0 & 24 & 39 &  0 &  0 \\
\midrule
\textbf{Overall} & \textbf{86} & \textbf{42} & \textbf{19} & \textbf{25} & \textbf{42} & \textbf{30} & \textbf{19} \\
\bottomrule
\end{tabular}
\caption{Failure-mode tags per dataset (non-exclusive).
\emph{EConv}\,=\,early convergence;
\emph{OffT}\,=\,gap-Q off-topic;
\emph{ICap}\,=\,iteration cap hit;
\emph{Jdg}\,=\,judge-rejected gold;
\emph{LowY}\,=\,low topic yield;
\emph{WkD}\,=\,$D_0$ recall = 0.}
\label{tab:fm-tags}
\end{table}

\subsection{Qualitative success examples}
\label{sec:appendix-qual-success}

Three worked examples in which Re:CAP recovers gold that flat
BM25 top-$500$ never reaches.  Each is a run from
\S\ref{sec:results-headline}; gap-Qs and topic labels are
verbatim from the iteration trace.

\paragraph{(i) Hedged-answer disambiguation (HotPotQA, hybrid).}
\textbf{Query.} \emph{``Kiley Dean was a back-up singer for what
famous singer, who was also known as the Queen of Pop?''}\quad
\textbf{Cell.} \texttt{hotpotqa-hybrid}, $D_0$ recall $= 0.50$
(\emph{Britney Spears} returned but not the
\emph{Queen of Pop} bridge), matched-$N_q$ flat $= 0.50$, flat-$500$
$= 0.50$; Re:CAP recall $= 1.00$ in $1$ iteration.\quad
\textbf{$A_0$ excerpt.} ``\emph{Kiley Dean sang back-up for
Britney Spears and Madonna \dots\ but they do not state which of
these singers was also known as the Queen of Pop.}''\quad
\textbf{Unreachable gold (flat-$500$).} \texttt{142056}
(\emph{Madonna} biographical article).\quad
\textbf{Recovering gap-Q.} \emph{``Which singer is widely
recognized as the Queen of Pop in music history?''}\quad
\textbf{Topic assigned.}
``\emph{Madonna is referred to as the `Queen of Pop'}''.\quad
\textbf{Interpretation.} The entity ledger pinned both candidates
(\emph{Britney Spears}, \emph{Madonna}) but a single
concept-anchored probe on the ambiguous \emph{Queen of Pop} role
surfaces the disambiguating article that flat retrieval ranks below
the top-$500$.

\paragraph{(ii) Triangulated multi-hop (MuSiQue, hybrid).}
\textbf{Query.} \emph{``What county is the city that shares a border
with the state capital of the state where Purrysburg is located?''}\quad
\textbf{Cell.} \texttt{musique-hybrid}, $D_0$ recall $= 0.25$
(only \emph{Purrysburg} retrieved), matched-$N_q$ flat $= 0.00$,
flat-$500$ leaves $4$ of $4$ supporting paragraphs unreachable;
Re:CAP recall $= 1.00$ in $2$ iterations.\quad
\textbf{$A_0$ excerpt.} ``\emph{The documents don't provide enough
information to determine this.  Doc 2 says Purrysburg is in
South Carolina, but none of the documents state South Carolina's
state capital\dots}''\quad
\textbf{Unreachable gold.}
\texttt{mq\_charleston\_south\_carolina\_9f68ae},
\texttt{mq\_forest\_acres\_south\_carolina\_35cf43},
\texttt{mq\_wwnq\_9c37a4},
\texttt{mq\_purrysburg\_south\_carolina\_254ff0}.\quad
\textbf{Iter-1 gap-Qs.} \emph{Geographical location of South
Carolina state capital Columbia; Which cities share borders with
Columbia, South Carolina; County jurisdiction of West Columbia,
South Carolina}.\quad
\textbf{Topics assigned.} ``\emph{Location and county of Forest
Acres, South Carolina}''; ``\emph{Geography and location of Forest
Acres, South Carolina}''.\quad
\textbf{Interpretation.} A canonical Re:CAP success: the loop
triangulates Purrysburg $\to$ South Carolina $\to$ Columbia
$\to$ Forest Acres in two iterations even though the initial answer
admits insufficient context --- the multi-hop pattern flat
retrieval cannot resolve in a single pass.

\paragraph{(iii) Cold-start bridge entity (HotPotQA, hybrid).}
\textbf{Query.} \emph{``What is the nationality of the film director
responsible for a 2008 American science fantasy film based on a
novel by Jeanne DuPrau?''}\quad
\textbf{Cell.} \texttt{hotpotqa-hybrid}, $D_0$ recall $= 0.00$
(neither film nor director retrieved), matched-$N_q$ flat $= 0.50$,
flat-$500$ misses the gold biography; Re:CAP recall $= 1.00$ in
$1$ iteration.\quad
\textbf{$A_0$ excerpt.} ``\emph{The documents provided do not
identify the specific 2008 American science fantasy film based on a
novel by Jeanne DuPrau, nor do they give the director's nationality
\dots}''\quad
\textbf{Unreachable gold.} \texttt{6167253}
(\emph{Gil Kenan} biography).\quad
\textbf{Recovering gap-Qs.} \emph{Director of City of Ember 2008
science fantasy film; Citizenship of Gil Kenan filmmaker}.\quad
\textbf{Topic assigned.}
``\emph{Nationality and biography of Gil Kenan, director of City of
Ember}''.\quad
\textbf{Interpretation.} The reader's ``insufficient context'' answer is
recovered into the topic registry; the gap-Q generator extracts the
DuPrau anchor, names the film (\emph{City of Ember}), pivots to the
director, and the judge promotes the biography to a new topic ---
the path flat retrieval cannot construct.

\subsection{Qualitative failure examples}
\label{sec:appendix-qual}

Three worked examples illustrating the three dominant bounded vs pooled-graded
failure modes.

\paragraph{(i) Bridge-entity erasure (MuSiQue, hybrid).}
\textbf{Query.} \emph{``A line with Williamsburg, Main Street and
another station are in a state that's next to an ocean.  When did
that ocean start to open up?''}\quad
\textbf{Cell.} \texttt{musique-hybrid}, recall $= 0.50$
(matched-$N_q$ flat $= 0.75$, flat-500 $= 0.75$), $D_0$ recall $= 0$,
$2$ iterations, $7$ topics, $361$ candidates judged.\quad
\textbf{Gold never surfaced.} \texttt{Newport News, Virginia};
\texttt{Virginia} (the bridge state).\quad
\textbf{Sample gap-Qs.} \emph{Geological history of Atlantic seafloor
formation near New York}; \emph{Earliest rifting events of Pangaea
affecting eastern North America}; \emph{Age of Atlantic Ocean crust
adjacent to Brooklyn shoreline}.\quad
\textbf{Diagnosis.} The bridge entity \emph{Virginia} is required to
link \emph{Williamsburg} to \emph{Atlantic Ocean}, but neither $D_0$
nor $A_0$ surface it (the loop guesses \emph{New York}/
\emph{Brooklyn} instead), so the entity ledger never anchors a gap-Q
on \emph{Virginia}.  Dominant \emph{OffT} $+$ \emph{WkD} pattern on
the BM25 cells of MuSiQue.

\paragraph{(ii) Hallucinated anchor (HotPotQA, BM25).}
\textbf{Query.} \emph{``What is the nationality of the star of The
Monster who was also in Swordswallers and Thin Men and The Savages?''}\quad
\textbf{Cell.} \texttt{hotpotqa-bm25}, recall $= 0.00$ (matched-$N_q$
flat $= 0.50$, flat-500 $= 0.50$), $D_0$ recall $= 0$, $1$ iteration,
$1$ topic, $0$ \textsc{NewTopic} verdicts, $137$ candidates judged.\quad
\textbf{Gold never surfaced.} \texttt{Zoe Kazan}; \texttt{The Monster
(2016 film)} (the correct bridge).\quad
\textbf{Sample gap-Qs.} \emph{Which country is Toma Caragiu from, the
actor in The Monster, Swordswallers and Thin Men, and The Savages?};
\emph{What is the citizenship of Toma Caragiu \dots}; \emph{Where was
Toma Caragiu \dots born?}\quad
\textbf{Diagnosis.} The reader hallucinated \emph{``Toma Caragiu''}
as the star in $A_0$ (Caragiu is a real Romanian actor but was not in
any of the listed films); the entity ledger pinned on this fabricated
name and every gap-Q probed around the wrong person.  The judge had
no opportunity to recover \emph{Zoe Kazan} because she was never
returned as a candidate.

\paragraph{(iii) Pooled-graded saturation (TREC-COVID, BM25 + $(Q{+}T)$-only).}
\textbf{Query.} \emph{``What is the mechanism of inflammatory response
and pathogenesis of COVID-19 cases?''}\quad
\textbf{Cell.} \texttt{trec-covid topic 38}, recall $= 0.02$
(matched-$N_q$ flat $= 0.12$, flat-500 $= 0.19$), $D_0$ recall $= 0.01$,
$1$ iteration (non-converged), $50$ topics (cap), $125$ \textsc{NewTopic}
verdicts, $249$ candidates judged.  Gold pool size: $\sim$$1{,}000+$
partially-relevant documents.\quad
\textbf{Gold judged-but-rejected (\textsc{SubTopic}).}
\texttt{17$\beta$-Estradiol\dots}\ folded under ``Estradiol and sex
hormone modulation of COVID-19 inflammation'';
\texttt{Why is SARS-CoV-2 infection more severe in obese men?\dots}
folded under ``Gut-lung axis\dots''.\quad
\textbf{Sample gap-Qs.} \emph{Role of viral spike protein in
triggering lung epithelial cell injury}; \emph{Interaction between
SARS-CoV-2 and ACE2 receptor leading to tissue damage};
\emph{Contribution of neutrophil extracellular traps to COVID-19
lung pathology}.\quad
\textbf{Diagnosis.} The gap-Qs are genuinely on-topic and the loop
produces $> 100$ \textsc{NewTopic} verdicts within the topic cap, but
the corpus has thousands of partially relevant documents and most
gold is never surfaced.  The judge over-aggregates novel sub-mechanisms
into existing topics --- a structural mismatch between binary-novelty
judging and pooled-graded relevance.

\section{Future Work}
\label{sec:appendix-future}

\begin{itemize}
\item \emph{Robust entity grounding} --- cross-check ledger entities
against $D_0$ before anchoring; downgrade anchors to soft preferences
when $D_0$ confidence is low.  Targets the residual $\sim$$33\%$ of
unrecovered bounded-evidence loss queries.

\item \emph{Judge-cost reduction} (highest engineering leverage per
\S\ref{sec:results-cost}): batch judging across multiple candidates
per LLM call; a lightweight cross-encoder pre-filter dropping $70$--$80\%$
of candidates; topic-aware short-circuit when the matched topic is
already saturated.

\item \emph{Multi-paraphrase gap-Qs per topic}: a queued lever for the
remaining \emph{OffT} failures the entity ledger does not catch.

\item \emph{Online gap discovery} as a production monitoring tool, and
the per-query gap-topic list as a human-readable audit artifact
that names the retriever's blind spots.

\item \emph{Multilingual extension} and \emph{graph-structured
topic registry} (replacing the flat list with a depth-3 rooted DAG,
enabling multi-resolution metrics and graph-merge deduplication).

\item \emph{Gaps as training signal}: discovered gaps as hard negatives
for retriever fine-tuning --- closing the loop from evaluation back
to improvement.

\item \emph{Gap inventory as synthetic qrels for retriever
benchmarking}: each $(Q, \text{gap-Q}, \text{judged-novel doc})$
triple is a relevance label produced without human annotation.  The
unreachable-gold result (\S\ref{sec:results-headline},
Table~\ref{tab:headline}: $9$--$29\%$ of recovered gold is absent
from flat BM25 top-$500$ on the bounded-evidence primaries, rising
to $48\%$ on TREC-COVID) shows these labels contain documents that
strong flat baselines cannot surface, suggesting the inventory can
serve as a low-cost evaluation set for comparing alternative ---
including inference-optimised --- retrievers without re-running the
full audit.
\end{itemize}

\section{Prompt Templates}
\label{sec:appendix-prompts}

Every prompt used by the Re:CAP loop is reproduced below with the
structured-output schema it is sent with, generated directly from the
pipeline's definitions so that the listings cannot diverge from the
code that runs. The schemas constrain the label set the model may
return and are therefore part of the specification: a template alone
does not determine the reported behaviour. The
information-equivalence judge of \S\ref{sec:appendix-e22} is included
for completeness; it belongs to that controlled experiment rather
than the loop. Two
presentation-only changes were applied: long lines are hard-wrapped,
and non-ASCII glyphs are transliterated
(\texttt{[+]} for a check mark, \texttt{[-]} for a ballot X,
\texttt{->} for a right arrow, \texttt{--} for an em dash). Prompt
wording is otherwise unmodified. Prompts use
\texttt{\{placeholder\}} substitution; temperature and model settings
are in Table~\ref{tab:llm-config}.

\begin{table*}[t]
\begin{tcolorbox}[promptbox, title={Topic extraction (Step 1) (1 of 2)}]
\begin{verbatim}
# Topic Extraction

Identify the distinct information facets present in the answer below.

## Input

**Query:** {query}

**Answer:**
{answer}

## Instructions

List every distinct piece of information this answer provides in response to the query.
A "topic" is a specific facet of information that helps answer the query.

- If the answer states it cannot answer, has no information, or is empty/unhelpful -> return an
  empty list.
- Only extract topics for **actual information provided**, not meta-statements about lack of
  information.
- Do **not** invent topics -- only extract what the answer actually contains.
\end{verbatim}
\end{tcolorbox}
\end{table*}

\begin{table*}[t]
\begin{tcolorbox}[promptbox, title={Topic extraction (Step 1) (2 of 2)}]
\begin{verbatim}
## Granularity rules

- A topic is an **information facet** (a type of information), not a single data point or
  instance.
- **Paragraph test:** each topic should warrant its own distinct paragraph. If two topics would
  produce overlapping paragraphs, merge them.
- If the answer lists multiple instances of the same type (e.g. several countries, people,
  events), group them into **one** topic describing the category.
  - [+] `Countries that have won the FIFA World Cup`
  - [-] `Brazil won 5 times` + `Germany won 4 times` + `Italy won 4 times`
- Target: **1-5 topics** for a typical answer. More than 8 almost certainly means
  over-enumeration.

## Output format

Each topic: a short descriptive label (7-12 words) describing what information is provided.
\end{verbatim}
\end{tcolorbox}
\end{table*}

\begin{table*}[t]
\begin{tcolorbox}[promptbox, title={Topic extraction (Step 1) --- response schema}]
\textbf{Enforced response schema} --- the model is constrained to return exactly these fields:
\begin{verbatim}
topics: list[str]
    List of distinct topic labels extracted from the answer. Each label is a short phrase
    (5-15 words) describing a specific piece of information. Empty list if the answer
    contains no useful information.
\end{verbatim}
\end{tcolorbox}
\end{table*}

\begin{table*}[t]
\begin{tcolorbox}[promptbox, title={Entity-ledger initialisation (Step 1b)}]
\begin{verbatim}
Extract LITERAL anchor strings from the query and the answer below.
These anchors will be used to ground follow-up search queries -- every
extracted anchor must appear VERBATIM in the source text (not a
paraphrase, not a normalisation).

ORIGINAL QUERY: {query}

INITIAL ANSWER:
{answer}

Two output lists:

1. QUERY ANCHORS -- literal named entities from the QUERY only.
   Include: people, organisations, locations, titles of works, dates,
   version numbers, named events. Do NOT include common nouns or
   generic concepts.

2. ANSWER ANCHORS -- literal named entities or numerical/temporal anchors
   from the ANSWER that are NOT already in the query. These are the
   *bridge* anchors -- the names the query did not mention but the
   answer discovered. They are typically the most valuable for finding
   missing evidence.

Rules:
- Each anchor must be a short literal substring of the source text.
  Use exact capitalisation and spelling as in the source.
- Do not include the same anchor in both lists -- query anchors take
  precedence.
- Skip anchors that are too generic to be useful for retrieval
  (e.g. "year", "country", "person").
- It is fine for either list to be empty.
- Target 1-6 anchors per list.
\end{verbatim}
\end{tcolorbox}
\end{table*}

\begin{table*}[t]
\begin{tcolorbox}[promptbox, title={Entity-ledger initialisation (Step 1b) --- response schema}]
\textbf{Enforced response schema} --- the model is constrained to return exactly these fields:
\begin{verbatim}
query_anchors: list[str]
    Named entities (people, organisations, locations, works, dates) extracted verbatim from
    the ORIGINAL QUERY. Each anchor is a short literal string that should appear in gap
    questions.
answer_anchors: list[str]
    Named entities or numerical/temporal anchors extracted verbatim from the ANSWER A_0 that
    are NOT already in query_anchors. These are the bridge entities -- the items the query
    did not mention but the answer discovered.
\end{verbatim}
\end{tcolorbox}
\end{table*}

\begin{table*}[t]
\begin{tcolorbox}[promptbox, title={Doc--topic reconciliation (Step 2) (1 of 2)}]
\begin{verbatim}
# Document-Topic Reconciliation

Reconcile retrieved documents against a list of known topics.

## Input

**Query:** {query}

**Known topics:**
{topic_list}

**Retrieved documents:**
{documents}

## Process (for each document)

### Step 1 -- Entity check
Is this document about the **same entity/subject** as the query? If not -> `OFF_TOPIC`.

### Step 2 -- Novelty check (only if same entity)
Compare the document's information against the known topics list.
- Covers the same ground as an existing topic (even with different details) -> `REDUNDANT` --
  cite the topic number(s).
- Provides a genuinely **new** type of information -> `NEW_TOPIC` -- provide a short label (7-12
  words).
\end{verbatim}
\end{tcolorbox}
\end{table*}

\begin{table*}[t]
\begin{tcolorbox}[promptbox, title={Doc--topic reconciliation (Step 2) (2 of 2)}]
\begin{verbatim}
## Granularity rules

- A "topic" is an **information facet** (a type/category of information), not a single data
  point, instance, or example.
- **Paragraph test:** would this topic warrant its own distinct paragraph that does not overlap
  with any existing topic? If not -> `REDUNDANT`.
- Another instance of a category already in the topic list -> `REDUNDANT`, not new.
  - Example: "Countries that won the World Cup" exists -> a document about a specific country's
    win is `REDUNDANT`.
- Sub-topics must represent a genuinely **different dimension**, not just deeper detail of the
  same dimension.
- When in doubt -> `REDUNDANT`. Prefer fewer, higher-quality topics over many overlapping ones.

## Classification labels

| Label | When to use |
|---|---|
| `OFF_TOPIC` | Not about the same entity, or not relevant |
| `REDUNDANT` | Covers same ground as known topics -- cite topic number(s) |
| `NEW_TOPIC` | Genuinely new information facet -- provide a 7-12 word label |
\end{verbatim}
\end{tcolorbox}
\end{table*}

\begin{table*}[t]
\begin{tcolorbox}[promptbox, title={Doc--topic reconciliation (Step 2) --- response schema}]
\textbf{Enforced response schema} --- the model is constrained to return exactly these fields:
\begin{verbatim}
documents: list[DocReconciliationEntry]
    One entry per document, in the same order as the input.

DocReconciliationEntry:
  doc_number: int
      1-based document number.
  entity_match: bool
      True if the document is about the same entity/subject as the query.
  maps_to: list[int]
      Topic numbers from the known list that this document covers.
  classification: REDUNDANT | NEW_TOPIC | SUB_TOPIC | OFF_TOPIC
      Classification of this document.
  new_topic_label: str | None
      If NEW_TOPIC or SUB_TOPIC: the label for the new topic. Otherwise null.
  parent_topic: str | None
      If SUB_TOPIC: the parent topic label or number. Otherwise null.
\end{verbatim}
\textsc{off\_topic} here is the reconciler's name for the outcome the novelty judge calls \textsc{irrelevant}; the two steps use different labels for the same decision and both count as non-novel.
\end{tcolorbox}
\end{table*}

\label{sec:appendix-probe-roles}
\noindent The five probe roles tagged by the gap-question generator
are:
\begin{itemize}
  \item \emph{entity-anchored} --- probe centred on a named entity from $L$;
  \item \emph{concept-anchored} --- probe centred on an abstract noun phrase rather than a specific entity;
  \item \emph{constraint-relaxed} --- drops a constraint of $Q$ to broaden retrieval;
  \item \emph{constraint-tightened} --- adds a constraint to narrow it;
  \item \emph{inverse-negation} --- probes for contradictory or counterexample evidence.
\end{itemize}
\begin{table*}[t]
\begin{tcolorbox}[promptbox, title={Gap-probing question generation (Step 3) (1 of 2)}]
\begin{verbatim}
### Task

You are generating SEARCH QUERIES that will be used to retrieve documents
likely to answer the ORIGINAL QUERY below. The goal is to surface
documents that the original query phrasing has NOT yet retrieved.

ORIGINAL QUERY: {query}

--- ENTITY LEDGER ---
These are LITERAL anchor strings extracted from the query and the current
answer. The "answer anchors" in particular are bridge entities that the
initial query did not mention but that subsequent retrieval should
target.

{entity_ledger}

--- TOPICS WITH COVERAGE STRENGTH ---
Each topic shows its current evidence count. Topics marked (weak) have
little supporting evidence and should be prioritised. Topics marked
(strong) are well-covered already -- do not over-probe them.

{topic_list_with_coverage}

--- PREVIOUS GAP QUESTIONS THAT YIELDED NO NEW TOPICS ---
Do NOT repeat or paraphrase any of these. Generate questions that probe
DIFFERENT retrieval directions.

{failed_gap_questions}

--- PROBE ROLES (your batch must be diverse) ---
Every question must declare exactly one role. Across the batch the roles
MUST vary -- do not pick the same role + same anchor twice.
\end{verbatim}
\end{tcolorbox}
\end{table*}

\begin{table*}[t]
\begin{tcolorbox}[promptbox, title={Gap-probing question generation (Step 3) (2 of 2)}]
\begin{verbatim}
- entity-anchored:      Pivot on a named entity from the LEDGER and ask
                        a different question about it. The literal
                        anchor string MUST appear verbatim in the
                        question. Use this when bridge entities matter.
- concept-anchored:     A query about a concept or mechanism in the
                        answer (no named entity). Use when the gap is
                        conceptual.
- constraint-relaxed:   The original query with ONE specific constraint
                        (date, location, qualifier) REMOVED. Use to
                        widen retrieval.
- constraint-tightened: The original query with ONE specific anchor from
                        the LEDGER ADDED. The anchor MUST appear
                        verbatim. Use to narrow retrieval onto a
                        specific entity.
- inverse-negation:     Ask the inverse / counter perspective (e.g.
                        "What does NOT support ...", "What contradicts
                        the claim that ..."). Use sparingly -- at most one
                        per batch.

--- HARD RULES ---

1. Generate EXACTLY {num_questions} questions.
2. Each question must seek information that helps answer the ORIGINAL
   query.
3. For roles `entity-anchored` and `constraint-tightened`: the `anchor`
   field MUST be one of the literal anchors from the ENTITY LEDGER, and
   it MUST appear verbatim inside the question text.
4. Roles MUST NOT all be the same. Across the batch, use at least 3
   different roles when {num_questions} >= 4.
5. If a topic has `(weak: 0 docs)` or `(weak: 1 doc)`, set
   `targets_topic_id` to that topic's id on at least one question
   (when possible without contradicting the diversity rule).
6. Do not repeat or paraphrase anything in the FAILED GAP QUESTIONS list.
\end{verbatim}
\end{tcolorbox}
\end{table*}

\begin{table*}[t]
\begin{tcolorbox}[promptbox, title={Gap-probing question generation (Step 3) --- response schema}]
\textbf{Enforced response schema} --- the model is constrained to return exactly these fields:
\begin{verbatim}
questions: list[RoleTaggedGapQuestion]
    Gap questions for this iteration. Each must declare its role + (if applicable) the
    literal anchor. Roles MUST vary across the batch -- repeating the same role with the
    same anchor is invalid.

RoleTaggedGapQuestion:
  role: entity-anchored | concept-anchored | constraint-relaxed | constraint-tightened |
  inverse-negation
      The structural probe role of this question. Must vary across the batch to enforce
      retrieval-space diversity.
  anchor: str | None
      If role is entity-anchored or constraint-tightened: the literal anchor string from
      the entity ledger that this question uses. MUST appear verbatim inside the question
      text. Null otherwise.
  targets_topic_id: int | None
      If this question is designed to probe a specific topic id from the topic list
      (typically a weakly-evidenced one), the topic id. Null if the question is
      broad/exploratory.
  question: str
      The complete gap question used as a retrieval query.
\end{verbatim}
\end{tcolorbox}
\end{table*}

\begin{table*}[t]
\begin{tcolorbox}[promptbox, title={Query reformulation (Step 4)}]
\textit{Generate alternative phrasings of a gap question to improve retrieval coverage. Invocations per query: \texttt{1\_per\_gap\_question}.}
\vspace{3pt}
\begin{verbatim}
# Query Reformulation

Generate alternative phrasings of a search query to improve retrieval coverage.

**Original query:** {original_query}
**Gap question:** {gap_question}

## Instructions

Generate exactly **{num_reformulations}** alternative phrasings of the gap question that might
retrieve different relevant documents. Each reformulation should:

- Use different vocabulary while preserving the intent
- Be distinct enough from the others to explore diverse retrieval paths
\end{verbatim}
\vspace{4pt}
\textbf{Enforced response schema} --- the model is constrained to return exactly these fields:
\begin{verbatim}
reformulations: list[str]
    List of reformulated query strings.
\end{verbatim}
\end{tcolorbox}
\end{table*}

\begin{table*}[t]
\begin{tcolorbox}[promptbox, title={Novelty judge (Step 5) (1 of 2)}]
\begin{verbatim}
# Document Evaluation -- Gap Judge

> Context: This is an academic information retrieval evaluation. The document may reference
sensitive historical events, conflicts, or social issues. Classify objectively without endorsing
any viewpoint.

Evaluate whether a candidate document contains novel, relevant information for answering a
query.

## Input

**Original query:** {query}

**Topics already found:**
{topic_list}

**Original answer (for reference):**
{answer}

**Candidate document:**
{document}

## Process

### Step 1 -- Relevance check
- Is this document about the **same entity/subject** as the query?
- Does it contain information that **directly helps answer** the query?
- If no to either -> `IRRELEVANT`.
\end{verbatim}
\end{tcolorbox}
\end{table*}

\begin{table*}[t]
\begin{tcolorbox}[promptbox, title={Novelty judge (Step 5) (2 of 2)}]
\begin{verbatim}
### Step 2 -- Novelty check (only if relevant)
Compare the document's information against the topics already found.
- Covers the same ground as an existing topic (even with different details) -> `REDUNDANT` --
  cite the existing topic number.
- Provides a genuinely **new** type of information -> proceed to Step 3.

### Step 3 -- Topic labelling (only if novel)
Create a short label (7-12 words) for the new information facet.

## Granularity rules

- A "topic" is an **information facet** (a type/category of information), not a single data
  point, instance, or example.
- **Paragraph test:** would this topic warrant its own distinct paragraph that does not overlap
  with any existing topic? If not -> `REDUNDANT`.
- Another instance of a category already in the topic list -> `REDUNDANT`, not new.
  - Example: "Results of Steelers vs Ravens games" exists -> a document about a specific game
    result is `REDUNDANT`.
  - Example: "Countries that won the World Cup" exists -> a document about a specific country's
    win is `REDUNDANT`.
- Sub-topics must represent a genuinely **different dimension**, not just deeper detail of the
  same dimension.
  - [+] New: Parent "Hugo Weaving as V" -> "Production challenges with mask acting" (different
    dimension)
  - [-] Redundant: Parent "Sauron created the One Ring" -> "One Ring forged in Mount Doom" (same
    dimension, more detail)
- When in doubt -> `REDUNDANT`. Prefer fewer, higher-quality topics over many overlapping ones.
\end{verbatim}
\end{tcolorbox}
\end{table*}

\begin{table*}[t]
\begin{tcolorbox}[promptbox, title={Novelty judge (Step 5) --- response schema}]
\textbf{Enforced response schema} --- the model is constrained to return exactly these fields:
\begin{verbatim}
verdict: NEW_TOPIC | SUB_TOPIC | REDUNDANT | IRRELEVANT | CONTRADICTORY
    The classification verdict for this document.
topic_label: str | None
    If NEW_TOPIC or SUB_TOPIC: a short descriptive label (5-15 words) for the new
    information facet. Otherwise null.
parent_topic: str | None
    If SUB_TOPIC: the parent topic label or number. Otherwise null.
existing_topics_covered: list[int]
    Topic numbers from the known list that this document also covers.
reason: str
    One sentence explaining the verdict.
\end{verbatim}
The enum admits a fifth label, \texttt{CONTRADICTORY}, outside the four-way taxonomy used throughout: it fired once in $149{,}063$ production verdicts and counts as non-novel wherever verdicts are collapsed.
\end{tcolorbox}
\end{table*}

\begin{table*}[t]
\begin{tcolorbox}[promptbox, title={Topic deduplication (Step 5b)}]
\begin{verbatim}
# Topic Deduplication

Deduplicate newly discovered information topics against the existing list.

## Input

**Existing known topics (already confirmed):**
{existing_topic_list}

**Newly discovered topics (need deduplication):**
{new_topic_labels}

## Instructions

1. **Cluster** new topic labels that are semantically equivalent (same concept, different
   wording).
2. **Merge instances** of the same category (e.g. "India's World Cup wins" and "Australia's
   World Cup wins" -> "Cricket World Cup winning countries").
3. For each cluster, pick the **best single label** as the canonical representative. Prefer
   category-level labels over instance-level ones.
4. **Check overlap:** if any cluster is semantically equivalent to an already-known topic, mark
   `overlaps_existing=true` and provide the existing topic ID number.

## Rules

- Merge labels that are instances/examples of the same information category into **one** cluster
  with a category-level label.
- A new label overlaps an existing topic if it describes another instance of the same category,
  or adds only minor detail to what that topic already covers.
- Only keep a new label as genuinely new if it represents a qualitatively **different** type of
  information (would warrant its own distinct paragraph).
- If a new label adds specificity or a new angle to an existing topic, it is **not** an overlap
  -- keep it as a genuine new topic.
\end{verbatim}
\end{tcolorbox}
\end{table*}

\begin{table*}[t]
\begin{tcolorbox}[promptbox, title={Topic deduplication (Step 5b) --- response schema}]
\textbf{Enforced response schema} --- the model is constrained to return exactly these fields:
\begin{verbatim}
clusters: list[TopicCluster]
    Clusters of semantically equivalent new topic labels. Each cluster becomes one topic.
    Labels that overlap with existing topics should have overlaps_existing=True.

TopicCluster:
  canonical_label: str
      The best single label representing this cluster of topics.
  merged_labels: list[str]
      All original labels that belong to this cluster (including the canonical).
  overlaps_existing: bool
      True if this cluster is semantically equivalent to an already-known topic.
  existing_topic_id: int | None
      If overlaps_existing=True, the ID of the existing topic it overlaps with.
\end{verbatim}
\end{tcolorbox}
\end{table*}

\begin{table*}[t]
\begin{tcolorbox}[promptbox, title={Reader (answer generation)}]
\textit{Generate an answer from retrieved documents. Invocations per query: \texttt{1}.}
\vspace{3pt}
\begin{verbatim}
# Answer Generation

Answer the question using **only** the documents provided. Do not use prior knowledge or any
information not contained in the documents.

## Documents

{documents}

## Question

{query}

## Instructions

- Base your answer solely on evidence in the documents above.
- If the documents don't contain sufficient information, say so explicitly.
\end{verbatim}
\end{tcolorbox}
\end{table*}

\begin{table*}[t]
\begin{tcolorbox}[promptbox, title={Information-equivalence judge (E2.2 only)}]
\textit{Used only for the controlled gold-deletion experiment (Appendix~\ref{sec:appendix-e22}); not part of the deployed Re:CAP loop.}
\vspace{3pt}
\begin{verbatim}
REMOVED:    {gold passage that was excluded}
CANDIDATE:  {doc surfaced by gap probing}

Does CANDIDATE contain the same core facts as
REMOVED? Output YES / PARTIAL / NO + reason.
\end{verbatim}
\end{tcolorbox}
\end{table*}

\end{document}